\documentclass[reqno]{amsart}
\usepackage[a4paper,left=2.6cm,right=2.4cm,top=2.2cm,bottom=2.5cm]{geometry}
\usepackage[english]{babel}
\usepackage[utf8]{inputenc}
\usepackage[T1]{fontenc}
\usepackage{amsmath, amsthm, amssymb}
\usepackage{graphicx}
\usepackage{subcaption}
\usepackage[hyphens]{url}
\usepackage{booktabs}
\usepackage{lineno}
\usepackage{algorithm}
\usepackage{algpseudocode}
\usepackage{siunitx}
\usepackage{makecell}
\usepackage{colortbl}
\usepackage{xcolor}
\usepackage{tikz}
\usepackage{pgfplots}
\usepackage{pifont}
\usepackage[draft,inline,marginclue]{fixme}
\usepackage{mathtools}

\usepackage [autostyle, english = american]{csquotes}
\MakeOuterQuote{"}
\usepackage[backend=biber,doi=true, isbn=false, url=false]{biblatex}
\definecolor{darkblue}{rgb}{0.0, 0.0, 0.55}
\definecolor{darkred}{rgb}{0.55, 0.0, 0.0}
\definecolor{darkgreen}{rgb}{0.0, 0.55, 0.0}

\usepackage{hyperref}
\hypersetup{
    colorlinks=true,
    linkcolor=darkblue,
    filecolor=magenta,
    urlcolor=cyan,
    citecolor=darkblue,
    pdftitle={Conditional LDDMM stochastic interpolant for generative shape modeling},
    pdfauthor={Sarah Katz, Francesco Romor, Jia-Jie Zhu, Alfonso Caiazzo},
    pdfkeywords={tbd}
}

\newtheorem{theorem}{Theorem}[section]

\theoremstyle{definition}
\newtheorem{definition}[theorem]{Definition}
\theoremstyle{remark}
\newtheorem{remark}[theorem]{Remark}
\theoremstyle{definition}

\newcommand{\R}{\mathbb{R}}

\newcommand{\argmin}{\operatornamewithlimits{argmin}}

\newcommand{\defeq}{\coloneqq} 
\newcommand{\by}{\mathbf{y}} 
\newcommand{\bx}{\mathbf{x}} 
\newcommand{\bhx}{\mathbf{x}^h} 
\newcommand{\bm}{\mathbf{m}} 
\newcommand{\bn}{\mathbf{n}}
\newcommand{\bc}{\mathbf{c}} 
\newcommand{\bu}{\mathbf{u}} 

\FXRegisterAuthor{fr}{afr}{FR}
\FXRegisterAuthor{ac}{aac}{AC}
\FXRegisterAuthor{jz}{ajz}{JZ}

\pgfplotsset{compat=1.17}

\begin{document}

\title[Conditional LDDMM stochastic interpolants]{LDDMM stochastic interpolants: an application to domain uncertainty quantification in hemodynamics}

\author[Katz]{Sarah Katz$^{1}$}
\author[Romor]{Francesco Romor$^{1}$}
\author[Zhu]{Jia-Jie Zhu$^{2}$}
\author[Caiazzo]{Alfonso Caiazzo$^{1}$}

\thanks{$^{1}$WIAS, Weierstrass Institute for Applied Analysis and Stochastics, Anton-Wilhelm-Amo-Str. 39, 10117 Berlin, Germany.}
\thanks{$^{2}$KTH Royal Institute of Technology,  Lindstedtsvägen 25, 114\,28\, Stockholm, Sweden.}

\begin{abstract}
We introduce a novel conditional stochastic interpolant framework for generative modeling of three-dimensional shapes. The method builds on a recent LDDMM-based registration approach to learn the conditional drift between geometries.
By leveraging the resulting pull-back and push-forward operators, we extend this formulation beyond standard Cartesian grids to complex shapes and random variables defined on distinct domains.
We present an application in the context of cardiovascular simulations, where aortic shapes are generated from an initial cohort of patients. The
conditioning variable is a latent geometric representation defined by a set of centerline points and the radii of the corresponding inscribed spheres.
This methodology facilitates both data augmentation for three-dimensional biomedical shapes, and the generation of random perturbations of controlled magnitude for a given shape. These capabilities are essential for quantifying the impact of domain uncertainties arising from medical image segmentation on the estimation of relevant biomarkers.
\end{abstract}
\maketitle
\tableofcontents

\section{Introduction}
\label{sec:introduction}
Numerical methods for the simulation of physical systems rely on mathematical models and on the definition and generation of 
computational domains that can accurately represent the problem of interest.
In many applications, the definition of the domain itself might be affected by uncertainties due, e.g., to measurement noise or process-dependent errors
making it necessary to quantify and manage how these affect the results and the relevant quantities of interest.

This work is motivated by the importance of these domain uncertainties in blood flow simulations, where CFD is often used in combination with available medical data 
(e.g., medical images and measurements) to infer biomarkers to diagnose and monitor cardiovascular pathologies.
Quantifying the sensitivity of numerical results to domain variability is driven by two main aspects.
On the one hand, cardiovascular models rely on anatomical data, such as MRI or CT scans, which are (manually, semi-automatically, or automatically) processed to create surface and volume meshes. Possible errors and biases introduced in this process result in unavoidable aleatoric and epistemic uncertainties in the computational domain used for the simulations.
On the other hand, understanding the influence of inter-patient variability is extremely important to draw conclusions relating different hemodynamics quantities beyond a single patient. 

Classical uncertainty quantification (UQ) methods aim at approximating the statistics of quantities of interest depending on the distribution of unknown model inputs. 
In the case of parametric uncertainties, this task can be addressed defining proper sampling strategies of a suitable parameter space. However, 
handling the geometrical uncertainties requires suitable methods for generating instances of anatomical shapes, which cannot be described by a low number of 
geometrical parameters.
Few works have addressed this issue limited to the variability of lumen diameter, in the case of coronary arteries~\cite{Sankaran2016} using an idealized geometry and
multiple patient data, and for the ascending aorta~\cite{Maher2021}, generating domain samples from two-dimensional cross sections obtained during the segmentation.
A more systematic approach was recently proposed in~\cite{bovsnjak2025geometric}, generating multiple samples of ascending aorta from an underlying description based on few geometrical parameters, and including a surface perturbation defined via Gaussian fields. 
Other methods to model inter-patient shape variability from a more general perspective considered sampling based on Statistical Shape Models (SSMs, see, e.g.,~\cite{Adams2020}), a technique used to characterize a patient population with a lower-dimensional parametric shape model which can be combined with a probabilistic model to generate new domains~\cite{goubergrits2022ct, thamsen2021synthetic, thamsen2020unsupervised}.
The idea of a machine learning generative model based on sampling in a low-dimensional latent space has been very 
recently explored in~\cite{tenderini2025deformable}, learning the latent representation
from diffeomorphic maps between an initial set of shapes, and in \cite{fabbri-iollo-etal-2025} considering a variational auto-encoder for the node coordinates.

Generative modeling aims at learning a mapping between a simple prior distribution (e.g., a Gaussian) and a complex target distribution known only through samples, enabling the generation of new realistic instances.
This paradigm has achieved remarkable success in image generation, where, for instance, the conditioning variable can be a text prompt guiding the synthesis process.
Among the most recent and effective approaches, conditional flow matching~\cite{lipman2022flow}, rectified flows~\cite{liu2022flow}, and stochastic interpolants~\cite{albergo2023stochastic} have been introduced as frameworks for generative modeling based on learning continuous-time transport maps between distributions.
These methods construct a time-dependent velocity field (or drift) that continuously transforms the prior into the target distribution, offering advantages in terms of training stability and sampling efficiency compared to classical diffusion models.
Extensions and theoretical connections to broader classes of stochastic processes have also been explored~\cite{liu2023genphysphysicalprocessesgenerative, duong2025telegraphersgenerativemodelkac}.
While these techniques have been extensively developed and applied in two-dimensional settings such as image~\cite{ramesh2022hierarchicaltextconditionalimagegeneration} or video~\cite{liu2024sorareviewbackgroundtechnology} synthesis, their application to three-dimensional geometric data, and in particular to the generation of anatomical shapes, remains largely unexplored.

We propose a new generative modeling method for three-dimensional shapes based on conditional stochastic interpolant and diffeomorphic shape registration.
The approach represents shapes as random variables in the three-dimensional Euclidean space. Using available data, the method learns the drift of a stochastic differential equation (SDE)
defining a stochastic process (\textit{stochastic interpolant}) between different distributions, depending on a conditioning variable in a latent space. 
New shapes are then generated by sampling the conditioning variable and evolving the corresponding SDE.

The main novelty concerns the usage of shape registration maps within the training phase to handle random variables defined on different physical domains. To this purpose, we used
a multilevel ResNet-LDDMM method~\cite{romor2025dataassimilationperformedrobust}, which can robustly and efficiently compute registration maps between anatomical shapes. Analogously to optimal transport-based stochastic interpolant methods, the paths generated by LDDMM stochastic interpolant minimize the LDDMM energy functional. This improves the sampling efficiency and the quality of the generated shapes, as the training process similarly to optimal transport-based stochastic interpolant methods, is guided by the LDDMM energy functional, which is a natural metric for shape registration.

In the context of computational fluid dynamics, a further aspect which cannot be neglected concerns the generation of computational meshes on the deformed domains. In the numerical examples shown in this paper forward simulations are based on a high-order discontinuous Galerkin solver~\cite{ExaDG2020}, which relies on a hierarchy of hexahedral grids. This issue is addressed by defining a suitable extension of the surface deformation onto the whole three-dimensional mesh by iteratively solving a non-linear elasticity problem.

We test the framework in the context of uncertainty quantification for aortic blood flows, validating the capability of the LDDMM stochastic interpolant to sample the shape space.


The rest of the paper is organized as follows. Section~\ref{sec:preliminaries} introduces the background of generative modeling based on 
stochastic interpolants and of shape registration using LDDMM and ResNet-LDDMM. Section~\ref{sec:lddmflow} focuses on the
novel LDDMM stochastic interpolant method, whilst section~\ref{sec:numerical_model_aorta} describes the mathematical models and the numerical methods for blood flow simulations. The application to geometrical uncertainty quantification is detailed in section~\ref{sec:uq},
numerical results are presented in section~\ref{sec:numerical_experiments}, and section~\ref{sec:conclusions} draws the main conclusions.

\begin{figure}
  \centering
  \includegraphics[width=1\textwidth]{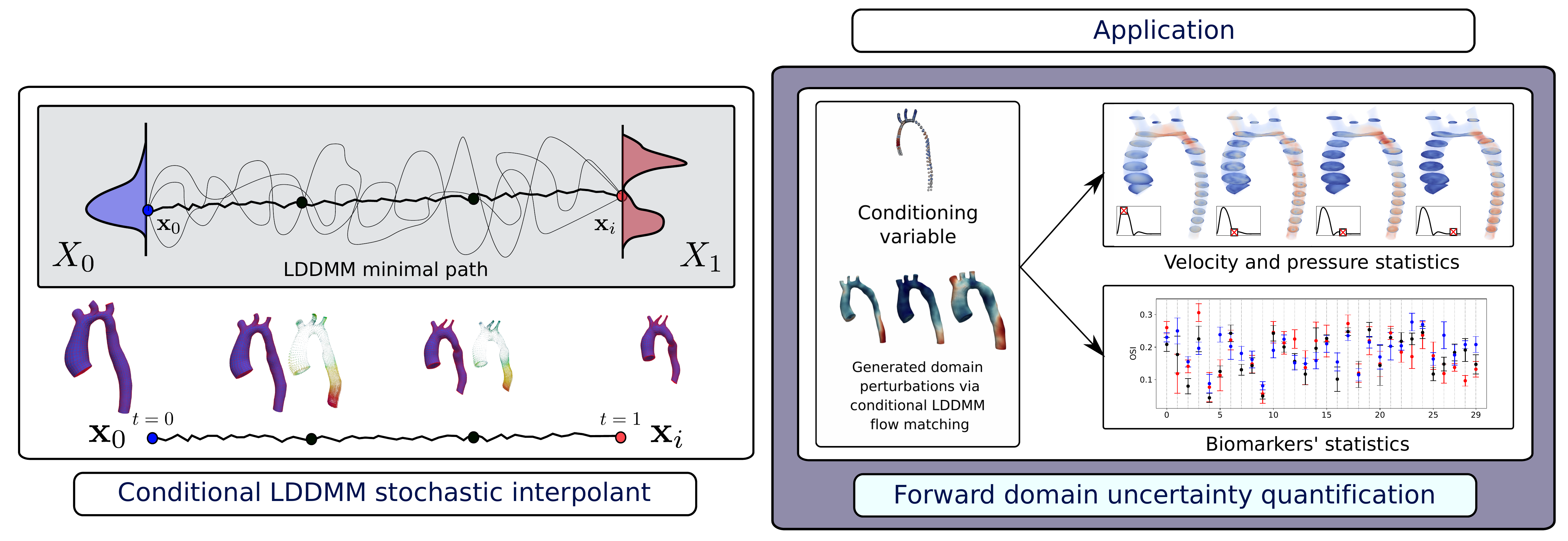}
  \caption{Overview of the proposed method and application in forward domain uncertainty quantification for aortic blood flows.}
\end{figure}

\section{Preliminaries}
\label{sec:preliminaries}
This section focuses on the background for introducing the LDDMM stochastic interpolant framework from a general point of view, reviewing the key concepts of stochastic interpolants in section~\ref{subsec:stochastic_interpolants} and the LDDMM method in section~\ref{subsec:lddmm}. Section \ref{subsec:resnet_lddmm_chamfer} discusses the
application of the recent multilevel ResNet-LDDMM in the case of registration of three-dimensional computational meshes.

\subsection{Stochastic interpolants}
\label{subsec:stochastic_interpolants}
Stochastic interpolants are a mathematical framework that connects different approaches for generative modeling, such as diffusion models, flow matching, and continuous normalizing flows. The main idea behind stochastic interpolants is to \textit{bridge}, in a continuous way, two different probability distributions via a time-dependent random process, which, once learned, can be used to generate new samples.
In what follows, let us denote the probability space as $(\mathcal{X}, \mathcal{A}, P)$, composed of a sample space $\mathcal{X}$, a $\sigma$-algebra of events $\mathcal{A}$, and a probability measure $P$.
For a random variable $X:(\mathcal{X}, \mathcal{A}, P)\rightarrow(\mathbb R^n, \mathcal{B}(\mathbb R^n))$, where $\mathcal{B}$ stands for the Borel $\sigma$-algebra, we denote by $\mu_X$ the probability measure induced by $X$ on $\mathbb{R}^n$ and by $p(\bx)$ its probability density function, when it exists.

 \begin{definition}[Stochastic interpolant]
Let $\mathcal{F}=\{\mathcal{F}_t\}_{t \in [0,1]}$ be a filtration supported on $(\mathcal{X}, \mathcal{A}, P)$, and let
$X_0, X_1:(\mathcal{X}, \mathcal{A}, P)\rightarrow(\mathbb{R}^n, \mathcal{B}(\mathbb{R}^n))$ be two random variables.
A stochastic interpolant between $X_0$ and $X_1$  is a stochastic process
\[
I:([0,1]\times\mathbb{R}^n, \mathcal{B}([0,1])\otimes\mathcal{F})\rightarrow(\mathbb{R}^n, \mathcal{B}(\mathbb{R}^n)),
\]
adapted to $\mathcal{F}$, that allows one to bridge two random variables $X_0$ and $X_1$, such that:
\begin{equation*}
  I_{t=0} = X_0, \quad I_{t=1} = X_1, \quad P\text{-}a.s.
\end{equation*}
\end{definition}

Let there be given two data distributions, which we assume are represented by two random variables $X_0$ and $X_1$, with probability densities $p(\bx_0)$ and $p(\bx_1)$ with respect to the Lebesgue measure. The stochastic interpolant between $X_0$ and $X_1$ is modeled as the solution of an SDE
\begin{equation}
  \label{eq:sde_drift}
\left\{
\begin{aligned}
  & dI_t = b^{\theta}_t(I_t) dt + \sigma_t dW_t, \; t \in [0,1],\\
  & I_0 \sim p(\bx_0), \; I_1 \sim p(\bx_1),
  \end{aligned}
  \right.
\end{equation}
where $W_t$ is a standard Brownian motion, $\sigma_t$ is a time-dependent diffusion coefficient, and
the time-dependent drift $b_t^{\theta}:\mathbb{R}^n\rightarrow\mathbb{R}^n$ needs to be learned
%
minimizing the following \textit{stochastic interpolant} loss function:
\begin{equation}\label{eq:flowmatchingloss}
  \mathcal{L}(b^{\theta}) = \int_0^1 \mathbb{E}_{I_t}[\|b^{\theta}_t(I_t) - u_t(I_t)\|^2] dt,
\end{equation}
depending on a parameter $\theta$, for a prescribed \textit{drift} $u_t:\mathbb{R}^n\rightarrow\mathbb{R}^n$.

\begin{figure}[htp!]
    \includegraphics[width=1\textwidth]{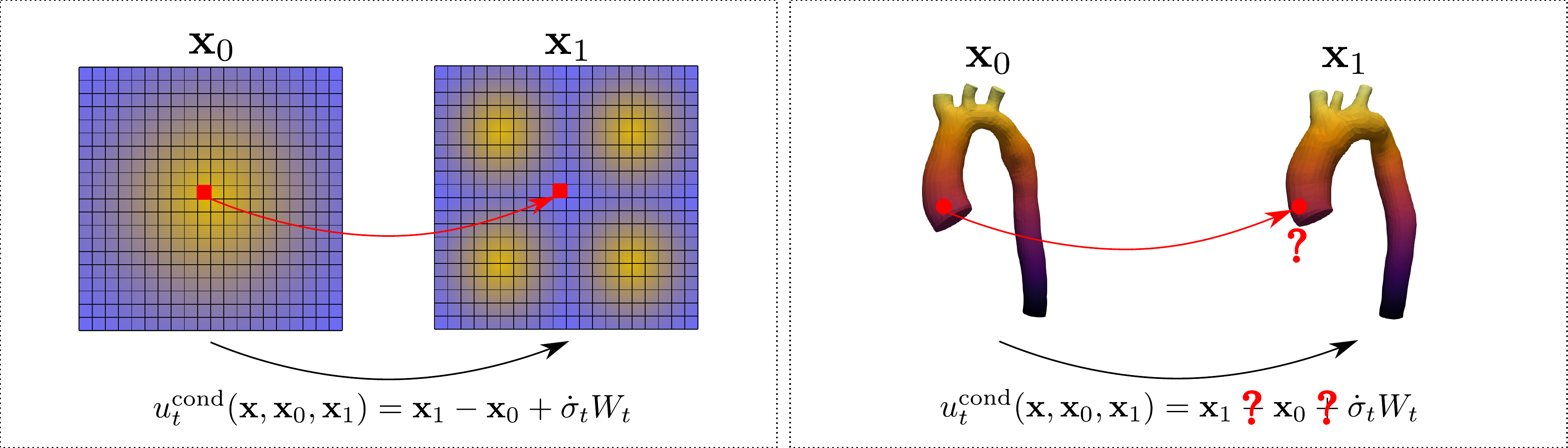}
    \caption{Left: On a two-dimensional grid, the prescribed velocity field $u^{\text{cond}}_t$ \eqref{eq:conditional_drift} can be defined by computing the difference between
    the realizations of the random variables $X_0$ and $X_1$. Right: In the case of a three-dimensional shape (described, e.g., as a surface mesh),
    a definition of $u^{\text{cond}}_t$ based on the difference \eqref{eq:conditional_drift} is no longer possible, since there is no direct correspondence between points.}
    \label{fig:2dvsshapes}
\end{figure}

Typically, during the training, the drift $u_t$ is conditioned on $\bx_0$ and $\bx_1$ (e.g., two images as in~Figure~\ref{fig:2dvsshapes}) sampled from $X_0$ and $X_1$, respectively, and is given in the form:
\begin{equation}
  \label{eq:conditional_drift}
  u_t: = u_t^{\text{cond}}(\bx_0, \bx_1) = \dot{\alpha}_t \bx_0 + \dot{\beta}_t \bx_1 + \dot{\sigma}_t W_t, \quad t\in[0,1],
\end{equation}
where $\alpha_t, \beta_t, \sigma_t:\mathbb{R}\rightarrow\mathbb{R}$ are deterministic functions satisfying $\alpha_0=1$, $\alpha_1=0$, $\beta_0=0$, $\beta_1=1$,
$\sigma_0=0$, $\sigma_1=0$.
The aim is to approximate a stochastic interpolant $\tilde{I}_t$ between $X_0$ and $X_1$ as
\begin{equation}
  \label{eq:stochastic_interpolant}
  \tilde{I}_t = \alpha_t X_0 + \beta_t X_1 + \sigma_t W_t, \quad t\in[0,1].
\end{equation}
A common choice is $\alpha_t = 1-t$, $\beta_t = t$, and $\sigma_t = \sigma \sqrt{t(1-t)}$ for some fixed parameter $\sigma>0$.

Considering the conditioned drift, the loss function \eqref{eq:flowmatchingloss} becomes
\begin{equation}
  \label{eq:flow_matching_loss}
  \mathcal{L}(b^{\theta}) = \int_0^1 \mathbb{E}_{X_0,X_1}[\|b^{\theta}_t(I_t) - u_t^{\text{cond}}(X_0, X_1)\|^2] dt,
\end{equation}
with the  vector field $u_t^{\text{cond}}$ defined in equation~\eqref{eq:conditional_drift}.

The definitions of the stochastic interpolant in the form \eqref{eq:stochastic_interpolant}, and the corresponding
conditional drift \eqref{eq:conditional_drift}, require computing the sums and differences of discretized realizations of two random variables $X_0$ and $X_1$ in $\mathbb{R}^n$.
This operation can be naturally defined in the case of images of the same size, discretized, e.g., as two- or three-dimensional grids with a one-to-one correspondence between pixels (see Figure~\ref{fig:2dvsshapes}, left). However, it cannot be directly extended to the case of random variables modeling physical fields on
general shapes (see Figure~\ref{fig:2dvsshapes}, right), which might be defined on point clouds with different cardinalities or through unstructured (surface or volume) meshes with different numbers of vertices and elements. A generalization to this case, and the application in the context of numerical methods for PDEs, will be discussed in section \ref{sec:lddmflow}.

\begin{remark}[Equivalent ODE flow formulation]
We can rewrite the stochastic interpolant formulation with respect to its probability density $p_{I_t}$ as
  \begin{equation}\label{eq:SI-equiv}
    \begin{cases}
      \partial_t p_{I_t} + \nabla\cdot(p_{I_t} v^{\text{SI}}_t) = 0, \quad t\in[0,1], \\
      p_{I_{t=0}} = p_{X_0}, \quad p_{I_{t=1}} = p_{X_1},
    \end{cases}
  \end{equation}
which is equivalent to equation~\eqref{eq:flow_matching_loss} via its corresponding Fokker-Planck equations, with $v^{\text{SI}}_t=b_t-\tfrac{\sigma_t^2}{2}\nabla\log{p_{I_t}}$.
Sampling from the stochastic interpolant then requires solving the ODE
  \begin{equation}
    \label{eq:ode_flow}
    \left\{
    \begin{aligned}
    & \partial_t \phi^{\text{SI}}_t(x) = v^{\text{SI}}_t(\phi^{\text{SI}}_t(x)), \quad t\in[0,1],\\
    & \phi^{\text{SI}}_0(x) = x\,,
        \end{aligned}
        \right.
  \end{equation}
  for a map $\phi^{\text{SI}}_t:[0,1]\times\mathbb{R}^n\rightarrow\mathbb{R}^n$, and setting $p_{I_t}$ as the push-forward of $p_{X_0}$ with respect to the map $\phi^{\text{SI}}$, i.e.,
  $p_{I_t} = p_{X_0} \circ \left(\phi_t^{\text{SI}}\right)^{-1}$.

   The existence of the solution $\phi^{\text{SI}}$ of \eqref{eq:ode_flow} is guaranteed under suitable regularity assumptions on $v^{\text{SI}}_t$, e.g., when  $v^{\text{SI}}\in L^2([0,1], L^2(\mathbb{R}^n))$~\cite{ambrosio2005gradient} and:
  \begin{equation}
    \label{eq:ode_flow_existence}
    \int_{[0,1]} \sup_{x\in B} \lVert v^{\text{SI}}_t(x)\rVert + \text{Lip}(v^{\text{SI}}_t, B) \, dt < \infty,\quad \forall B \subset \mathbb{R}^n \text{ compact}.
  \end{equation}

\end{remark}

\begin{remark}[Regularizers for stochastic interpolants]
  \label{rmk:stochastic_interpolants_reg}
The definition \eqref{eq:stochastic_interpolant} is limited to linear interpolants between $X_0$ and $X_1$, which may not be optimal in terms of minimizing the
kinetic energy density. For this purpose, the minimization can be formulated as
  \begin{align*}
    W^2_2(p_{X_0}, p_{X_1}) &= \inf_{p_{I_t}, v^{\text{SI}}_t} \int_0^1 \mathbb{E}_{I_t}[\|v^{\text{SI}}_t\|^2] dt ,\quad\mbox{ s.t. $p_{I_t},\ v^{\text{SI}}_t$ satisfy \eqref{eq:SI-equiv},}\\
    &\mbox{with $\mathbb{E}_{I_t}[\|v^{\text{SI}}_t\|^2]=\int_{\mathbb{R}^n}\lVert v^{\text{SI}}_t(x)\rVert^2dp_t(x)$}
  \end{align*}
  where  $W^2_2$ stands for the 2-Wasserstein distance in the dynamic formulation (Benamou–Brenier) between the probability densities $p_{X_0}$ and $p_{X_1}$.
 When approximating physical fields obtained as solutions of PDEs, additional constraints may be imposed on the drift $b_t$ to enforce physical consistency, e.g., divergence-free constraints for incompressible fluids or a prescribed kinetic energy rate of change (see, e.g., ~\cite{muecke2025physicsawaregenerativemodelsturbulent}).
\end{remark}

\subsection{LDDMM for image registration}
\label{subsec:lddmm}
The problem of image registration concerns finding a smooth transformation between a \textit{source/template} image (defined, e.g., as a point cloud or a mesh) and a \textit{target} image. The Large Deformation Diffeomorphic Metric Mapping (LDDMM) method~\cite{Trouv1998} addresses this task by modeling the transformation as the flow of an ordinary differential equation (ODE), ensuring smooth and invertible mappings.

Let us consider the source and the target objects denoted by $S$ and $T$, respectively ($S,T \subset\mathbb{R}^d$), and defined by their characteristic functions $\chi_S:\mathbb{R}^d\rightarrow\mathbb{R}$ and $\chi_T:\mathbb{R}^d\rightarrow\mathbb{R}$, respectively.
Without loss of generality, let us assume that both objects are contained in an open bounded set $U$, i.e.,
\begin{equation*}
  \text{supp}(\chi_S), \text{supp}(\chi_T) \subset U \subset \R^d.
\end{equation*}
The LDDMM method seeks a time-dependent velocity field $v^{\text{\tiny LDDMM}}_t:U \rightarrow \R^d$, such that the
flow $\phi^{\text{\tiny LDDMM}}_t:U \rightarrow U$, solution to
\begin{equation}
  \label{eq:lddmm_flow}
  \left\{
  \begin{aligned}
  & \partial_t\phi^{\text{\tiny LDDMM}}_t(\bx) = v^{\text{\tiny LDDMM}}_t(\phi^{\text{\tiny LDDMM}}_t(\bx)), \quad t \in [0,1],\\
  & \phi^{\text{\tiny LDDMM}}_0(\bx) = \bx,
  \end{aligned}
    \right.
\end{equation}
smoothly transforms the source object $S$ into the target $T$.

We assume that
\begin{equation}
  \label{eq:lddmm_velocity_assumption}
  v^{\text{\tiny LDDMM}}\in L^2(I, H^s(U)), \quad \int_{I}\lVert v^{\text{\tiny LDDMM}}_t \rVert_{H^s(U)}^2 \, dt < \infty,
\end{equation}
so the Sobolev embedding theorem implies $v\in L^2(I, \mathcal{C}^{1,\alpha}(U))$ with $\alpha=s-\tfrac{d}{2}-1>0$. Hence, for each $t \in [0,1]$, the flow $\phi^{\text{\tiny LDDMM}}_t$ is a diffeomorphism, $\phi^{\text{\tiny LDDMM}}\in\mathcal{C}^{0}(I, \text{Diff}^{1}(U))$.
The transformation of the source image at time $t$ is given by:
\begin{equation*}
  \chi_{S_t}(\bx) = \chi_S(\left(\phi^{\text{\tiny LDDMM}}_t\right)^{-1}(\bx)),\quad \forall t\in [0,1].
\end{equation*}
The LDDMM method seeks to minimize the following augmented energy functional:
\begin{equation}
  \label{eq:lddmm_energy}
  E(v) = \int_0^1 \|v^{\text{\tiny LDDMM}}_t\|_{H^s(U)}^2 \, dt + \lambda \int_U |\chi_{S_t}(\bx) - \chi_T(\bx)|^2 \, d\bx,
\end{equation}
where $\|v^{\text{\tiny LDDMM}}_t\|_{H^s(U)}$ ensures smoothness of the velocity field, and $\lambda > 0$ is a regularization parameter balancing the smoothness of the transformation and the fidelity to the target image. We refer the reader to~\cite{Dupuis1998} for additional details on the existence of a minimizer with the required regularity.

\begin{definition}[Push-forward and pull-back operators]
Given a diffeomorphism $\phi:U\rightarrow U$, we can define the push-forward operator $\phi_{\#}$ and the pull-back operator $\phi^{\#}$ that allow us to transport fields defined on $U$ via:
  \begin{equation*}
    \phi_{\#}(f) = f \circ \phi^{-1}, \quad \phi^{\#}(f) = f \circ \phi,
  \end{equation*}
  for any $f:U\rightarrow\mathbb{R}^m$. These operators will be useful in section~\ref{sec:lddmflow} to define a map between different computational domains.
\end{definition}

\begin{remark}[Analogies between LDDMM and diffusion models]
The LDDMM method and stochastic interpolants (section \ref{subsec:stochastic_interpolants}) are defined in different settings (deterministic and probabilistic) and address different problems. However, both approaches are based on flows of ODEs that aim to transform a source (subdomain or probability measure, respectively) into a target through suitable time-dependent velocity fields -- compare equations~\eqref{eq:ode_flow} and~\eqref{eq:lddmm_flow}, as well as the regularity assumptions~\eqref{eq:ode_flow_existence} for the stochastic interpolant and~\eqref{eq:lddmm_velocity_assumption} for the velocity field in LDDMM.
\end{remark}

\subsection{ResNet LDDMM for registration of surface meshes}
\label{subsec:resnet_lddmm_chamfer}
Let us define a generic mesh as $\mathcal{M} = (\Omega, \bm, E)$, where $\Omega\subset\mathbb{R}^d$ stands for the (physical) domain or a piecewise smooth manifold associated with it, $\bm = \{\bm^i\}_{i=1}^{N}\subset\mathbb{R}^d$ is the array of the mesh vertices' coordinates, and $E$ is the set of $d$-dimensional cells defined as subsets of vertices in $\bm$ (for instance, $E$ is a set of subsets of $4$ vertices for tetrahedral meshes in 3D).
In what follows, we assume that there exist pairwise bi-Lipschitz homeomorphisms between shapes.
This assumption does not affect the generality of the framework and is consistent with the setting where all the considered shapes represent the same underlying physical system, e.g., a PDE on different domains with a given set of boundaries and corresponding
boundary conditions -- see section~\ref{sec:numerical_experiments}.
However, we do not assume any a priori relation on the sizes of the discrete spaces (number of vertices, faces, or elements) of the different shapes.

Instead of characteristic functions (as done in section~\ref{subsec:lddmm}), we now consider source and target shapes represented by computational meshes $\mathcal{M}_S=(\Omega_S, \bm_S, E_S)$ and $\mathcal{M}_T=(\Omega_T, \bm_T, E_T)$, respectively.
In this case, the minimization problem \eqref{eq:lddmm_energy} is solved by minimizing a shape dissimilarity measure between the meshes.
A common choice is the so-called Chamfer distance between the point clouds given by mesh vertices, defined by
\begin{equation}\label{eq:chamfer-0}
  d_{\text{Chamfer}}(\bm_S, \bm_T) = \sum_{\bx \in \bm_S} \min_{\by \in \bm_T} \|\bx - \by\|^2 + \sum_{\by \in \bm_T} \min_{\bx \in \bm_S} \|\bx - \by\|^2.
\end{equation}

In this setting, the LDDMM registration problem can be formulated as: find the time-dependent velocity field $v^{\text{\tiny LDDMM}}_t:U \rightarrow \R^d$ that minimizes the energy functional:
\begin{equation}
  \label{eq:lddmm_chamfer_energy}
  E(v) = \int_0^1 \|v^{\text{\tiny LDDMM}}_t\|_{H^s(U)}^2 \, dt + \lambda \, d_{\text{Chamfer}}(\phi^{\text{\tiny LDDMM}}_1(\bm_S), \bm_T),
\end{equation}
where $\phi^{\text{\tiny LDDMM}}_t$ is the flow generated by $v^{\text{\tiny LDDMM}}_t$ (solution of equation~\eqref{eq:lddmm_flow}), and $\lambda > 0$ is a regularization parameter.

Depending on the choice of the numerical optimization method and the parametrization of the velocity field $v^{\text{\tiny LDDMM}}_t$, several variants of LDDMM can be defined.
In what follows, we focus on the recent and efficient multilevel ResNet LDDMM~\cite{romor2025dataassimilationperformedrobust} (based on~\cite{9775044}), in which the velocity field is parametrized using a multilevel ResNet architecture, and the minimization considers a generalization of the Chamfer distance \eqref{eq:chamfer-0},
accounting for dissimilarity between boundary elements and mesh centerlines, and with regularization terms to
preserve mesh quality during the deformation (see~\cite{romor2025dataassimilationperformedrobust} for details).
In this case, the regularity of the registration map changes from the space of diffeomorphisms $\phi^{\text{\tiny LDDMM}}_t\in\text{Diff}^1(U)$ to the space of bi-Lipschitz homeomorphisms, which is more suitable for applications in which the shapes to be registered are not necessarily diffeomorphic (e.g., due to the presence of corners or edges). The invertibility constraint is not implicitly imposed on the ResNet architecture, but it can be checked a posteriori~\cite{romor2025dataassimilationperformedrobust}.
Moreover, the size of the mesh is refined during the training stage, allowing for efficient computation and scalability to large datasets of 3D shapes.

\begin{figure}
  \includegraphics[width=1\textwidth]{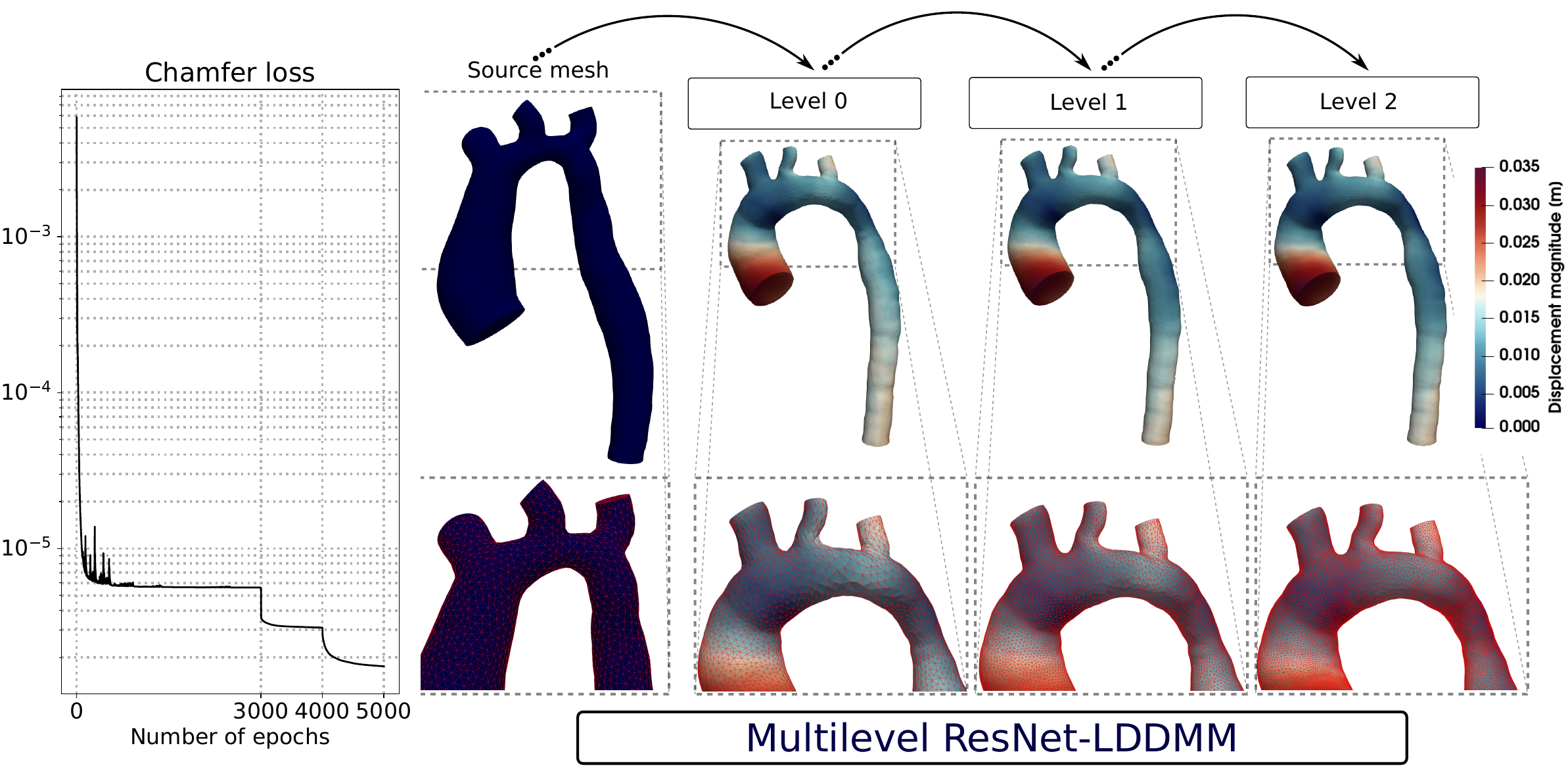}
  \caption{Example of application of the multilevel ResNet-LDDMM~\cite{romor2025dataassimilationperformedrobust} for the registration of computational meshes
  of aortic shapes. In this example, meshes are refined after 3,000, 4,000, and 5,000 epochs.}
  \label{fig:registration_multigrid}
\end{figure}

\section{Conditional LDDMM stochastic interpolant}
\label{sec:lddmflow}
\subsection{Stochastic interpolant between numerical solutions on different shapes}
This work is motivated by the application of generative modeling in the context of the numerical solution of PDEs, in which the computational domain
is discretized by a mesh and the solution is sought in a discrete space, e.g., of piecewise polynomials defined on the mesh elements.

Building on the framework of stochastic interpolants (section~\ref{subsec:stochastic_interpolants}), we consider
data distributions and random variables representing domain coordinates or numerical solutions of a given PDE, supported on a pool of continuous, three-dimensional domains, denoted as $\{\Omega_i\}_{i=0}^{N_{\text{shapes}}}$.
In this case, it is necessary to define a map between different computational domains on which different fields are supported. This is a crucial difference with respect to the application to the more classical case of image generation, where data structures are naturally defined on objects of the same size (e.g., images).

Let us assume that there exists an open, bounded, and connected set $U \subset \mathbb{R}^3$ such that $\Omega_i \subset U$ for $i=0,\hdots,N_{\text{shapes}}$, and
let us introduce a background Hilbert space $\mathbb{H}(U)$, e.g., a functional space on $U$, on which all the numerical solutions corresponding to the different computational domains are defined.

Among the available domains, let $\Omega_0 \subset \mathbb{R}^3 $ denote the \textit{template} (or reference) shape,
and let $\mathcal{M}_0=(\Omega_0, \bm_0, E_0)$ be the corresponding computational mesh.
Using the notation of the probabilistic framework introduced in subsection~\ref{subsec:stochastic_interpolants}, we
consider a source random variable $X_0:(\mathcal{X}, \mathcal{A}, P)\rightarrow \mathbb{H}(U)$ supported on $\Omega_0$, whose realizations $\bx_0\in \mathbb{H}(U)$ belong to the background Hilbert space, and whose discretizations $\bhx_0\in \mathbb{R}^{n_0}$ have the dimension
($n_0$) of the space of piecewise polynomial functions (e.g., basis functions of a finite element method) defined on the mesh $\mathcal{M}_0$.

The target random variable $X_1:(\mathcal{X}, \mathcal{A}, P)\rightarrow \mathbb{H}(U)$ will represent any of the remaining shapes $\Omega_i\subset\mathbb{R}^3$, $i\in\{1,\dots, N_{\text{shapes}}\}$, with realizations $\bx_i\in \mathbb{H}(U)$, and discretizations $\bhx_i \in \mathbb{R}^{n_i}$ supported on the corresponding meshes $\mathcal{M}_i=(\Omega_i, \bm_i, E_i)$
with $n_i$ degrees of freedom.
As noted in section~\ref{subsec:resnet_lddmm_chamfer}, no relation is assumed a priori on the sizes of the discretizations corresponding to different shapes; in particular,
$n_i \neq n_j$ for $i \neq j$.

\begin{remark}[General realizations of $X_0$ and $X_1$]
The framework has so far been introduced from a general perspective, i.e., considering that the realizations $\bx_0, \bx_1 \in \mathbb{H}(U)$ of $X_0$ and $X_1$, and the corresponding
discretizations $\bhx_0\in\mathbb{R}^{n_0}$ and $\bhx_i\in\mathbb{R}^{n_i}$, can represent any scalar or vector fields defined on the computational mesh.
In particular, the distributions could represent numerical solutions of an incompressible flow (e.g., velocity and pressure), a deformation field defined on the computational mesh, or the coordinates of the support points of the finite element spaces. Section~\ref{sec:uq} will focus on the latter case, as an application of shape generative modeling.
\end{remark}

\begin{remark}
In this study, for the purpose of validating the conditional LDDMM stochastic interpolant method, we restrict ourselves to a fixed, single template domain.
In general, it is possible to consider varying domains $\Omega_j\subset\mathbb{R}^3$ also for the random variable $X_0$, which will increase the overall computational effort required.
\end{remark}

\subsection{Generative modeling with conditional LDDMM stochastic interpolants}
\label{subsec:lddmm_flow_matching}

We propose a new generative model that combines stochastic interpolants (section~\ref{subsec:stochastic_interpolants}) with ResNet-LDDMM registration (section \ref{subsec:resnet_lddmm_chamfer}), in order to define a suitable conditioned drift for data distributions on variable domains.

Firstly, let us observe that, to apply the framework introduced in section~\ref{subsec:stochastic_interpolants}, it is necessary to extend the definition of noise to the
background space $\mathbb{H}(U)$, i.e., to define a $Q$-Wiener process $W_t^Q$ on $\mathbb{H}(U)$, with the covariance operator $Q:\mathbb{H}(U)\rightarrow\mathbb{H}(U)$ being a symmetric, positive, trace-class operator. We consider an extension based on the standard theory of $Q$-Wiener processes in Hilbert spaces; see~\cite{da2014stochastic} for more details.

We denote by $C:(\mathcal{X}, \mathcal{A}, P)\rightarrow(\mathbb{R}^c, \mathcal{B}(\mathbb{R}^c))$, $c>0$, an additional \textit{conditioning} random variable that will trigger the generative process. In the field of image~\cite{ramesh2022hierarchicaltextconditionalimagegeneration} or video~\cite{liu2024sorareviewbackgroundtechnology} generation, for example, $C$ typically represents the embedding of a text prompt.

\begin{figure}[htp!]
  \includegraphics[height=3.5cm,trim=28cm 0cm 0cm 0cm,clip=true]{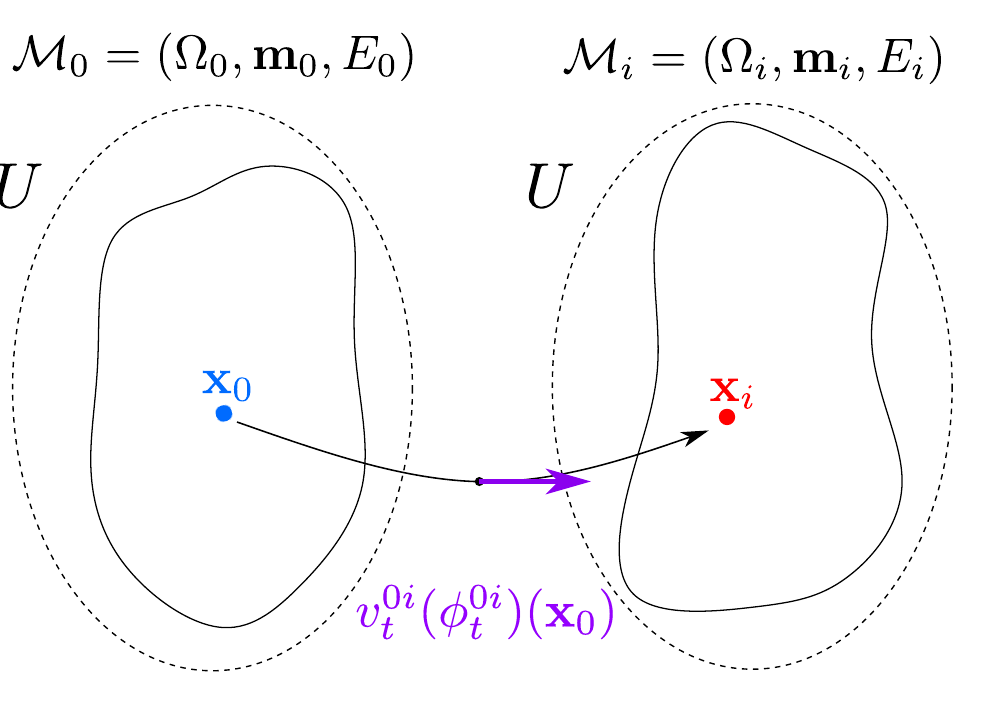}\hspace{.5cm}
  \includegraphics[height=3.5cm,trim=0cm 0cm 18.5cm 0cm,clip=true]{img/conditioned_field.pdf}
  \caption{The conditional drift $u_t^{\text{cond}}(\bx_0, \bx_i)$ is defined based on the LDDMM registration.
  When $\mathbf{x}_0$ and $\bx_i$ represent domain coordinates (left), the conditional drift is computed from the registration field ${\phi}^{0,i}$ between the two realizations. When
$\mathbf{x}_0$ and $\bx_i$ represent physical fields/solutions of PDEs (right), the conditional drift is computed, at each intermediate step $t \in [0,1]$, considering
the push-forward of ${\phi}^{0,i}$ (registration of $\bx_0$ onto $\bx_i$) and the pull-back of ${\psi}^{0,i}$ (registration of $\bx_i$ onto $\bx_0$).
}\label{fig:sketch_ut_cond}
  \end{figure}

Let two samples $\bx_0$ and $\bx_i$ of two random variables $X_0$ and $X_1$ respectively, be given, supported on the meshes $\mathcal{M}_0$ (template shape) and $\mathcal{M}_i$.
The generative model follows the approach described in section \ref{subsec:stochastic_interpolants}, considering the stochastic interpolant solution to the SDE
\begin{equation}
  \label{eq:lddmm_flow_matching_spde}
  \left\{
  \begin{aligned}
  & dI_t = b^{\theta}_t(I_t, C) dt + \sigma_t dW^Q_t,\\
  & I_{t=0} = X_0, \quad I_{t=1} = X_1,
  \end{aligned}
  \right.
\end{equation}
where the coefficient $b^{\theta}_t(I_t, C)$ is trained by minimizing the loss function
\begin{equation}
  \label{eq:lddmm_flow_matching_loss}
  \mathcal{L}(b^{\theta}) = \int_0^1 \mathbb{E}_{X_0,X_1, C}[\|b^{\theta}_t(I_t, C) - u_t^{\text{cond}}(X_0, X_1)\|^2] dt.
\end{equation}
for a suitable conditional drift $u_t^{\text{cond}}$. 
\begin{figure}[htp!]
  \includegraphics[width=\textwidth]{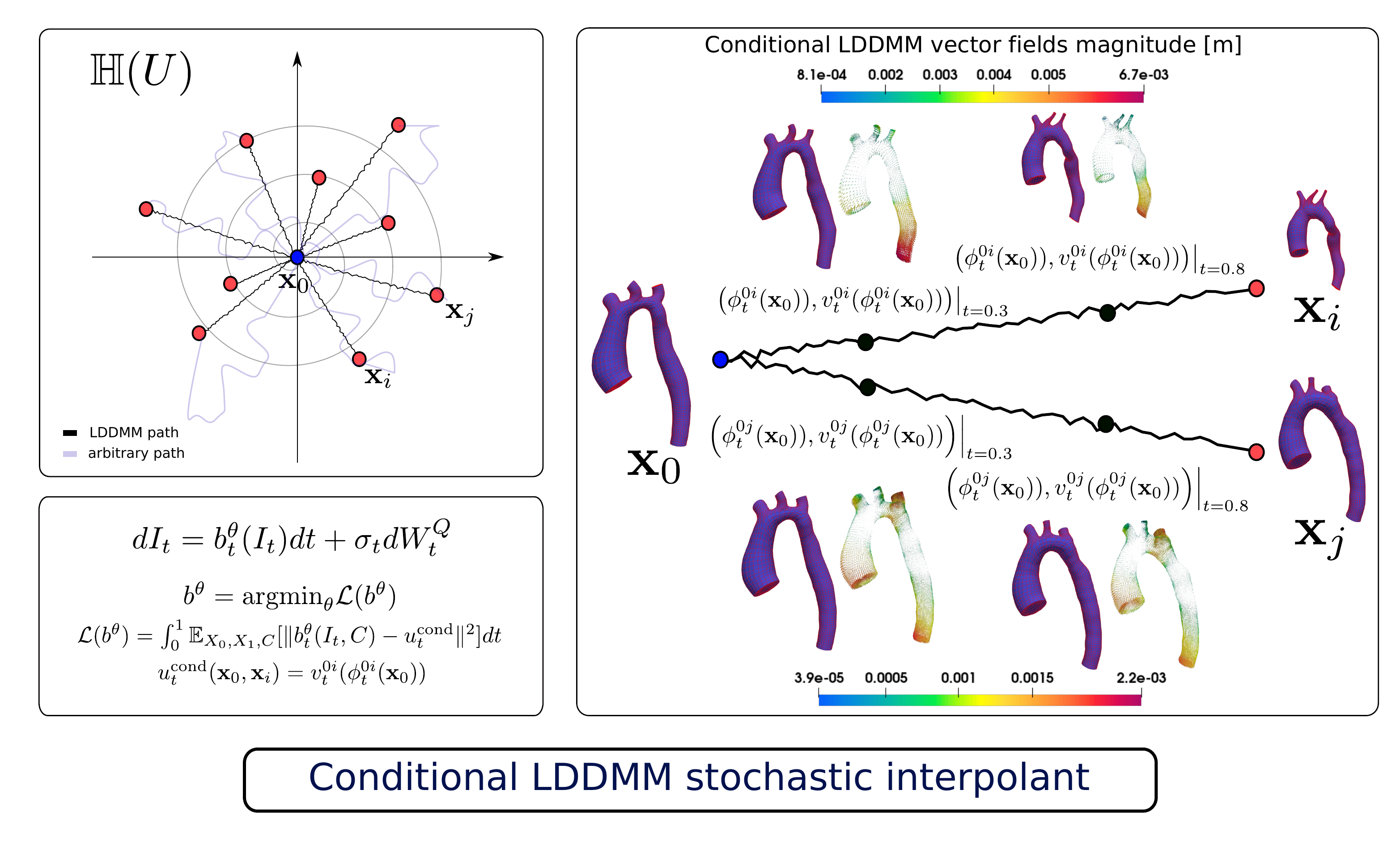}
  \caption{Graphical overview of the proposed conditional LDDMM stochastic interpolant method. The map defining the
  conditional drift $b^{\theta}$ as a function of the conditioning variable $C$ is learned registering the available shapes onto a common reference ($\bx_0$),
  in order to find a stochastic interpolant between the two distributions.
  }
\end{figure}
Unlike the case discussed in section \ref{subsec:stochastic_interpolants}, prescribing $u^{\text{cond}}:[0,1]\times\mathbb{H}(U)\times\mathbb{H}(U)\rightarrow\mathbb{H}(U)$,
requires defining a time-dependent velocity field $u_t^{\text{cond}}(\bx_0, \bx_i)$ on an \textit{intermediate} domain (at time $t$) between $\Omega_0\subset U$ and $\Omega_i\subset U$.
This task will be addressed by computing the LDDMM registration map $\phi_t^{0,i}:\mathbb{R}^3\rightarrow\mathbb{R}^3$ between the meshes $\mathcal{M}_0$ and $\mathcal{M}_i$ (i.e., between the corresponding point clouds $\bm_0$ and $\bm_i$) with the multilevel ResNet LDDMM framework (see section~\ref{subsec:resnet_lddmm_chamfer}).
Namely, if the
  realizations $\bx_0$ and $\bx_i$ represent sets of coordinates within the computational domains (the case considered in this work),
  we set
\begin{equation}
  \label{eq:conditional_drift_lddmm_coordinates}
  u_t^{\text{cond}}(\bx_0, \bx_i) = v^{0,i}_t({\phi}^{0,i}_t(\bx_0))+ \sigma_t \dot{W}^Q_t, \quad t\in[0,1],\qquad\text{(LDDMM flow matching)}
\end{equation}
where $v^{0,i}_t$ is the LDDMM velocity field obtained by registering 
$\mathcal{M}_0$ to $\mathcal{M}_i$, equation~\eqref{eq:lddmm_flow} (see also figure~\ref{fig:sketch_ut_cond}, left) for the training of $b^{\theta}_t(I_t, C)$
\eqref{eq:lddmm_flow_matching_loss} for a given realization of the conditioning variable.
Next, we sample different solutions $I_t[\bc]$ of \eqref{eq:lddmm_flow_matching_spde} and define the new deformed domain as
$\mathbb{E}[I_t[\bc]]$.
The training and generative phases of the proposed conditional LDDMM stochastic interpolant method are summarized in Algorithms~\ref{alg:lddmm_fm_training} and~\ref{alg:lddmm_fm_generation}, respectively. More details on the training procedure for the conditioning drift are provided in appendix~\ref{sec:training_details}.

\begin{algorithm}[!tph]
\caption{Conditional LDDMM Stochastic Interpolant -- Training Phase}
\label{alg:lddmm_fm_training}
\begin{algorithmic}[1]
\Require Template mesh $\mathcal{M}_0$, target meshes $\{\mathcal{M}_i\}_{i=1}^{N_{\text{shapes}}}$, conditioning variables $\{\bc_i\}_{i=1}^{N_{\text{shapes}}}$
\Ensure Trained drift network $b^{\theta}_t$
\For{$\text{epoch} = 1$ to $N_{\text{epochs}}$}
    \State Sample batch of indices $\mathcal{B} \subset \{1, \ldots, N_{\text{shapes}}\}$
    \For{$i \in \mathcal{B}$}
        \State Compute registration map $\phi_t^{0,i}$ between $\mathcal{M}_0$ and $\mathcal{M}_i$ and velocity $v_t^{0,i}$ (using the multilevel ResNet LDDMM -- section~\ref{subsec:resnet_lddmm_chamfer})
        \State Sample time: $t \sim \text{Uniform}(0, 1)$
        \State Compute intermediate state: $I_t = \phi_t^{0,i}(\bx_0)$
        \State Compute conditional drift: $u_t^{\text{cond}}(\bx_0, \bx_i) = v_t^{0,i}(\phi_t^{0,i}(\bx_0))+ \sigma_t \epsilon$, where $\epsilon \sim \mathcal{N}(0, I)$
        \State Compute loss: $\mathcal{L}_i = \|b^{\theta}_t(I_t, \bc_i) - u_t^{\text{cond}}(\bx_0, \bx_i)\|^2_2$
    \EndFor
    \State Update parameters: $\theta \leftarrow \theta - \nabla_{\theta}\frac{1}{|\mathcal{B}|}\sum_{i\in\mathcal{B}}\mathcal{L}_i$
\EndFor
\State \Return $b^{\theta}_t$
\end{algorithmic}
\end{algorithm}

\begin{algorithm}[!tph]
\caption{Conditional LDDMM Stochastic Interpolant -- Generative Phase}
\label{alg:lddmm_fm_generation}
\begin{algorithmic}[1]
\Require Template sample $\bx_0$ on mesh $\mathcal{M}_0$, conditioning variable $\bc$,
trained drift network $b^{\theta}_t$, time discretization $\{t_k\}_{k=0}^{K}$ with $t_0=0$, $t_K=1$
\Ensure Generated sample $\left.\mathbb{E}[I_t[\bc]]\right|_{t=1}$ on deformed domain
\State Initialize: $I_0 \leftarrow \bx_0$
\State Initialize: $S \leftarrow 0$
\For{$s = 1$ to $N_{\text{samples}}$}
    \State Initialize: $I_0^{(s)} \leftarrow \bx_0$
    \For{$k = 0$ to $K-1$}
        \State Compute drift: $b_k^{(s)} = b^{\theta}_{t_k}(I_{t_k}^{(s)}, \bc)$
        \State Sample noise: $\xi_k^{(s)} \sim \mathcal{N}(0, I)$
        \State Update state: $I_{t_{k+1}}^{(s)} = I_{t_k}^{(s)} + b_k^{(s)} \cdot (t_{k+1} - t_k) + \sigma_{t_k}\sqrt{t_{k+1} - t_k} \cdot \xi_k^{(s)}$
    \EndFor
    \State Accumulate sample: $S \leftarrow S + I_{t_K}^{(s)}$
\EndFor
\State \Return $\left.\mathbb{E}[I_t[\bc]]\right|_{t=1} \approx \frac{1}{N_{\text{samples}}} S$
\end{algorithmic}
\end{algorithm}

\begin{remark}[Generalization to physical fields on meshes]
  If the random variables describe physical fields defined on the computational mesh (e.g., piecewise polynomial numerical solutions of PDEs), one can first introduce
  maps between the discrete spaces on the meshes $\mathcal{M}_0$ and $\mathcal{M}_i$ based on the push-forward $(\phi_t^{0,i})_{\#}$ and pull-back $(\phi_t^{0,i})^{\#}$ operators induced by $\phi_t^{0,i}$, in order to transport fields from $\mathcal{M}_0$ to $\mathcal{M}_i$ and vice versa. The conditional drift can be defined
  via
  \begin{equation}
    \label{eq:conditional_drift_lddmm}
    u_t^{\text{cond}}(\bx_0, \bx_i) = (\psi_t^{0,i})^{\#}(\bx_i)-(\phi_t^{0,i})_{\#}(\bx_0)+ \sigma_t \dot{W}^Q_t, \quad t\in[0,1],
  \end{equation}
  where 
  $\psi_t^{0,i}:\Omega_i\subset U\rightarrow\Omega_0\subset U$ is defined as the backward transformation $\psi_t^{0,i}:=\phi_{-t}^{i,0}$
(figure~\ref{fig:sketch_ut_cond}, right).
\end{remark}

\section{Inter-patient numerical modeling of aortic blood flow}
\label{sec:numerical_model_aorta}
The conditional LDDMM stochastic interpolant framework introduced in the previous section can be applied to any numerical model to handle variability of computational domains related via bi-Lipschitz homeomorphisms.  To illustrate its capabilities, we apply the proposed  method to data augmentation and geometric uncertainty quantification for the case of aortic hemodynamics. This section introduces the experimental setup, numerical methods, and data sources underpinning the results in the following sections \ref{sec:uq} and \ref{sec:numerical_experiments}. Specifically, we introduce the dataset of aortic geometries (section \ref{sec:dataset_acquisition}) and the application of ResNet-LDDMM for registration (section \ref{sec:mg_resnet_lddmm_registration}). We then detail the mathematical model and solver calibration (section \ref{sec:numerical_model_calibration}), concluding with the strategy for transporting high-quality hexahedral meshes (section \ref{subsec:mesh_transport}).

\subsection{Acquisition of the dataset}
\label{sec:dataset_acquisition}
For the training of the conditioned LDDMM stochastic interpolant generative model, we employ a dataset $\mathcal S = \left\{ \Omega_i \right\}_{i=0,n_S-1}$
consisting of $n_S=1261$ aortic synthetic shapes ($1209$ for training and $52$ for testing)
generated with statistical shape modeling (SSM) from an initial dataset of $228$ patient surfaces.
The patient dataset was acquired with 3D steady-state free-precession (SSFP) magnetic resonance imaging (MRI) (voxel resolution 2 mm $\times$ 2 mm $\times$ 4 mm, resolution used for surface reconstruction 1 mm $\times$ 1 mm $\times$ 2 mm).
The SSM is built on a centerline encoding, describing each geometry by a fixed number of centerline points ($n_{\text{cntrl}}=390$) and the corresponding radii of the largest inscribed spheres with center on those points. Hence, the encoding vector has size $390\times 4$.
All shapes are topologically equivalent and represent a common portion of the aorta, containing aortic arch, thoracic aorta, and three additional branches.

For a complete description of the procedure, we refer the reader  to~\cite{goubergrits2022ct, thamsen2021synthetic, thamsen2020unsupervised}.
A reduced version of this dataset has been recently used in ~\cite{romor2025dataassimilationperformedrobust} to validate the ResNet-LDDMM method for shape registration.

\begin{figure}[htp!]
  \includegraphics[width=1\textwidth]{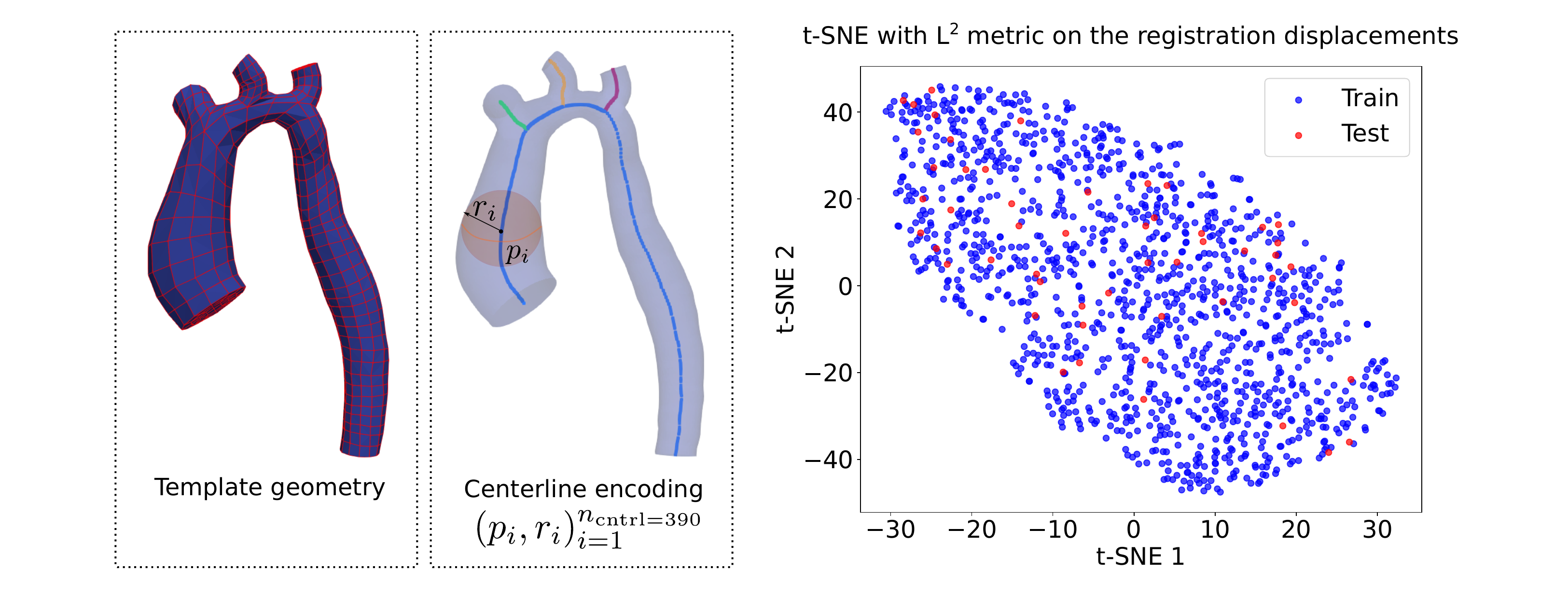}
  \caption{Dataset of aortic geometries. Left: template geometry, and example of centerline encoding with inscribed radii. Right: t-SNE clustering of the dataset based on the similarity matrix of the LDDMM registrations from the template geometry.}
  \label{fig:cluster}
\end{figure}

Figure~\ref{fig:cluster} (left) shows the template geometry selected for the scope of this work (surface mesh).
Figure~\ref{fig:cluster} (right) shows the test and training  shapes in a two-dimensional representation as the result of a clustering with a t-SNE~\cite{2020SciPy-NMeth}
embeddings based on the similarity matrix $D^i = \left(d_{ij}^i\right)$ of the displacements from the template, i.e., $d_{ij}^i := \lVert \phi^{0,i}_1 - \phi^{0,j}_1 \rVert_{L^2}$.

\subsection{Multi-grid ResNet-LDDMM registration of aortic geometries}
\label{sec:mg_resnet_lddmm_registration}
We apply the multi-grid ResNet-LDDMM registration (section~\ref{subsec:resnet_lddmm_chamfer}, see also~\cite{romor2025dataassimilationperformedrobust}),
to compute the registration maps $\phi_i$, $i=1,\hdots,n_S-1$ from the chosen template geometry $\Omega_0$ to each different target geometry $\Omega_i$.
We refer the reader to~\cite{romor2025dataassimilationperformedrobust} for a detailed analysis of the registration accuracy in terms of the Chamfer distance, as well as for a
detailed study on the robustness of the approach with respect to the choice of the template geometry.

\begin{figure}[htp!]
  \includegraphics[width=1\textwidth]{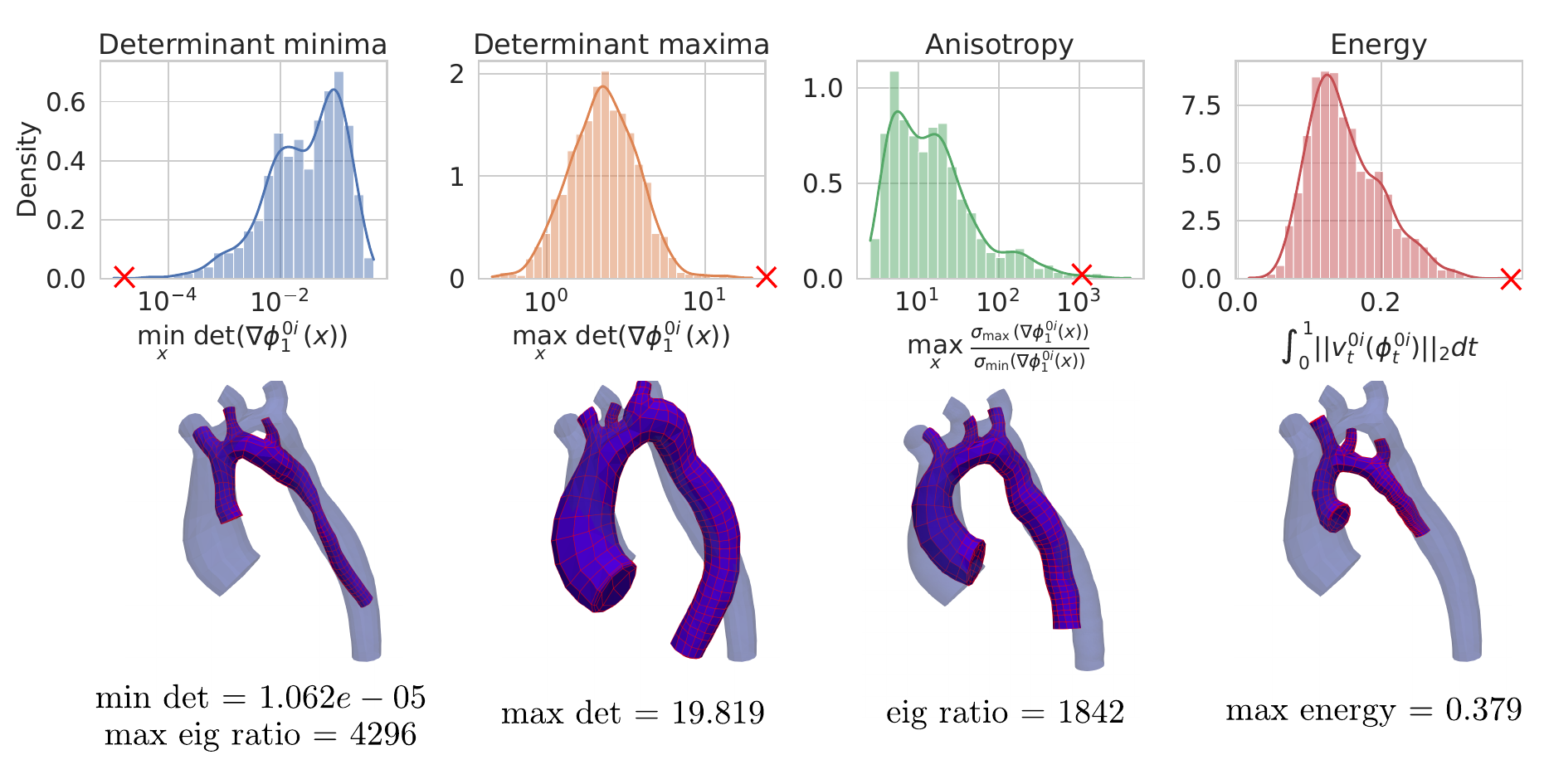}
  \caption{Distributions of selected quantities of interest of the registered dataset over the $1239$ training and test geometries.}
  \label{fig:reg_distributions}
\end{figure}

The multi-grid ResNet-LDDMM registration is performed on tetrahedral surface meshes, considering three levels of refinement during the training, as described  in~\cite{romor2025dataassimilationperformedrobust}. To perform the numerical simulations, the surface registration maps are then extended in the bulk domain, and used
to push forward the hexahedral mesh defined on the template to each target domain (Figure~\ref{fig:mesh_hierarchy_transportation} left).
This is possible because the mappings $\phi^{0,i}_1$ are defined continuously on the ambient space $\mathbb{R}^3$ and are independent of the specific mesh discretization of the domains, except through the optimization problem solved during the LDDMM registration.
However, given the high variability of the shape surfaces, a priori it is not guaranteed that the extension based on interpolation preserves the quality and the consistency of the original mesh
(e.g., leading to degenerating elements). The approach employed for the extension of the registration will be described in section ~\ref{subsec:mesh_transport}.
To illustrate the variability of the dataset, Figure~\ref{fig:reg_distributions} shows the distributions of relevant quantities of interest, such as the minimum and maximum deformation gradient, the ratio between maximum and minimum principal stretches, and the energy of the registration map.

\subsection{Computational hemodynamics}
\label{sec:numerical_model_calibration}
\begin{figure}[htp!]
  \includegraphics[width=0.32\textwidth]{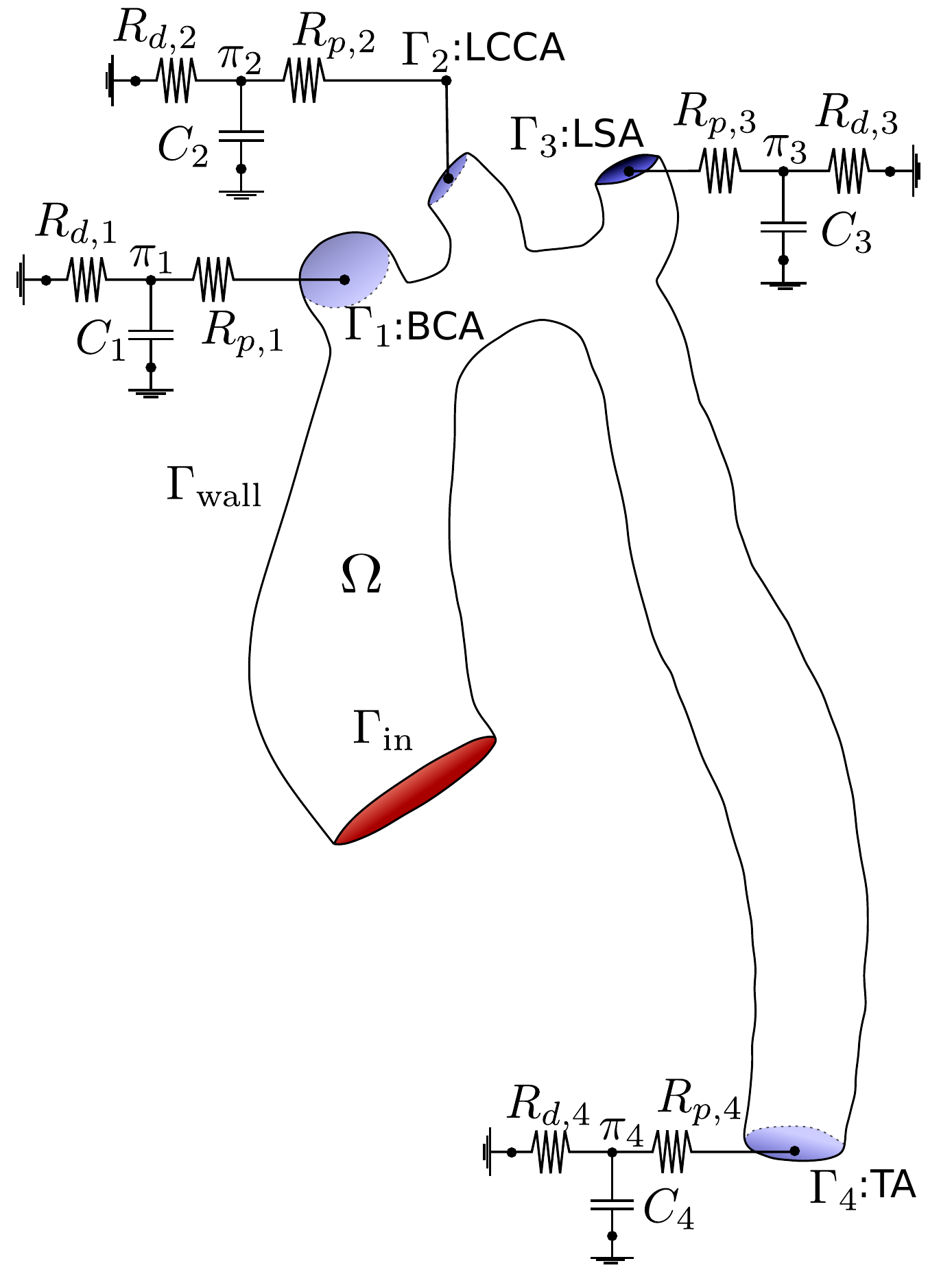}\hspace{1cm}
  \includegraphics[width=0.42\textwidth]{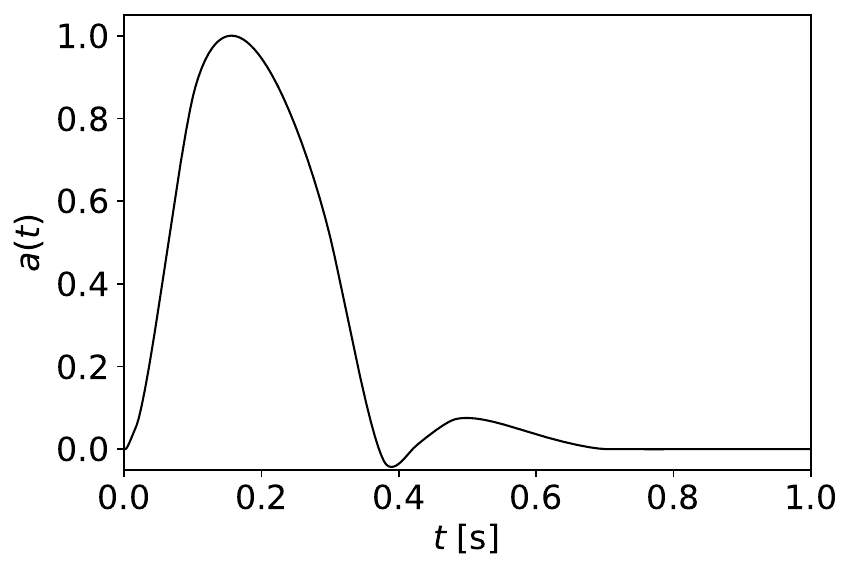}
  \caption{Left: sketch of the computational domain for the computational blood flow model \eqref{eq:3dnse}-\eqref{eq:3dnse-bc}. Right: Time--dependent inflow profile.}
  \label{fig:domain}
\end{figure}

Let us denote by $\Omega \subset \mathbb R^3$ a computational domain whose boundary is defined by a generic shape of the considered dataset.
The boundary of $\Omega$ is decomposed into a vessel wall ($\Gamma_{\rm wall}$), an inlet boundary ($\Gamma_{\rm in}$), and four outlet boundaries (BCA, LCCA, LSA, TA, represented with $\Gamma_i$, $i\in\{1,2,3,4\}$), as shown in Figure~\ref{fig:domain}.

In $\Omega$, we model the blood as an incompressible, Newtonian flow, whose dynamics is described by the velocity $\bu:\Omega \to \mathbb R^3$ and the pressure fields $p:\Omega \to \mathbb R$, solution to the incompressible Navier--Stokes equations
\begin{equation}\label{eq:3dnse}
  \left\{
  \begin{aligned}
    \rho \partial_t \mathbf{u}+\rho  \mathbf{u}\cdot\nabla\mathbf{u}
    -\mu\Delta\mathbf{u}+\nabla p &= \mathbf{0}, && \text{in }\Omega,\\
    \nabla\cdot\mathbf{u} &= 0, && \text{in }\Omega,\\[0.4em]
  \end{aligned}
  \right.
\end{equation}
where $\rho=\SI{1.06e3}{\kilogram\per\meter^3}$ stands for the blood density, and $\mu=\SI{3.5e-3}{\second\cdot\pascal}$ is the dynamic viscosity.
Equations \eqref{eq:3dnse} are completed by Dirichlet boundary conditions on the inlet,
no-slip boundary conditions on the vessel wall, and by a so-called lumped-parameter model on the four outlet boundaries:
\begin{equation}\label{eq:3dnse-bc}
  \left\{
  \begin{aligned}
    \mathbf{u} &= \mathbf{u}_{\rm in}=a(t)\beta(\mathbf{x}), && \text{on }\Gamma_{\rm in},\\
    \mathbf{u} &= \mathbf{0}, && \text{on }\Gamma_{\text{wall}},\\
    \bigl(-pI+\mu(\nabla\mathbf{u}+\nabla\mathbf{u}^T)\bigr)\mathbf{n}
    &= -P_i\mathbf{n}, && \text{on }\Gamma_i,\ \forall i.
  \end{aligned}
  \right.
\end{equation}

The velocity prescribed at inlet is modeled as a time-dependent function $a(t)$ multiplied by a parabolic profile $\beta(\mathbf{x})$.
The outlet pressures $P_i$, $i=1,\hdots,4$, are computed as a function of the boundary flow rates
$Q_i:=\int_{\Gamma_i} \mathbf{u}\cdot\mathbf{n}$, $i \in\{1,2,3,4\}$,  via a three-elements (RCR) Windkessel model~\cite{westerhof2009arterial}:
\begin{equation}\label{eq:wk-rcr-i}
\left\{
\begin{aligned}
C_{d,i}\frac{d\pi_i}{dt}+\frac{\pi_i}{R_{d,i}}=Q_i,\qquad\qquad\qquad\qquad\ \;\,\quad&\text{on}\ \Gamma_i\ \forall i,\\
   P_i=R_{p,i}Q_i+\pi_i,\qquad\qquad\qquad\qquad\ \;\,\quad&\text{on}\ \Gamma_i\ \forall i,\\
\end{aligned}
\right.
\end{equation}

depending on an auxiliary distal pressure $\pi_i$, a proximal resistance $R_{p,i}$, a distal resistance $R_d$, and a capacitance $C_d$.
These parameters model the effect of the neglected cardiovascular system, in terms of the resistance of the arteries in the vicinity of the outlet boundary,
the downstream resistance, and the compliance of the arterial system. The choice of the Windkessel parameters strongly influences the resulting hemodynamics
and their calibration typically depends on the flow regime under consideration, on subject-specific anatomical features, and on the available clinical data.
In this work, we adopt a calibration strategy described in~\cite{romor2025dataassimilationperformedrobust}, which is based on prescribing a target flow split among the outlets, and enables the representation of both physiological and pathological scenarios in a consistent way. We refer to~\cite{romor2025dataassimilationperformedrobust} for the details about the procedure.


The system of equations \eqref{eq:3dnse}-\eqref{eq:3dnse-bc} is discretized in space with
discontinuous mixed-order DG2-DG1 finite elements and in time using a high-order dual splitting scheme (see~\cite{Fehn2021phd,Fehn2020hybrid,Kronbichler2019fast,Krank2017,Kronbichler2021sc,Fehn2018efficiency,Fehn2018robust,Fehn2017stability}
for details on the numerical method and its implementation).
For modeling the effect of turbulence, a $\sigma$-model is used ~\cite{katz2023impact}.
The model is implemented using \texttt{ExaDG}~\cite{ExaDG2020}, which is based on the open-source library \texttt{deal-II}~\cite{arndt2024deal}
and relies on a matrix-free implementation on hierarchical hexahedral meshes.

The hierarchical structure of meshes is necessary for applying the
geometric multigrid preconditioners which are critical for the efficiency of the methods, and must be preserved
when registering the template meshes onto different shapes. This issue will be discussed in the following section \ref{subsec:mesh_transport}.

\subsection{Preservation of hexahedral mesh hierarchies structure upon registration}
\label{subsec:mesh_transport}
The registration maps between aortic shapes (section~\ref{sec:mg_resnet_lddmm_registration}) are computed to match only the surface meshes, since considering a full three--dimensional registration would considerably increase the computational burden.

A more efficient alternative consists then in properly defining extensions of the surface maps into the interior domain and transporting the mesh accordingly, using, e.g., radial basis function interpolation, or solving an auxiliary problem. Additionally, smoothing steps might be required in the presence of large deformations to improve mesh quality, avoid the degeneration of mesh cells, and preserve mesh regularity properties which might be required or affect the stability of the numerical method.

The considered numerical solver, implemented in \texttt{ExaDG}~\cite{ExaDG2020}, employs geometric multigrid preconditioners on nested hexahedral meshes. Hence, the hierarchical structure needs to be preserved when extending the surface displacement map to the mesh vertices in the volume domain.

The extension algorithm considered in this work consists of an iterative algorithm solving a non-homogeneous linear elastic problem and a smoothing step.
Namely, let $\mathcal M_0$ and $\mathcal{M}_i$ be two different meshes, the template mesh and the target mesh generated with the conditional LDDMM stochastic interpolant algorithm~\ref{alg:lddmm_fm_generation} respectively, and let  us denote with $\Psi_{\text{s}}\defeq \left.\mathbb{E}[I_t[\bc]]\right|_{t=1}$ the displacement of the surface mesh nodes from $\mathcal M_0$ to $\mathcal{M}_i$.
To handle large deformations, the surface displacement $\Psi_{\text{s}}$ is split in $N_{\text{steps}}$ iterations. Setting $\Psi_{0}=I_d$,
for $n=1, \ldots, N_{\text{steps}}$, we solve a fictitious linear elastic problem for the intermediate displacement field.
 \begin{equation}
  \label{eq:intermediate elastic problem-psi}
  \left\{
  \begin{aligned}
  \nabla \cdot \left[ \frac{E(\mathbf{x})}{2(1+\nu)} \left( \nabla \Psi_{n} + (\nabla \Psi_{n})^T \right) + \frac{E(\mathbf{x})\nu}{(1+\nu)(1-2\nu)} (\nabla \cdot \Psi_{n}) I \right] &= \mathbf{0}, && \text{in }\Psi_{n-1}(\Omega_0),\\
  \Psi_{n}|_{\partial \Omega} &= \tfrac{e^{-\tfrac{n}{N_{\text{steps}}}}-1}{e^{-1}-1}\ \Psi_{\text{s}}, && \text{on }\partial\Psi_{n-1}(\Omega_0),
  \end{aligned}
  \right.
\end{equation}
where the non-homogeneous Young modulus  $E(\mathbf{x})$ is defined as the square of the finite element cell aspect ratio and
the Poisson's ratio $\nu=0.3$ is fixed.
At the end of each step, inverted cells are detected and iteratively corrected adjusting the value of the displacement field $\Psi_n$ at the nodes.

The surface solution obtained after $N_{\text{steps}}$ is then processed by a smoothing algorithm that defines internal displacement limiting the number of cells with
bad aspect ratio (details of the smoothing step are provided in appendix~\ref{sec:mesh_smoothing}).

\begin{figure}[htp!]
  \centering
  \includegraphics[width=1\textwidth]{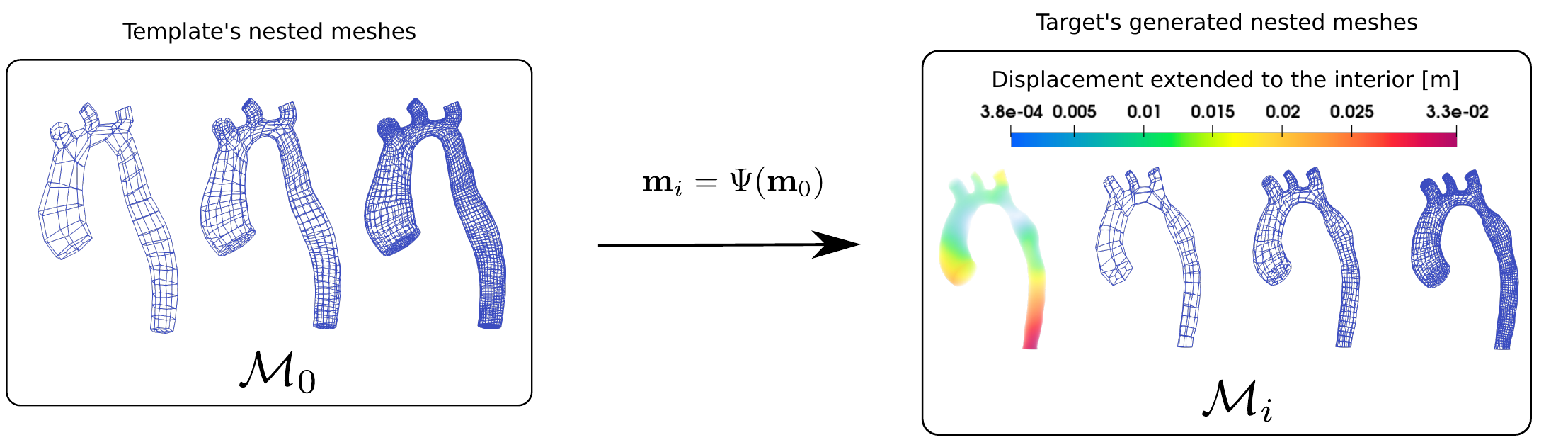}
  \caption{Transport of nested hexahedral mesh hierarchies needed for geometric multigrid preconditioners.}
  \label{fig:mesh_hierarchy_transportation}
\end{figure}

Figure~\ref{fig:mesh_hierarchy_transportation} shows an example of the nested hexahedral mesh hierarchy transported from $\mathcal{M}_0$ to the target generated mesh $\mathcal{M}_i$, associated to a particular realization $\mathbf{c}_i\in\mathbb{R}^c$ of the conditioning random variable (section \ref{subsec:generative}).

\begin{remark}[Generation of the template mesh]
  The described mesh extension procedure has been used to define also the nested hexahedral mesh $\mathcal{M}_0$ on the template geometry, transporting it from another mesh~\cite{Bonjak2025}, not contained in our original dataset.
\end{remark}

\section{Application to shape uncertainty quantification in cardiovascular models}
\label{sec:uq}
Studying the sensitivity of quantities of interest with respect to domain uncertainties requires the generation of random domain perturbations. In the case of anatomical shapes, which are not described by a small number of parameters, this step needs a suitable and robust procedure that preserves the physiological features of the anatomy and fits into the considered numerical scheme in terms, e.g., of mesh topology and mesh quality.

\subsection{Numerical methods for geometric uncertainty quantification}
\label{subsec:generative}

We apply the conditional LDDMM stochastic interpolant method to generate domain perturbations that will be used to study the impact of geometrical uncertainty.
Namely, shapes are generated through deformations  $\Psi:\hat{\Omega} \rightarrow\Omega$ 
of a reference domain $\hat{\Omega} \subset \mathbb R^3$, obtained from LDDMM flows (as described in section~\ref{sec:lddmflow}),
by varying a conditioning variable $\bc$:
\begin{equation}\label{eq:c-to-Psi}
\bc\in\mathbb{R}^c \mapsto  \{b_t^{\theta}[\bc]\}_{t\in[0,1]} \mapsto  \Psi^{\bc} \coloneqq \mathbb{E}[\left.I_t\right|_{t=1}[\bc]],
\end{equation}
where $I_t$ is obtained from the SDE
\begin{equation}\label{eq:lddmm-sde}
  \left\{\begin{aligned}
  dI_t[\bc] & = b^{\theta}_t[\bc](I_t) dt + \sigma_t dW^Q_t, \\
  \quad I_{t=0} & = \text{Id}\,.
  \end{aligned}
    \right.
\end{equation}
depending on the drift $b^{\theta}[\bc]$. For ease of notation, we also denote by $\Psi^{\bc}$ the LDDMM displacement field generated by the
conditional stochastic interpolant \eqref{eq:c-to-Psi}-\eqref{eq:lddmm-sde}, extended in the interior domain as described in subsection~\ref{subsec:mesh_transport}.

\begin{remark}[Regularity of the perturbation]
The spatial regularity of the domain perturbation $\mathbf{x} \mapsto \Psi^{\mathbf{c}}(\mathbf{x})$ and the parameter-to-perturbation map $\mathbf{c} \mapsto \Psi^{\mathbf{c}}$ are governed by the regularity of the vector field $b$ with respect to $\mathbf{x}$ and the parameter $\theta$, respectively. When $b^{\theta}$ and the map $\mathbf{c} \mapsto \{b_t^{\theta}[\mathbf{c}]\}_{t \in [0, 1]}$ are discretized via neural networks, their regularity is inherited from the choice of activation functions. Specifically, while ReLU activations provide Lipschitz continuity (often associated with $H^s$ for $s < 3/2$ in this context), tanh or sigmoid activations yield analytic (Gevrey-1) regularity. Furthermore, the use of Gevrey bump functions allows for the recovery of Gevrey-$\sigma$ regularity with $\sigma > 1$. Several studies have been dedicated to analyzing the regularity of parameter-to-solution maps: analytic regularity (see, e.g.,~\cite{Henrquez2021, castrillon2016analytic}), $(b,\epsilon,p)$-holomorphic maps~\cite{dölz2025fullydiscreteanalysisgalerkin}, in particular for the stationary incompressible Navier-Stokes equations~\cite{Cohen2018}, and the Gevray regularity~\cite{chernov2024analytic, harbrecht2024gevrey, djurdjevac2025uncertaintyquantificationstationarytimedependent}.
\end{remark}

The vanilla Monte Carlo method will be used to compute estimators of the expectation of a selected quantity of interest 
$Q\left(\mathbf{u}(\Omega),p(\Omega)\right)$, defined as a function of the Navier--Stokes equations on a perturbed domain $\Omega$, i.e.,
\begin{equation*}
E[Q] \approx \frac{1}{M} \sum_{i=1}^M Q(\mathbf u(\Omega_i),p(\Omega_i)), 
\end{equation*}
where $M$ is the number of samples, $\Omega_i := \Psi^{\bc_i}(\hat{\Omega})$, and $\bc_i \sim \Lambda_P(C)$, the probability distribution of the conditioning variable $C$, as defined in subsection~\ref{subsec:lddmm_flow_matching}. 

\begin{remark}
Alternative numerical strategies to compute statistics of quantities of interest include variants of the Monte Carlo methods (e.g., quasi-Monte Carlo, multilevel Monte Carlo), 
polynomial chaos expansions, realized through stochastic collocation~\cite{nobile2008sparse} (tensor quadrature rules, sparse quadrature rules, etc.), and stochastic Galerkin methods~\cite{Babuska2004}.
\end{remark}

\begin{remark}
For the sake of completeness, it is worth mentioning that in the context of uncertainty quantification, 
the most common numerical method to induce random domain perturbations is the truncated 
Karhunen-Loève expansion of Gaussian random fields (see, e.g., ~\cite{bovsnjak2025geometric}) defined on the reference domain $\hat{\Omega}$ 
with a prescribed covariance kernel (e.g., RBF kernel, Matérn kernel, etc.):
\begin{equation}
\Psi(\hat{\bx}; \bc) = \hat{\bx} + \sum_{i=1}^K \sqrt{\lambda_i} \psi_i(\hat{\bx}) c_i, \quad \bc = (c_1, \ldots, c_K) \sim \mathcal{N}(0, I),
\end{equation}
where $\{(\lambda_i, \psi_i)\}_{i=1}^K$ are the first $K$ eigenpairs of the covariance operator associated with the chosen kernel.
In this case, the regularity of the bases $\{\psi_i\}_{i=1}^K$ of the expansion determines the regularity of the perturbations. Other techniques include the randomization of parametrized geometrical descriptions such as free-form deformation~\cite{manzoni2012shape}, radial basis functions, Fourier and wavelet representations, or ad-hoc parametrized deformations~\cite{Bonjak2025}.
\end{remark}

\subsection{Domain perturbations for aortic geometries}
\label{sec:generation_perturbations}
In the case of aortic shapes under study, the conditioning variable $\mathbf{c}$ is defined as a low-dimensional representation of the aortic shapes, composed of
the coordinates of $94$ centerline control points and the associated inscribed radii, i.e., 
$\bc = (\mathbf{p}, r) \in\mathbb{R}^{94\times 4}$.
The time-dependent conditioned vector field $b^{\theta}_t[\bc]:\mathbb{R}^3\rightarrow\mathbb{R}^3$ is approximated using
a graph neural network made of consecutive MeshGraphNet layers~\cite{pfaff2020learning}.
The training procedure follows algorithm~\ref{alg:lddmm_fm_training} and is described in more detail in appendix~\ref{sec:training_details}. 
Notice that, during training, the original LDDMM stochastic interpolant loss (equation~\eqref{eq:lddmm_flow_matching_loss}) is approximated with mini-batches of size $n_{\text{b}}$ randomly sampled from the $1209$ training shapes:
\begin{equation*}
\mathcal{L}_{\text{batch}}(\theta) = \frac{1}{n_{\text{b}}} \sum_{i=1}^{n_{\text{b}}} \mathbb{E}_{t, \epsilon}\left[\|b^{\theta}_t(I_t[\bc_i], \bc_i) - v_t^{0i}(\phi_t^{0i}(\mathbf{m}_0)) - \sigma_t \epsilon\|^2_2\right],
\end{equation*}
where $\bc_i =  (\mathbf{p}_i, r_i)$ is the conditioning variable associated with the $i$-th training shape. 
For each batch, the functions are evaluated on the same set of points $\mathbf{m}_0$ (corresponding to the nodes of the hexahedral template mesh), and then mapped
onto the deformed geometries via the corresponding registration map. Since our aim is to generate perturbations of given aortic shapes, at the end of training, we fine-tune the model on the $30$ test shapes for a few epochs.
Once the model is trained and fine-tuned, we can generate perturbations of a given test shape associated with $\bc$ by simulating the SDE 
\eqref{eq:lddmm-sde} and setting the generated perturbation as $\Psi = \mathbb{E}[\left.I_t\right|_{t=1}[\bc]]$.

\begin{figure}[htp!]
  \centering
  \includegraphics[width=1\textwidth]{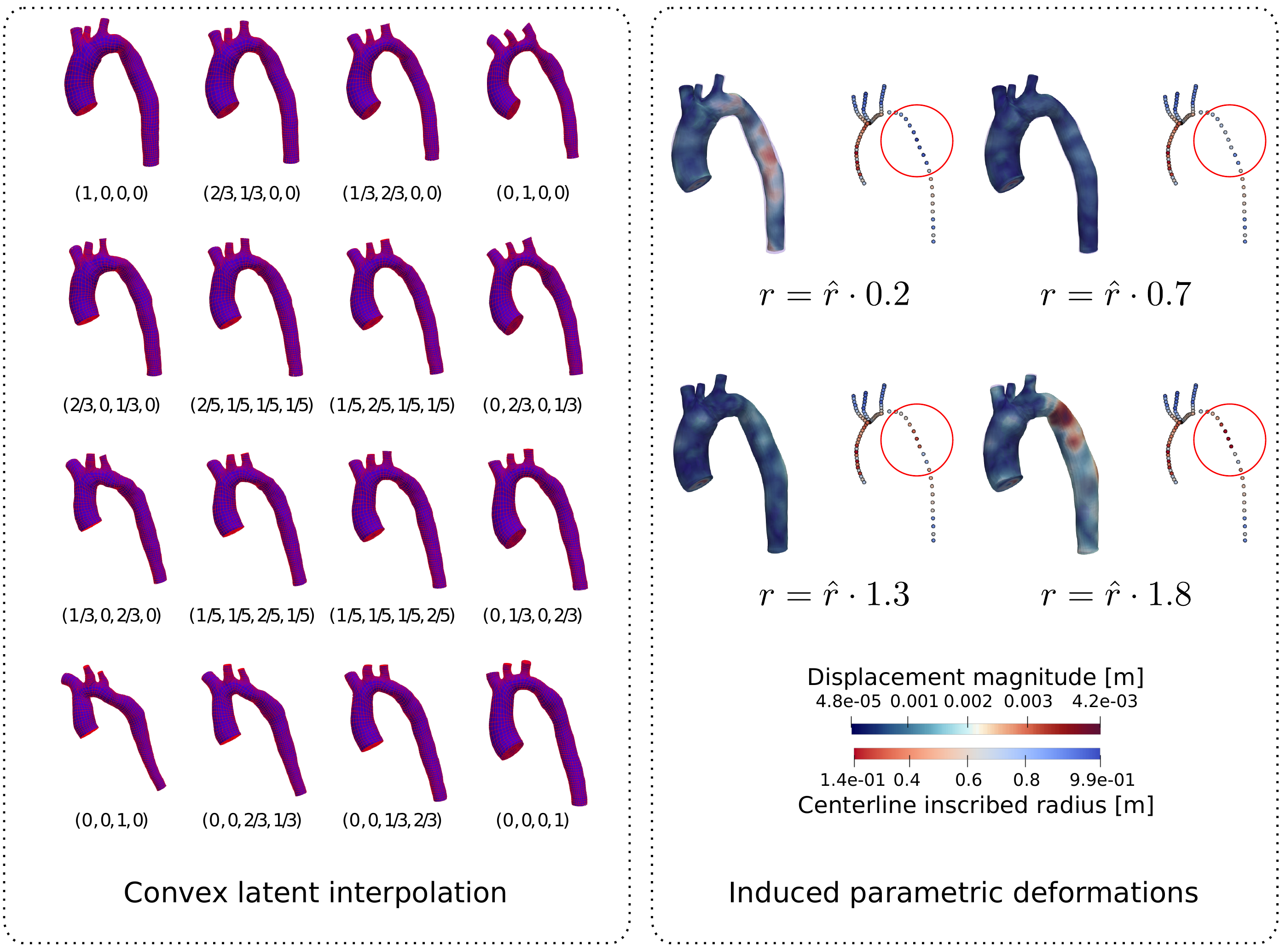}
  \caption{Left: Example of shape generated from a convex interpolation on the latent space of four test geometries. 
  Right: Limitations of the method with respect to local parametric deformations. We impose a variation of the inscribed radii
  by a factor of $\alpha_r \in \{0.2, 0.7, 1.3, 1.8\}$, limited to the centerline points close to a selected location (red circles). This results in perturbed shapes with
  global radius variations.}
  \label{fig:interpolation}
\end{figure}

In particular, Figure~\ref{fig:interpolation}, left, shows an example of shapes generated from a convex latent interpolation.
In this case, we have chosen $4$ test geometries, with corresponding latent coordinates $\bc_1, \ldots, \bc_4$,  
and considered the conditioning variable associated with the convex combinations 
\[
\bc_{\text{interp}} = \sum_{i=1}^4 \alpha_i \bc_i, \;\mbox{with}\; \alpha_i\in[0,1], \;\sum_{i=1}^4\alpha_i=1.
\]
We can observe that the interpolated shapes represent smooth transitions between the four chosen geometries, and that 
the intermediate conditioning variables yield realistic aortic shapes, which were not contained in the original dataset.

\begin{figure}[htp!]
  \includegraphics[width=0.8\textwidth]{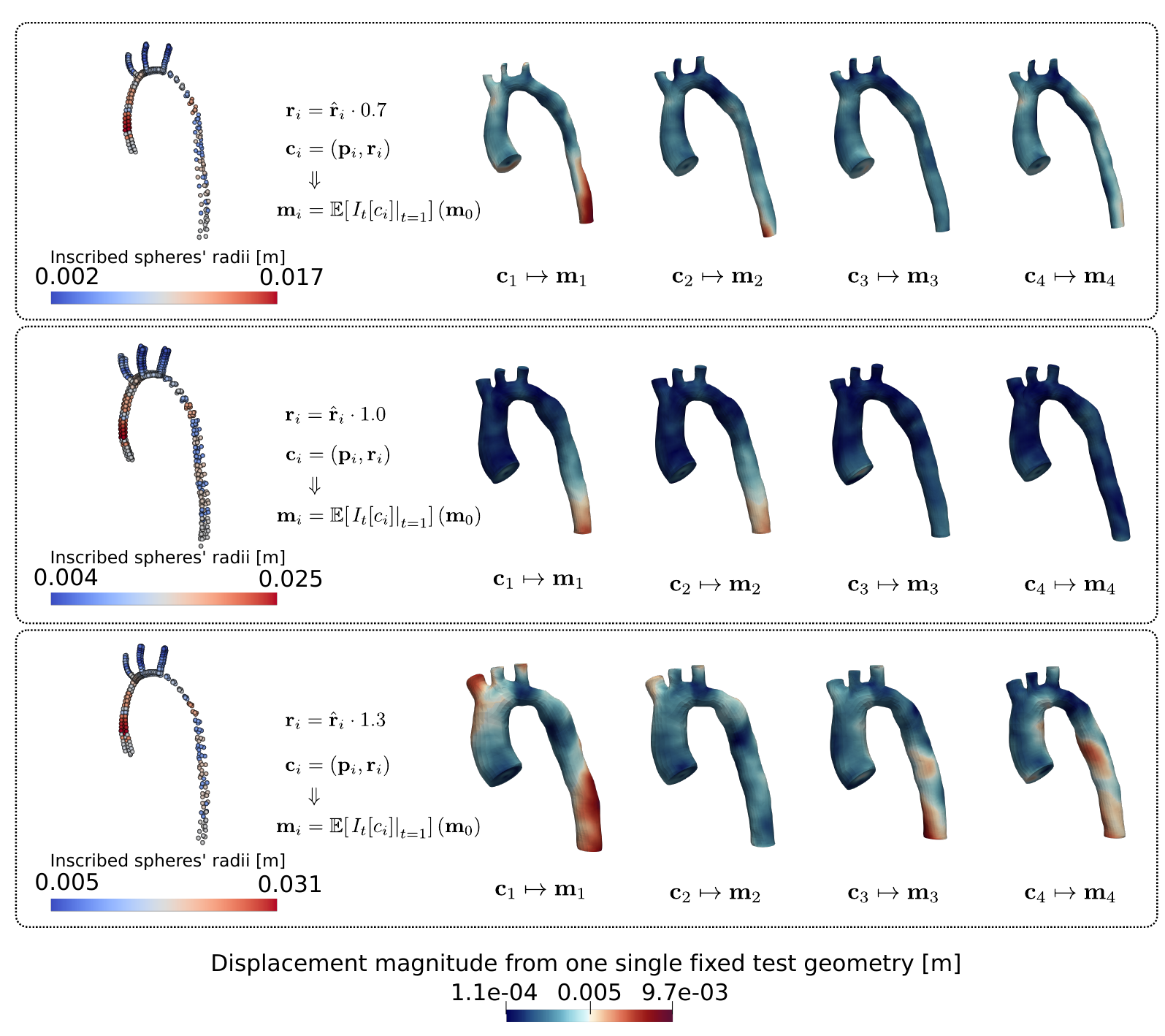}
  \caption{Shape generation with conditional LDDMM stochastic interpolant.
The three different rows correspond to a different factor applied to the reference inscribed radii: $0.7$, $1.0$, and $1.3$; in each row, we show the four generated samples, varying only the Gaussian random field employed to perturb the control points coordinates, reported on the left. The displacement magnitude of the associated registration maps from the template to the generated test shape is also shown.
}
  \label{fig:sampling}
\end{figure}

Additionally, it is possible to increase the variability of the generated shapes by varying uniformly the radii of inscribed spheres and perturbing  
the control points' coordinates with a Gaussian random field with RBF kernel.  Figure~\ref{fig:sampling} shows some examples of shapes generated with this approach.
This strategy will be used in section~\ref{sec:shape_uq_numerical_experiments}
for the shape uncertainty quantification study.

The proposed stochastic interpolant LDDMM allows one to generate domain perturbations of aortic shapes based on their latent representations $\bc$. 
On the other hand, local deformations, such as enlargements or contractions at specific locations, which are relevant
for certain applications (e.g., presence of aneurysms or stenoses), might be difficult to achieve. 
To illustrate this issue, we selected a test geometry and generated perturbed shapes by varying locally the latent coordinates corresponding to the inscribed radii close to 
a given location by a factor of $\alpha_r \in \{0.2, 0.7, 1.3, 1.8\}$. The results of this experiment (Figure~\ref{fig:interpolation}, right) show that the radii were globally 
increased or decreased with respect to the original radii. This behavior can be explained by the fact that the training dataset does not contain shapes with local variations of the inscribed radii, and thus the model is not able to learn such local deformations. Possible strategies to overcome this issue include augmenting the training dataset with shapes with local variations of the inscribed radii.

\begin{remark}
No metric constraints were imposed during the training of the generative model. Hence, changing the radius of the inscribed spheres by a specific factor does not guarantee that the generated shape will reflect this change accurately. 
This can be seen in Figure~\ref{fig:sampling}, which shows that the value of the inscribed spheres' radii and the associated displacement magnitude are of the same order of magnitude but not exactly the same. Adjustments and enhancements of the method to improve the accuracy of these parametric deformations will be the subject of future research.
\end{remark}

\section{Numerical results}
\label{sec:numerical_experiments}
\newcommand{\legendnote}{Black: $\alpha_r = 1.3$ (30\% radius increase), blue: $\alpha_r = 1$, red:  $\alpha_r = 0.7$ (30\% radius decrease). }
\newcommand{\compnote}{Black: larger batch of $100$ simulations; red: original smaller batch. }
\newcommand{\legendnoteline}{Solid lines represent means, dotted lines represent standard deviations. }

\subsection{Simulation setup}
\label{sec:simulation_setup}

\begin{figure}[!ht]
  \includegraphics[width=0.7\textwidth]{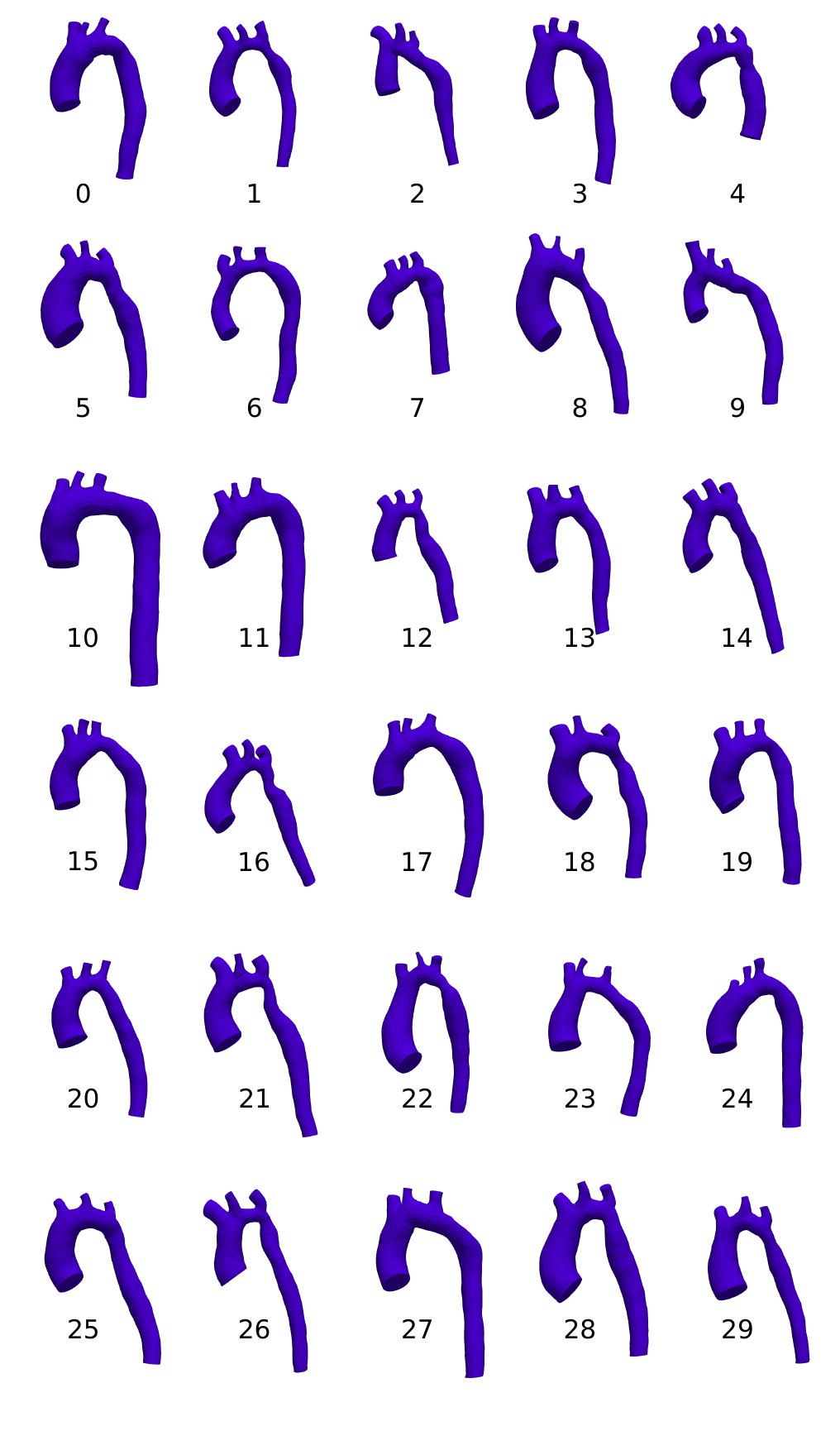}
  \caption{The 30 test geometries used to generate the different batches.}
  \label{fig:all_test_geometries}
\end{figure}

\begin{figure}[htp!]
\begin{center}
  \includegraphics[width=0.11\textwidth]{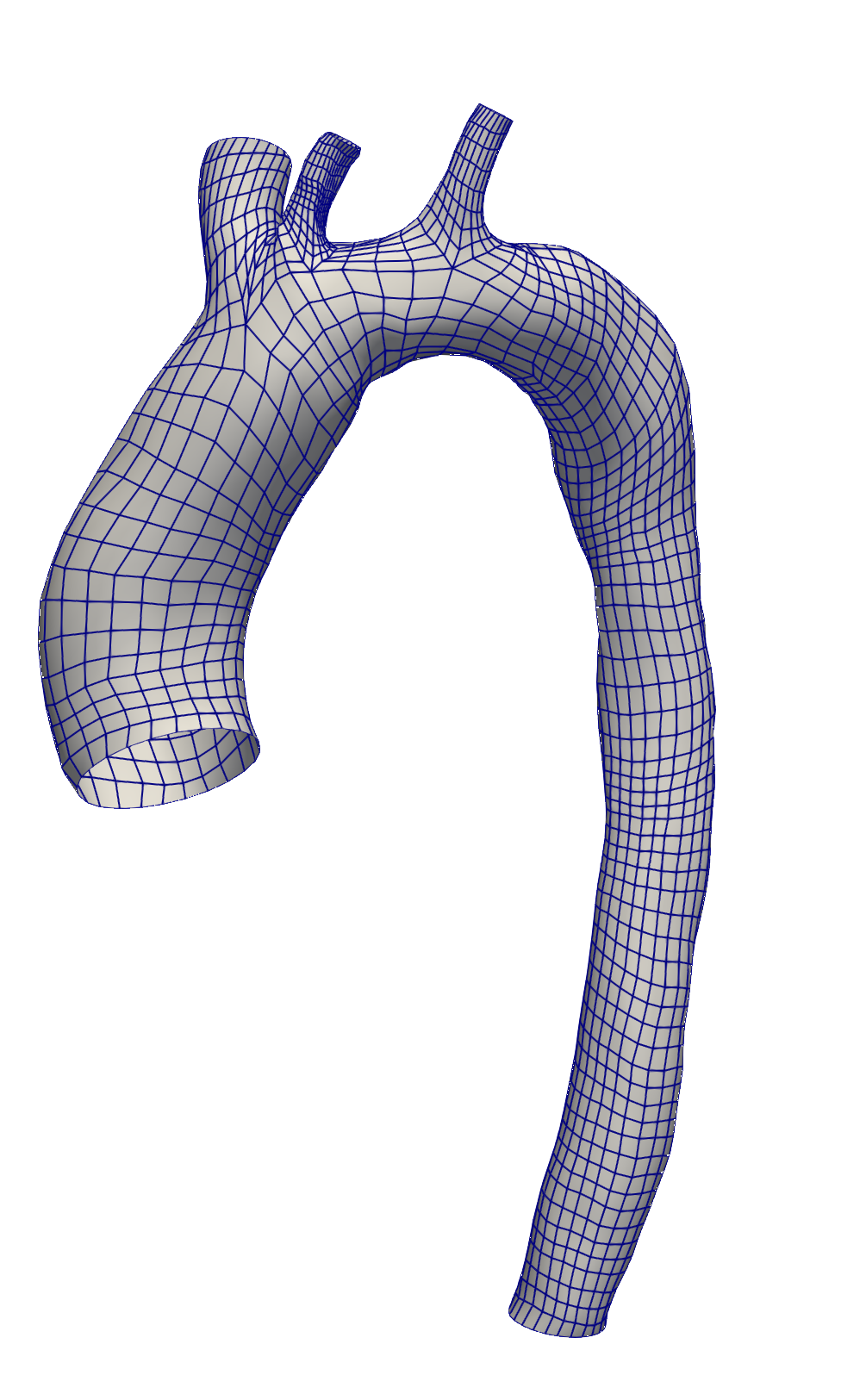}\hspace{.5cm}
  \includegraphics[width=0.11\textwidth]{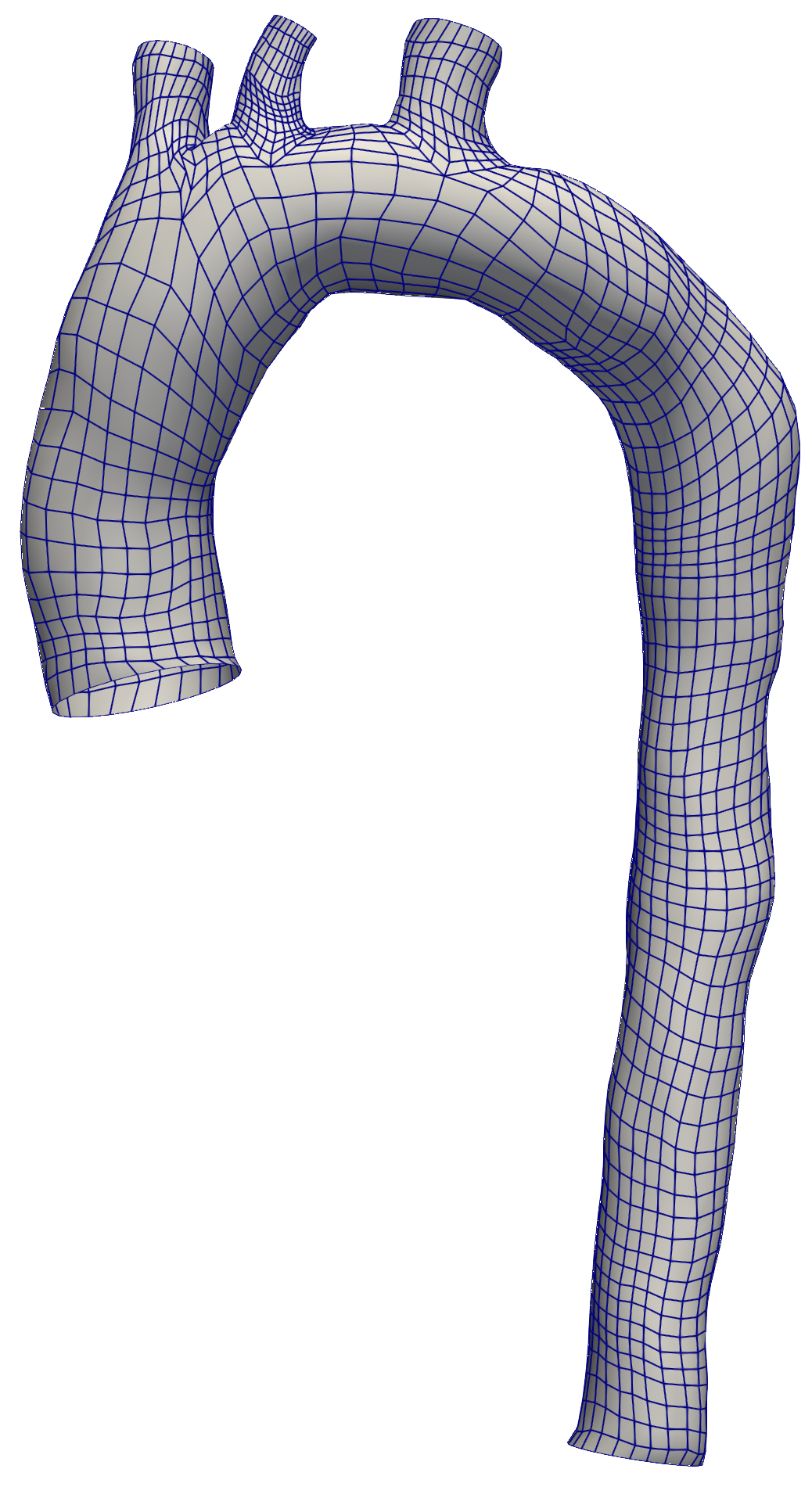}\hspace{.5cm}
  \includegraphics[width=0.11\textwidth]{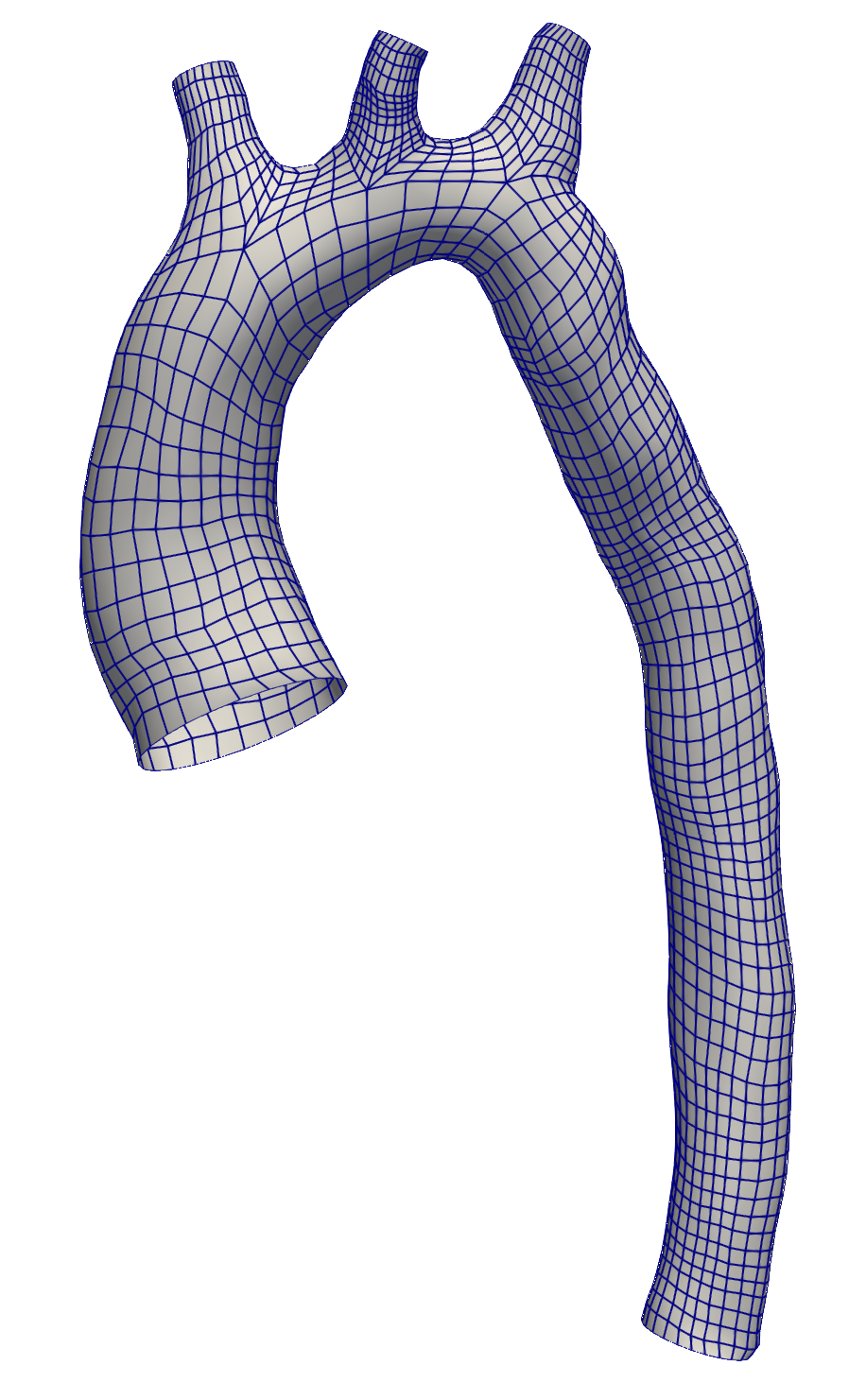}\hspace{.5cm}
  \includegraphics[width=0.11\textwidth]{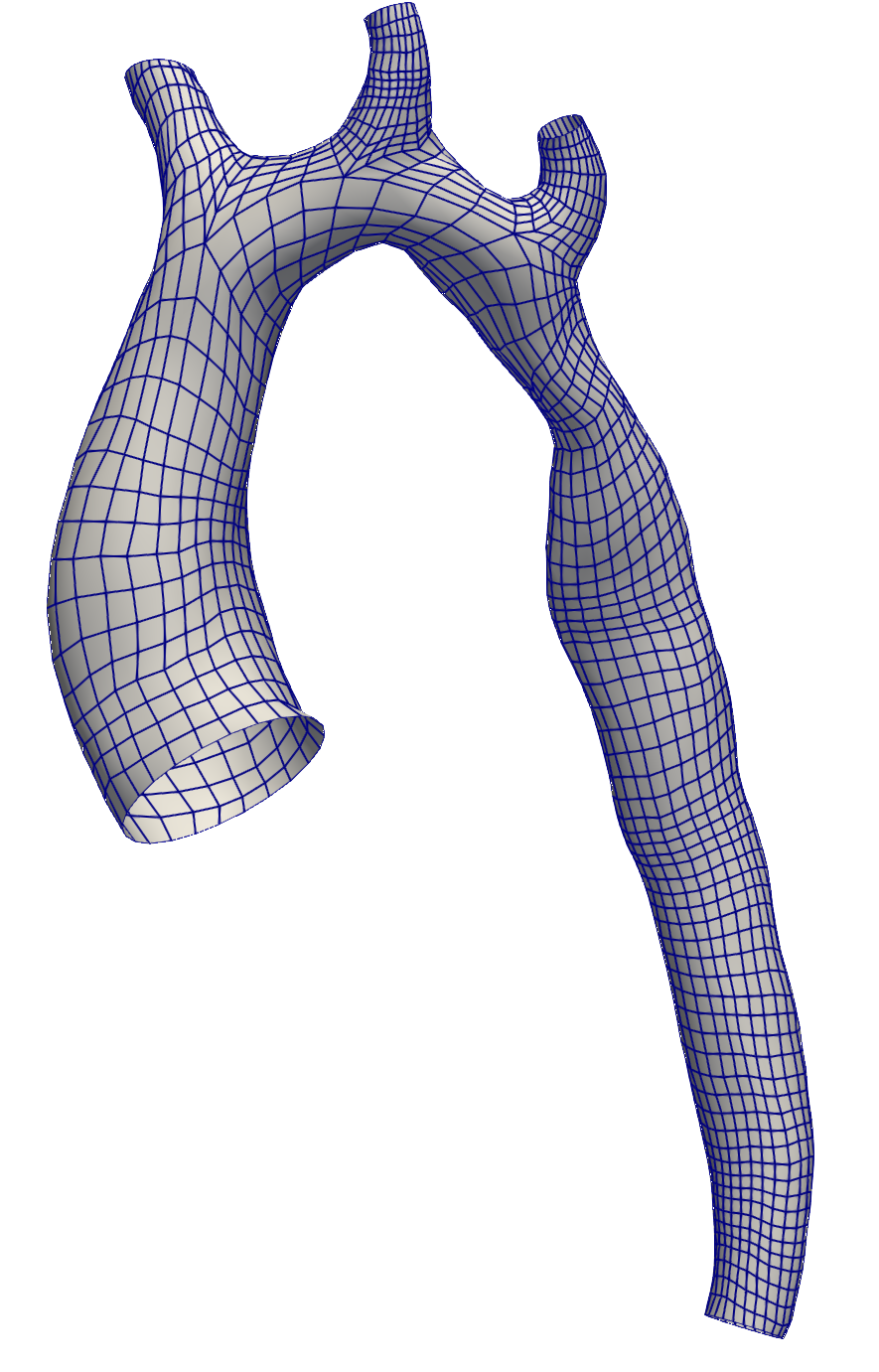}\\
  \includegraphics[width=0.11\textwidth]{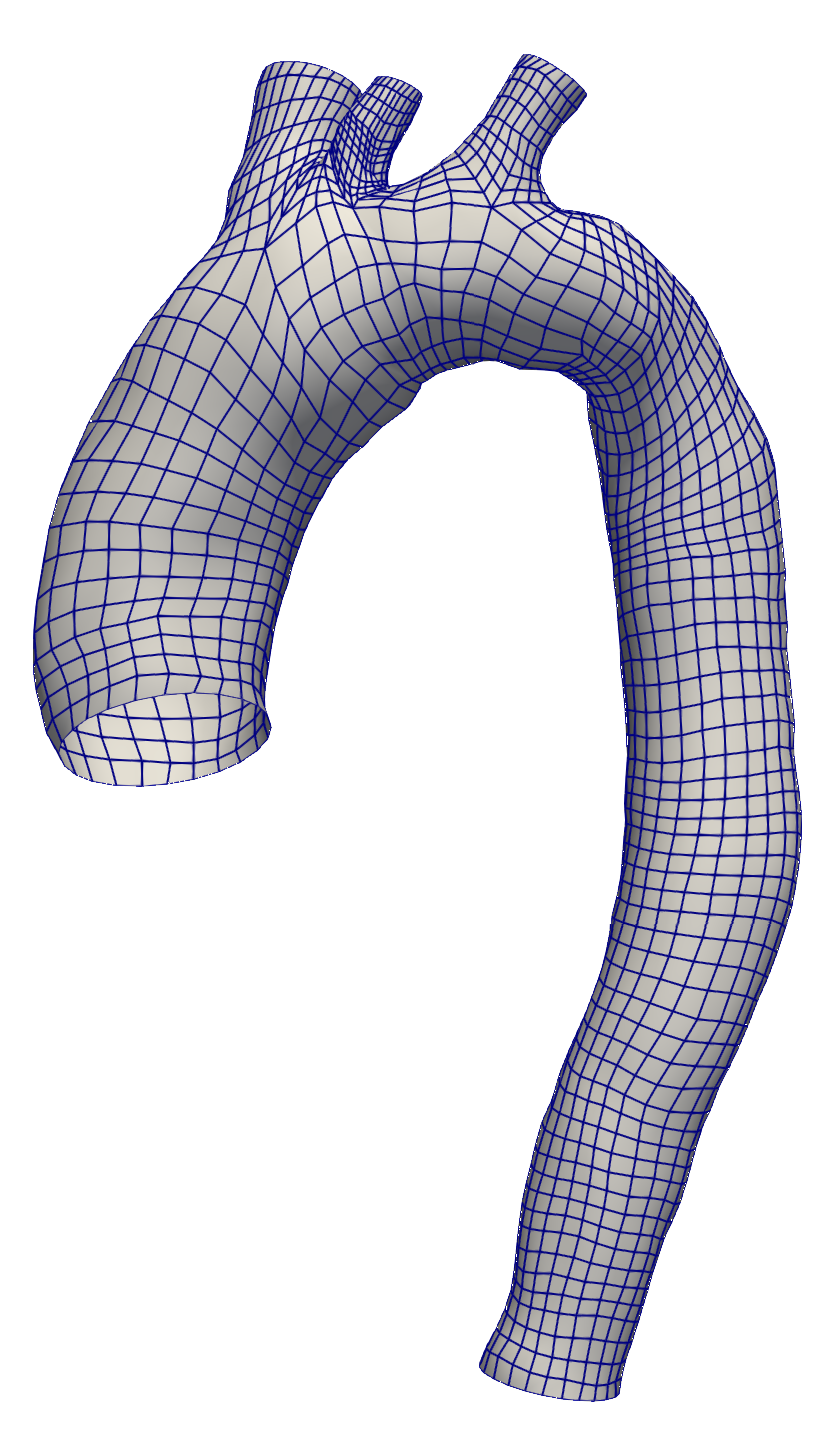}\hspace{.5cm}
  \includegraphics[width=0.11\textwidth]{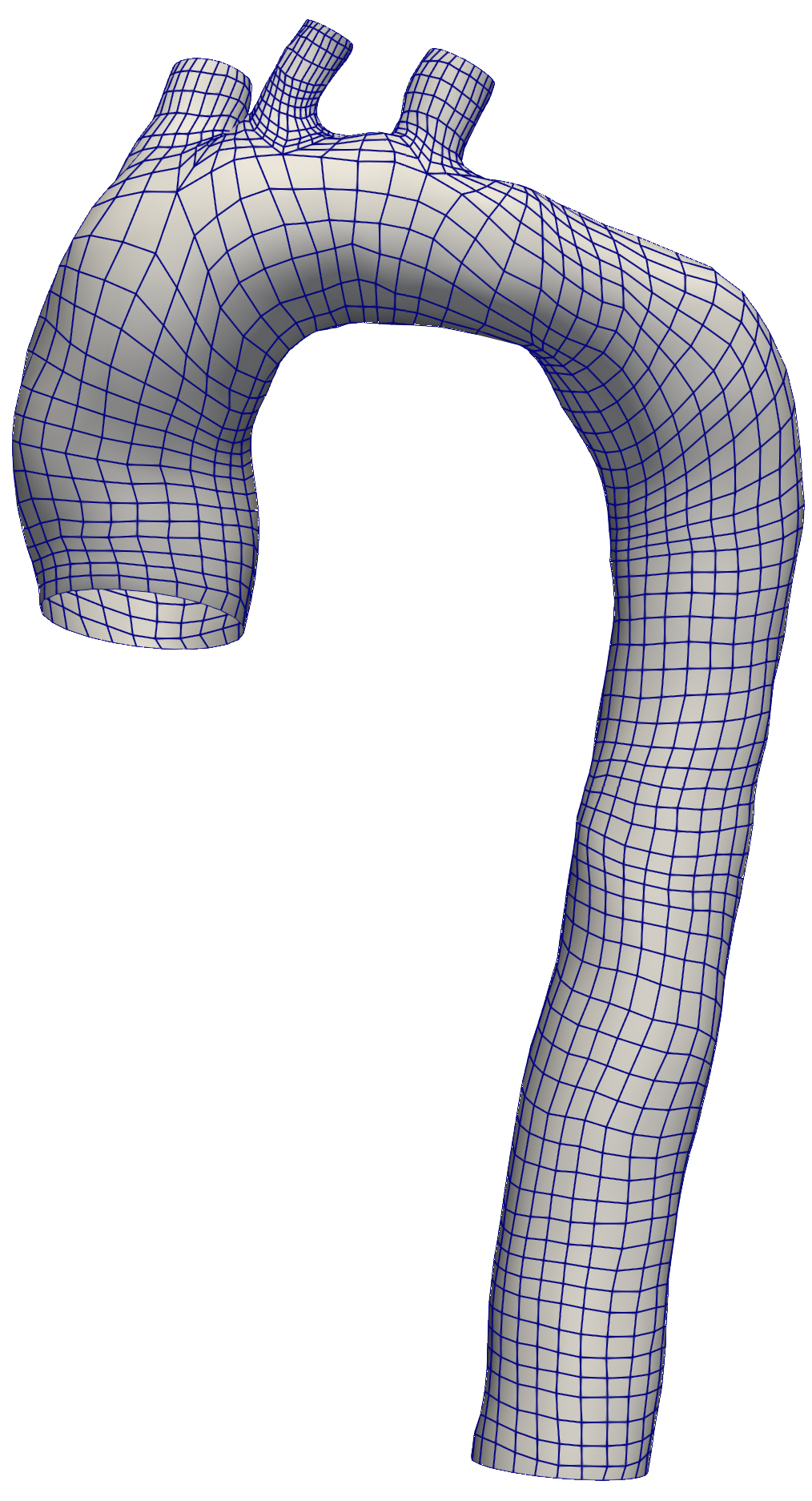}\hspace{.5cm}
  \includegraphics[width=0.11\textwidth]{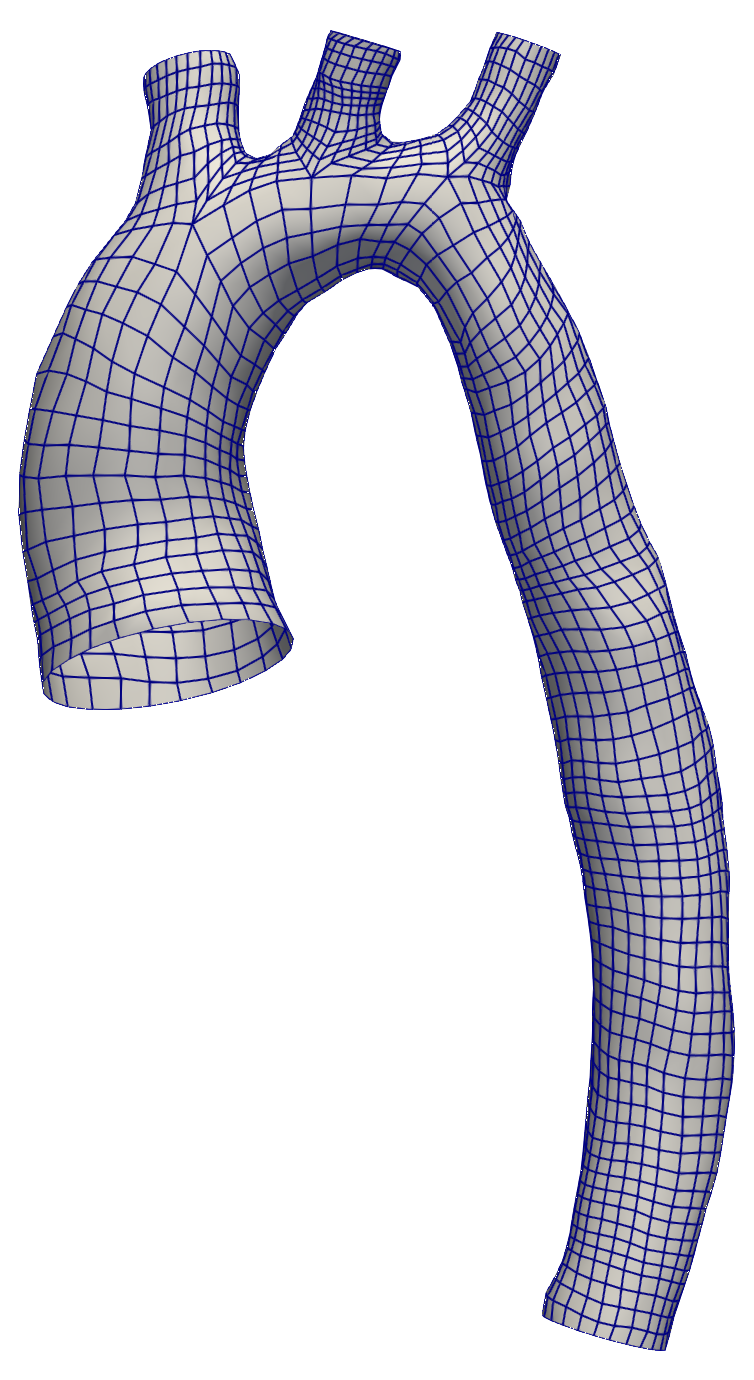}\hspace{.5cm}
  \includegraphics[width=0.11\textwidth]{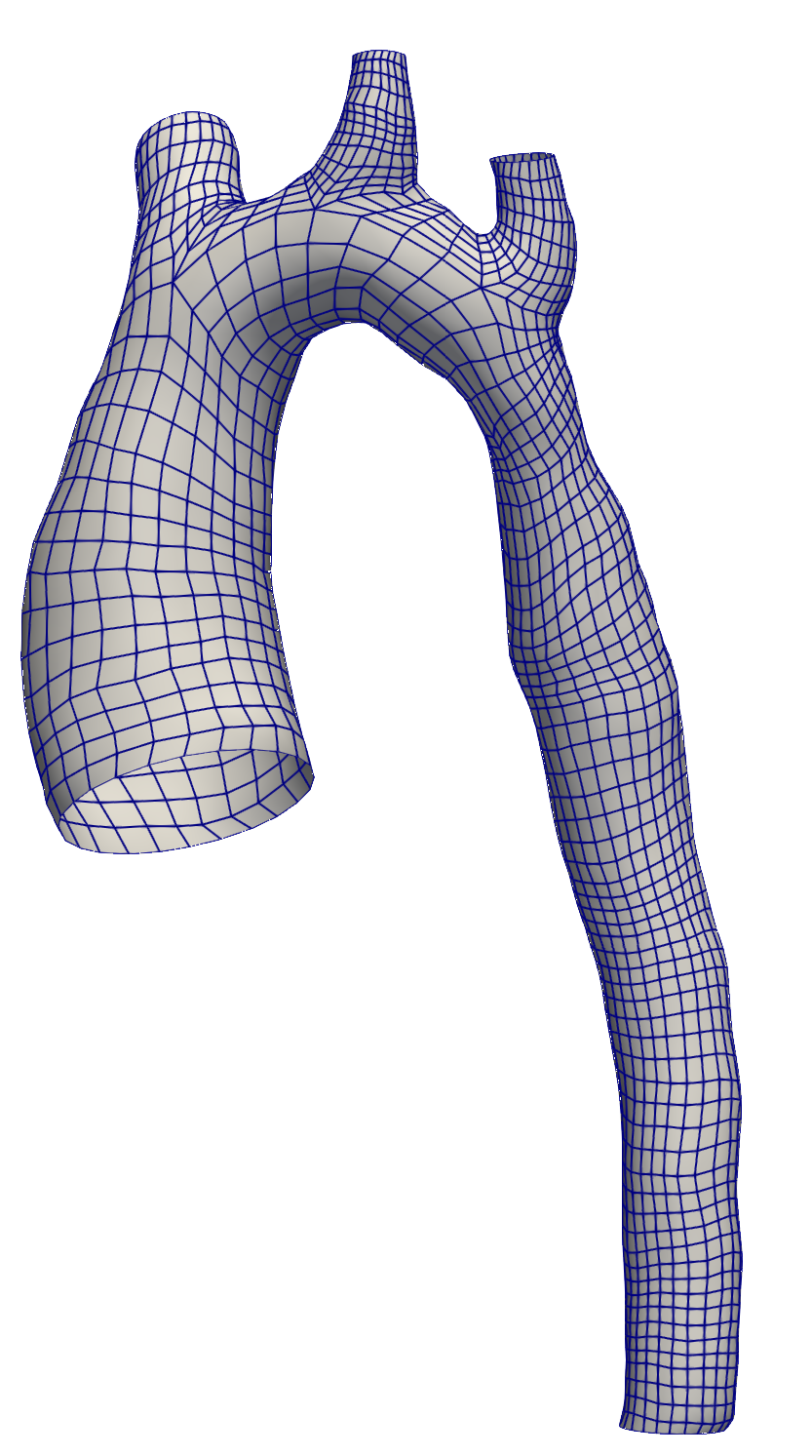}\\
  \includegraphics[width=0.11\textwidth]{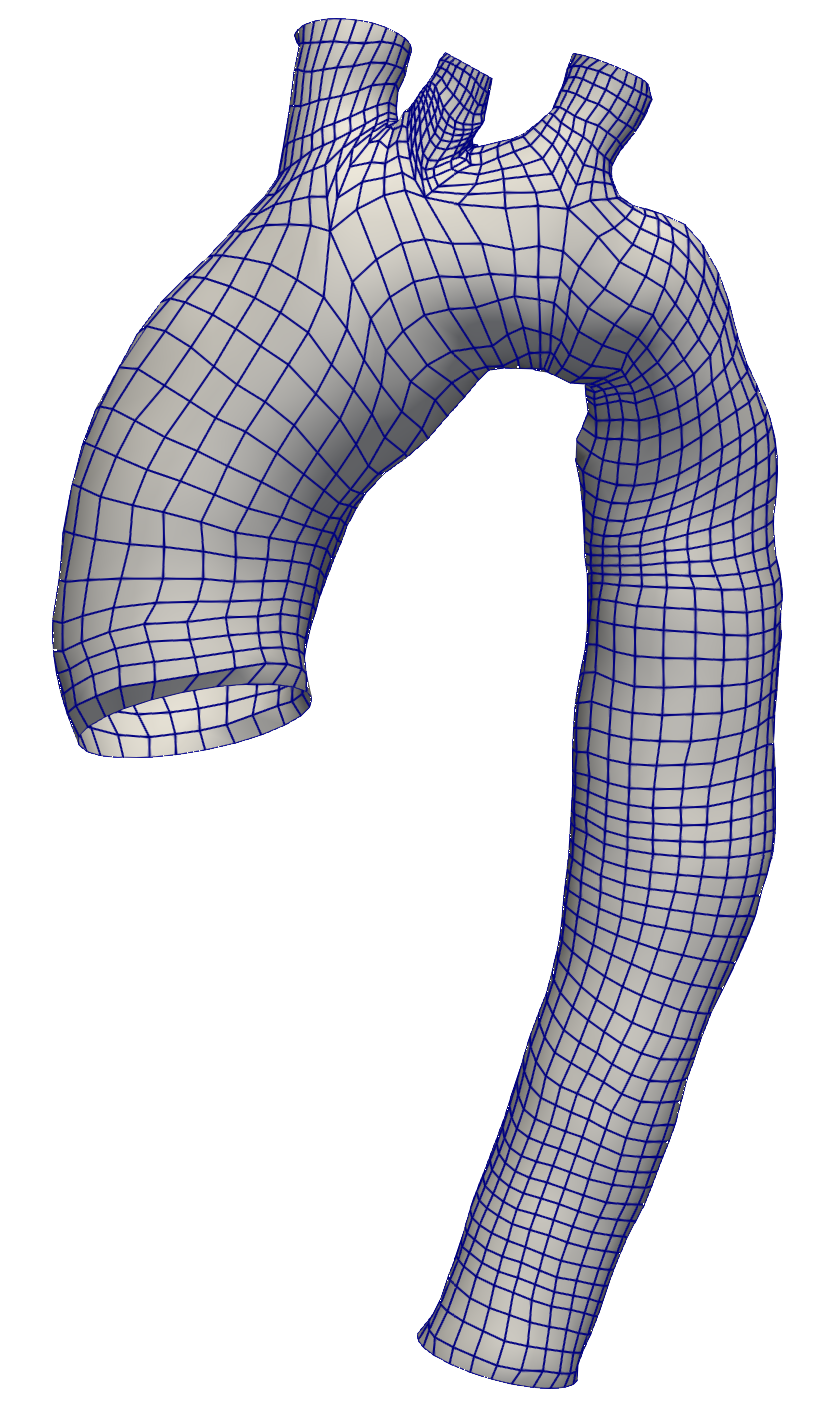}\hspace{.5cm}
  \includegraphics[width=0.11\textwidth]{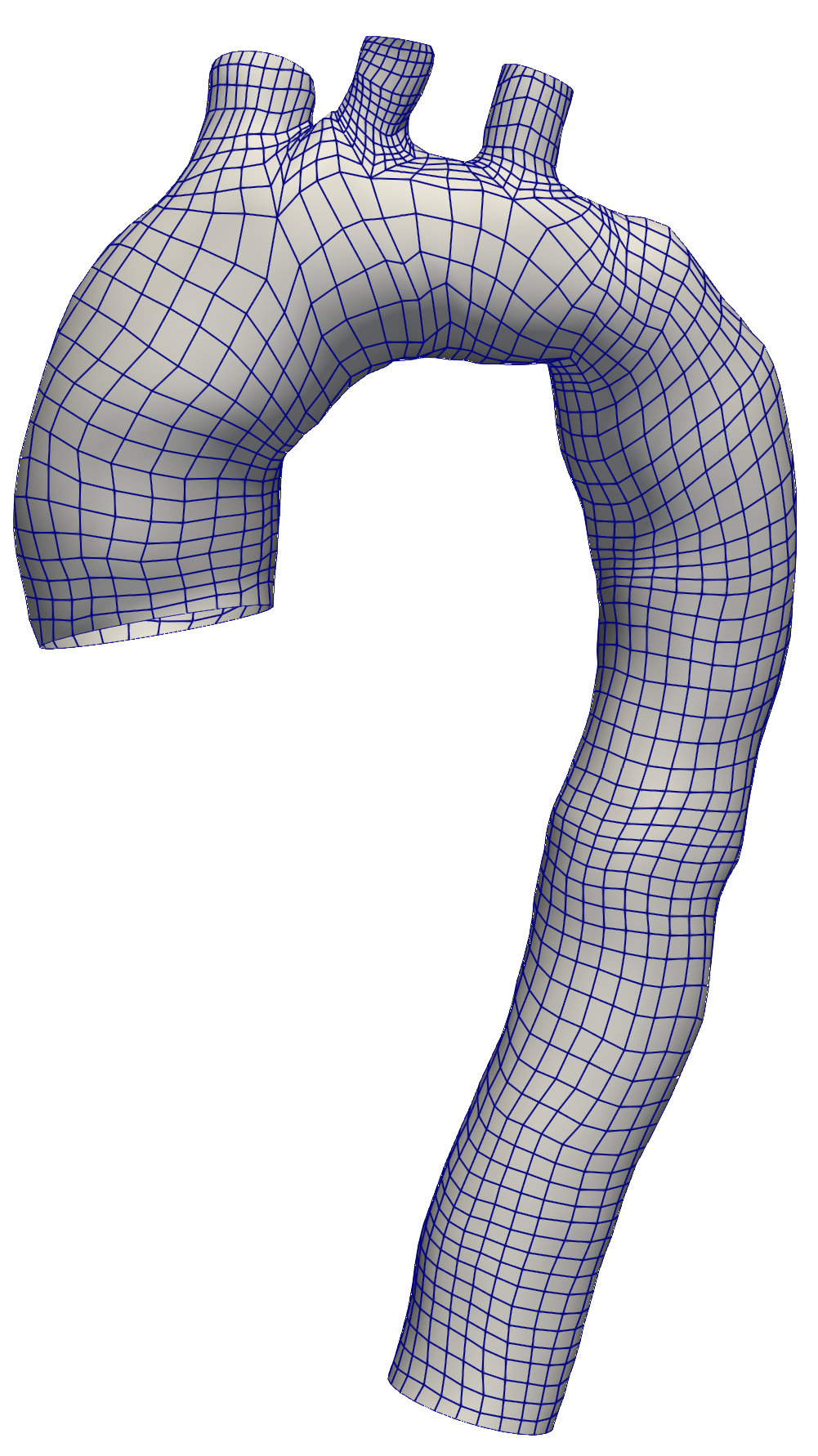}\hspace{.5cm}
  \includegraphics[width=0.11\textwidth]{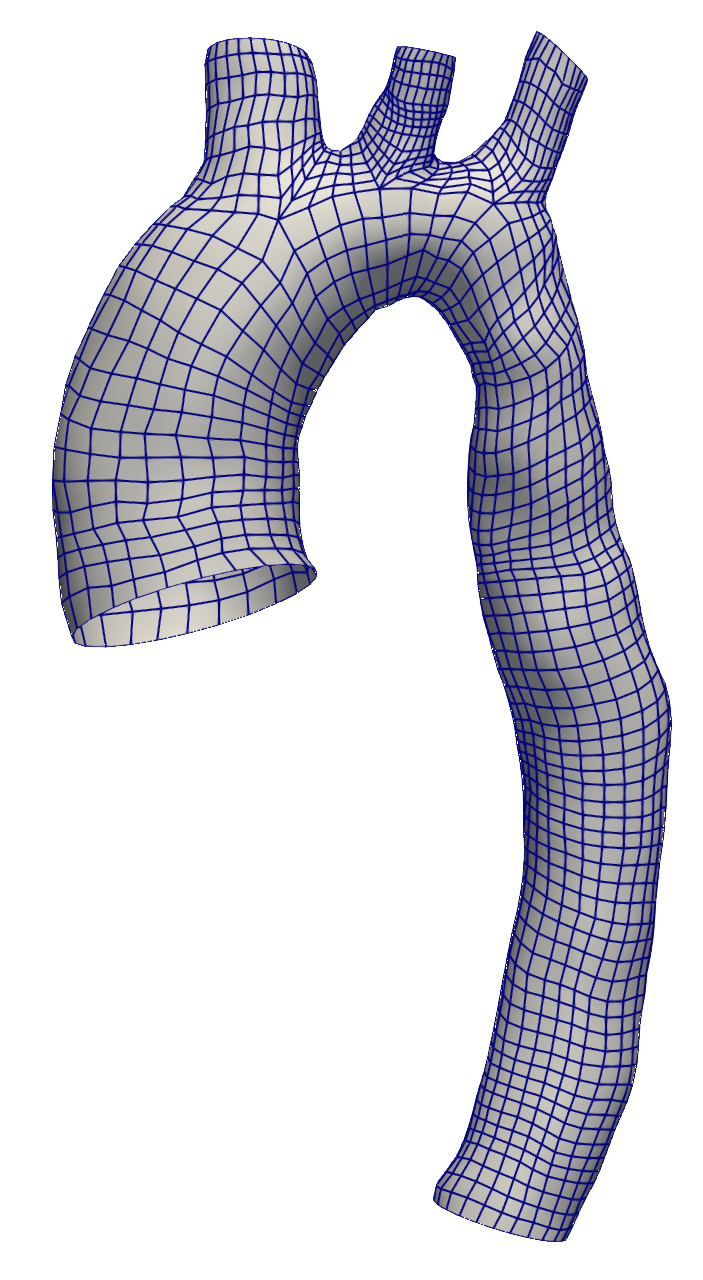}\hspace{.5cm}
  \includegraphics[width=0.11\textwidth]{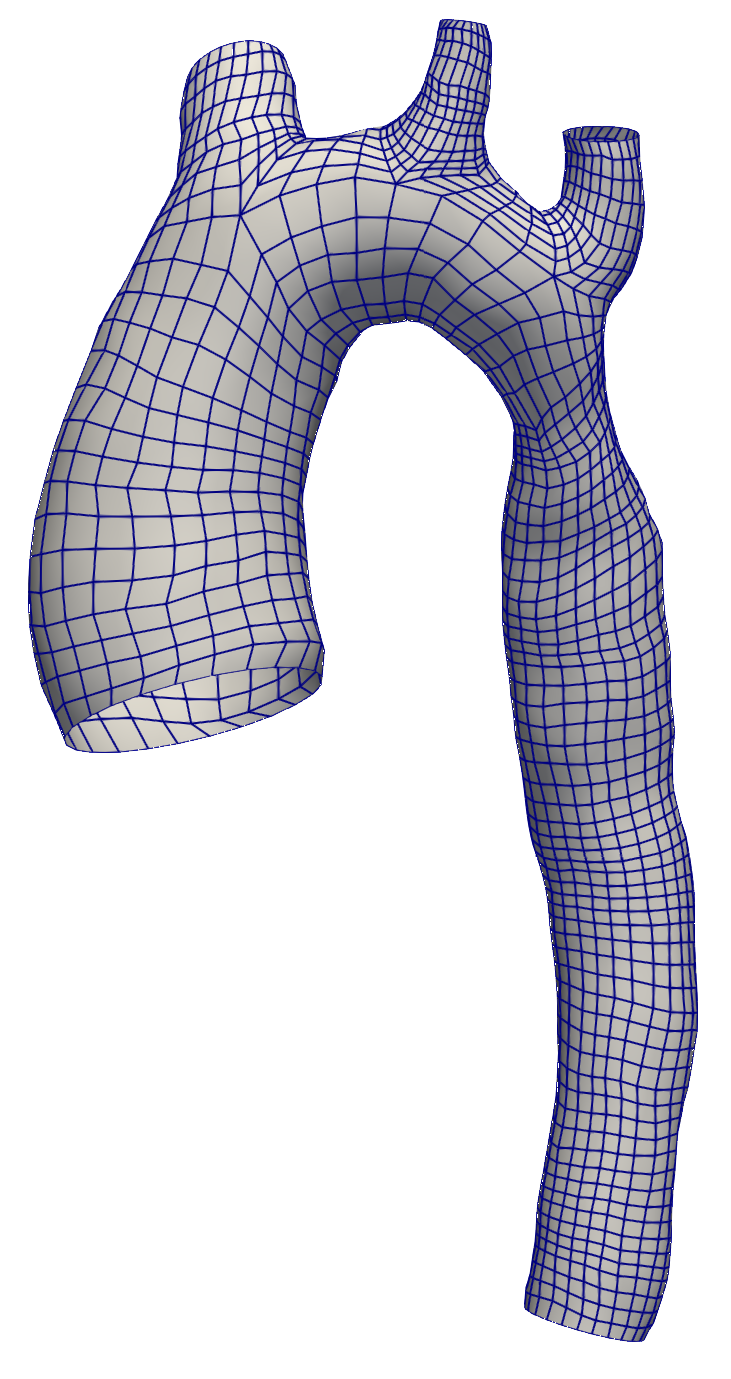}\\
  \end{center}
  \caption{Examples from batches 0, 10, 20 and 29 (from left to right). Top: perturbations with radius reduced by 30\% (factor $\alpha_r = 0.7$).
  Middle: perturbations with similar radius (factor $\alpha_r = 1$). Bottom: perturbations with radius increased by 30\% (factor $\alpha_r = 1.3$).}
  \label{fig:shape_examples_batches}
\end{figure}

Numerical simulations for uncertainty quantification have been run based on a test dataset containing 30 geometries.
For each of these, we generated three additional sets of $n_\mathrm{s} = 10$ perturbed domains by imposing a radius variation by a factor
$\alpha_r \in \{0.7, 1, 1.3\}$. Figure~\ref{fig:all_test_geometries} shows the whole test dataset, while Figure~\ref{fig:shape_examples_batches} shows a selection from the generated batches of geometries after the perturbation. The choice $\alpha_r \in \{0.7, 1.3\}$ is made to test the generative model and mesh extension algorithms with high deformations: they may not produce physiologically realistic geometries, but they are useful to explore the behavior of the model over a wide range of variations.

In all considered cases, the generative model yielded valid surfaces. However, for certain geometries,
particularly when reducing the radius of already relatively narrow ones, it was not possible to generate a valid volume mesh or to guarantee mesh quality
sufficient for stable simulations. These geometries (46 in total out of the full set of 900) were removed from the corresponding batches and not considered further in the study.
The number of valid and excluded cases in each batch is detailed in Table~\ref{tab:incomplete_batches}.

\begin{table}[!ht]
  \begin{center}
    \caption[Incomplete batches]
    {
      Valid geometries for each batch (numbered from $0$ to $29$). Each cell contains the number of full simulations available for each
      batch and for the corresponding value of $\alpha_r$. For the batches not listed here, all 10 perturbations, for all three values of $\alpha$, were available.
      The last row provides the total number of cases for each $\alpha_r$.
    }

    \label{tab:incomplete_batches}
    \begin{tabular}{l|c|c|c|c|c|c|c|c|c|c|c}
    & \multicolumn{10}{c|}{Incomplete batches} \\
$\alpha_r$
& 0 & 2 & 7 & 16 & 17 & 18 & 21 & 23 & 24 & 25 & total \\ \hline
$0.7$
& 6 & 4 & 0 & 1 & 10 & 8 & 5 & 10 & 8 & 10 & 262 \\
$1.0$
& 10 & 10 & 10 & 10 & 9 & 10 & 10 & 5 & 9 & 9 & 292 \\
$1.3$
& 10 & 10 & 9 & 10 & 10 & 10 & 10 & 10 & 6 & 10 & 295 \\
\end{tabular}
  \end{center}
\end{table}

In each geometry, the Navier--Stokes equations are solved numerically as described in Section~\ref{sec:numerical_model_calibration}.
The same periodic inflow profile, with a period of 1 second,
is used for all simulations, rescaled according to the area of the inlet surface in order to provide the same volumetric peak
of $\SI{400}{\cubic\centi\meter\per\second}$ (see Figure~\ref{fig:domain}).
No-slip boundary conditions on the vessel walls are enforced weakly due to the discontinuous Galerkin method employed by the \texttt{ExaDG} solver.
The parameters defining the boundary conditions are calibrated only once for each batch, depending on the base test geometry. The calibration is
based on modeling the flow split between the four outlets as a function of the areas, as described in~\cite{romor2025dataassimilationperformedrobust}.
Simulations are run for three cardiac cycles in order to reach a periodic state, and the final cycle is considered for evaluating the quantities of interest.

\begin{figure}[h!]
\begin{center}
  \includegraphics[width=0.12\textwidth]{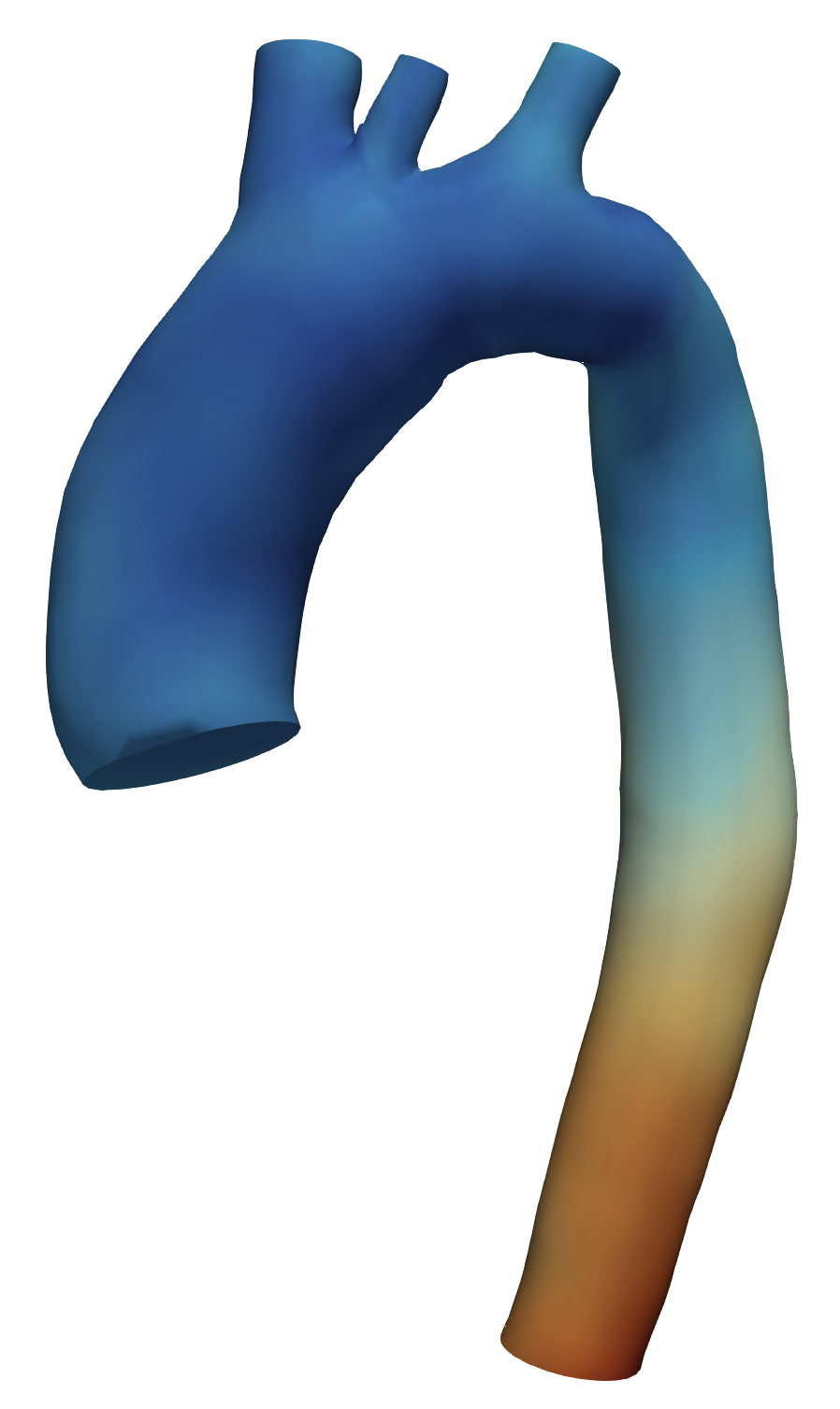}\hspace{1cm}
  \includegraphics[width=0.12\textwidth]{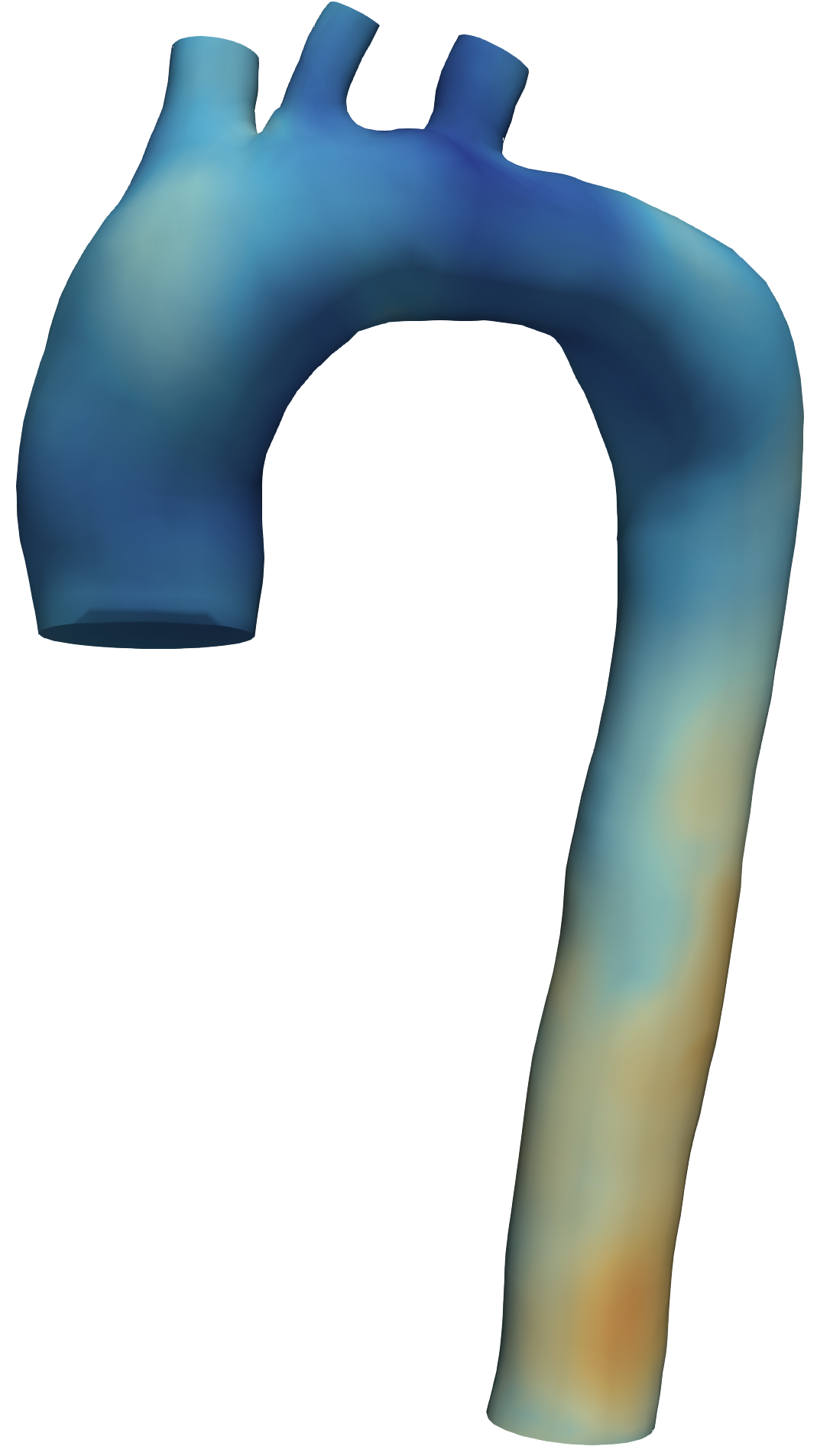}\hspace{1cm}
  \includegraphics[width=0.12\textwidth]{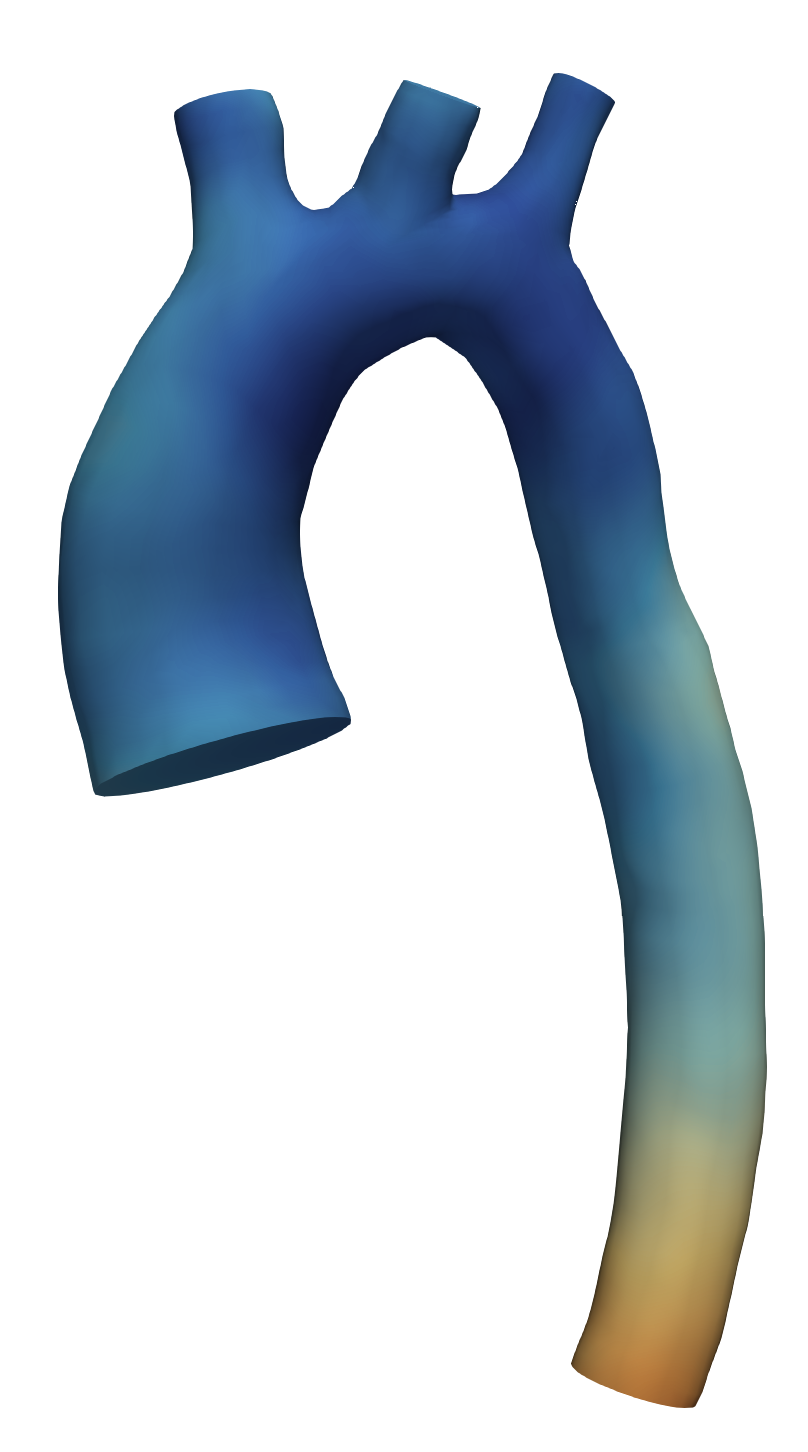}\hspace{1cm}
  \includegraphics[width=0.12\textwidth]{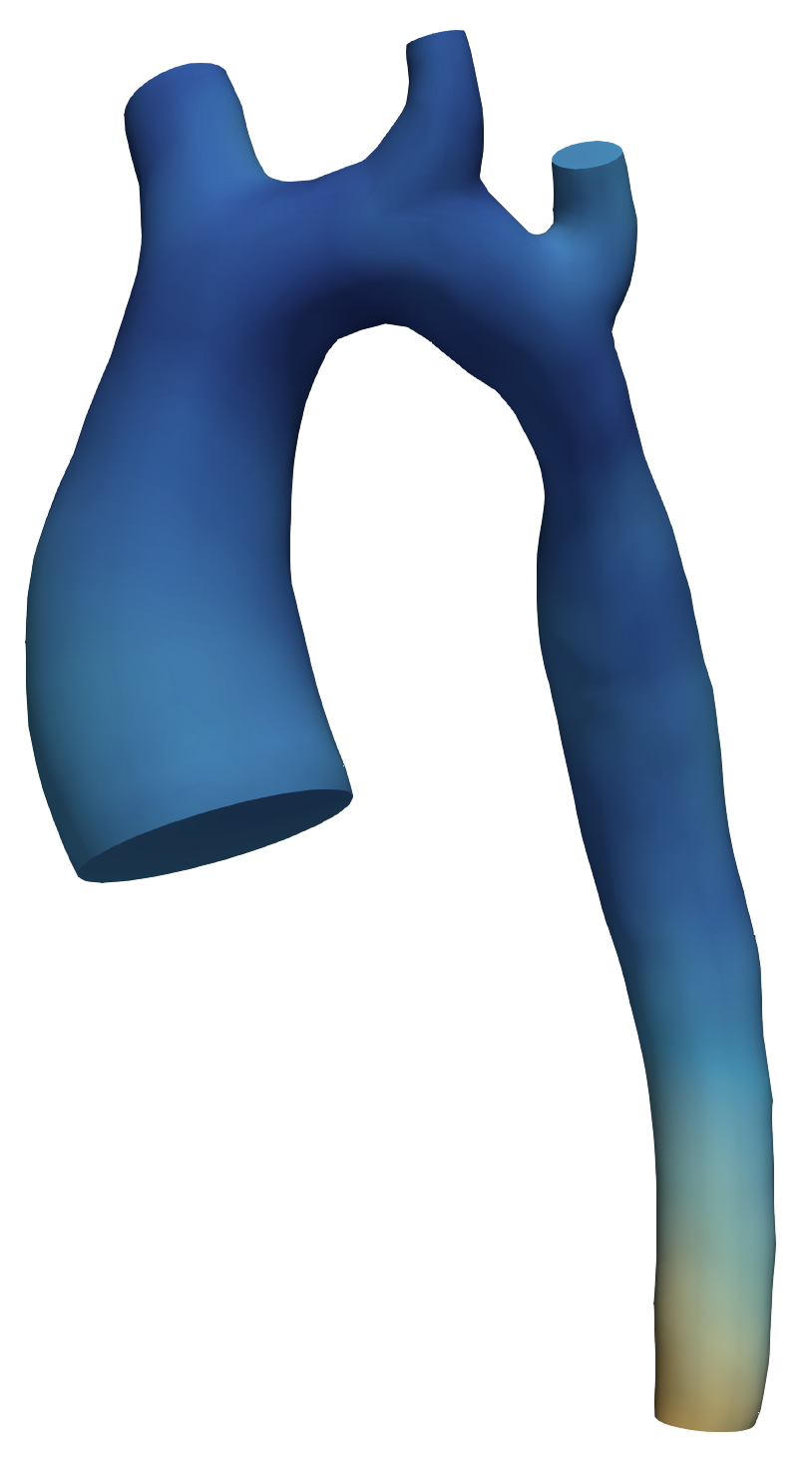}\\[0.2cm]
  \includegraphics[width=0.35\textwidth]{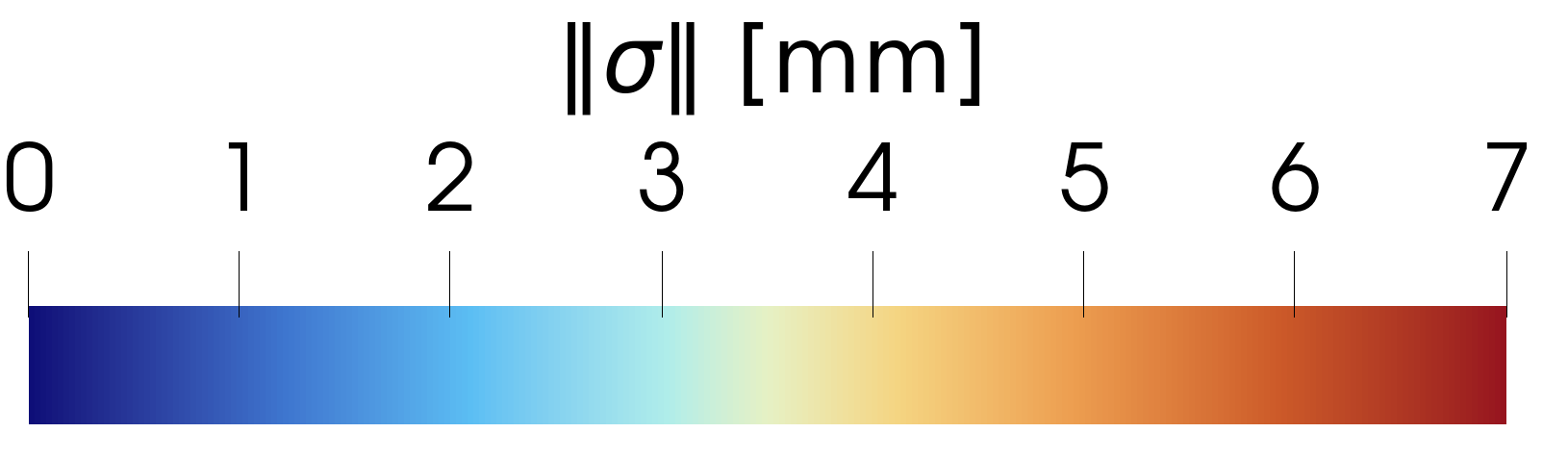}
  \end{center}
  \caption
  {
    Meshes generated from average coordinates for the batches 0, 10, 20, and 29. The surface color
    displays the sample standard deviation.
  }
  \label{fig:batch_variation}
\end{figure}

\begin{figure}[h!]
  \includegraphics[width=0.75\textwidth]{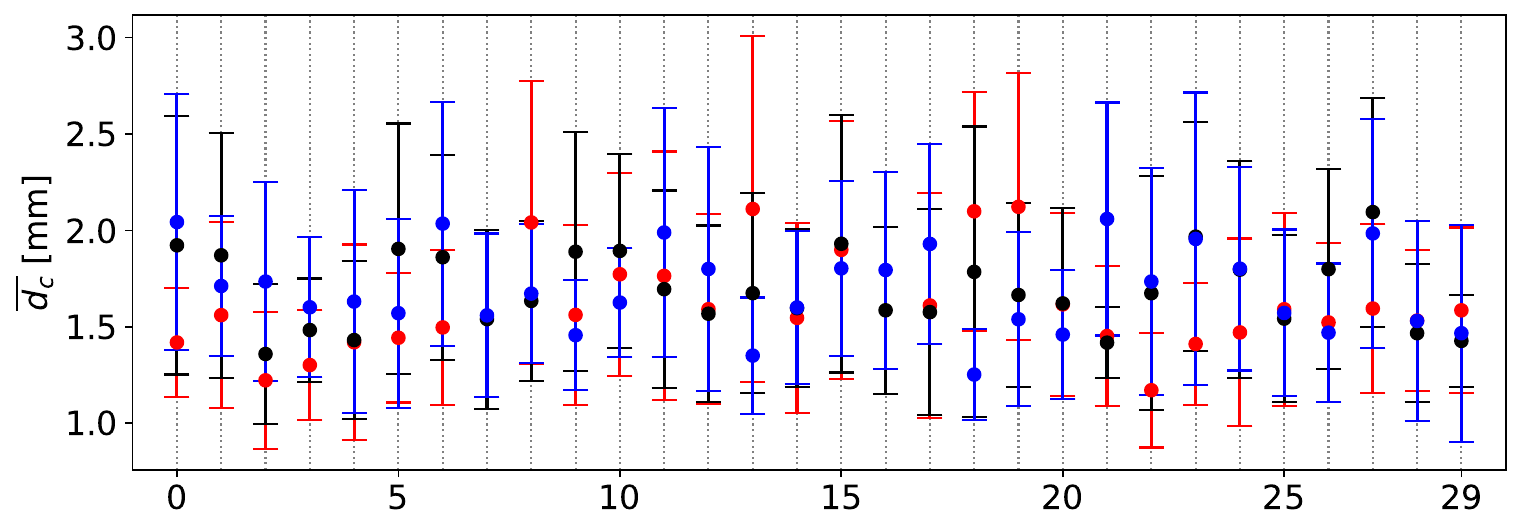}
  \includegraphics[width=0.75\textwidth]{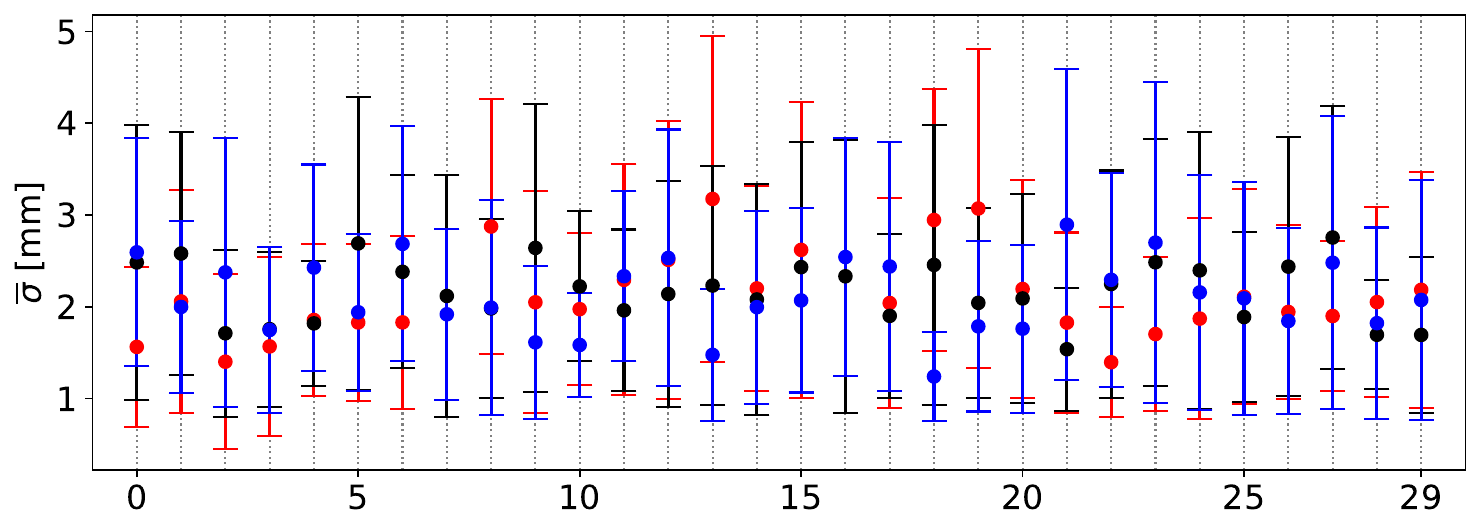}
  \caption
  {
    Shape variability within each batch. Top: For each batch index (0 to 29), the bars show mean and sample standard deviation of all pairwise Chamfer distances
    of the shapes within the batch. Bottom: For each batch index (0 to 29), the bars show mean and sample standard deviation of
vertex coordinate standard deviations.
  }
  \label{fig:mesh_stats}
\end{figure}

\begin{figure}[h!]
  \centering
  \includegraphics[width=0.7\textwidth]{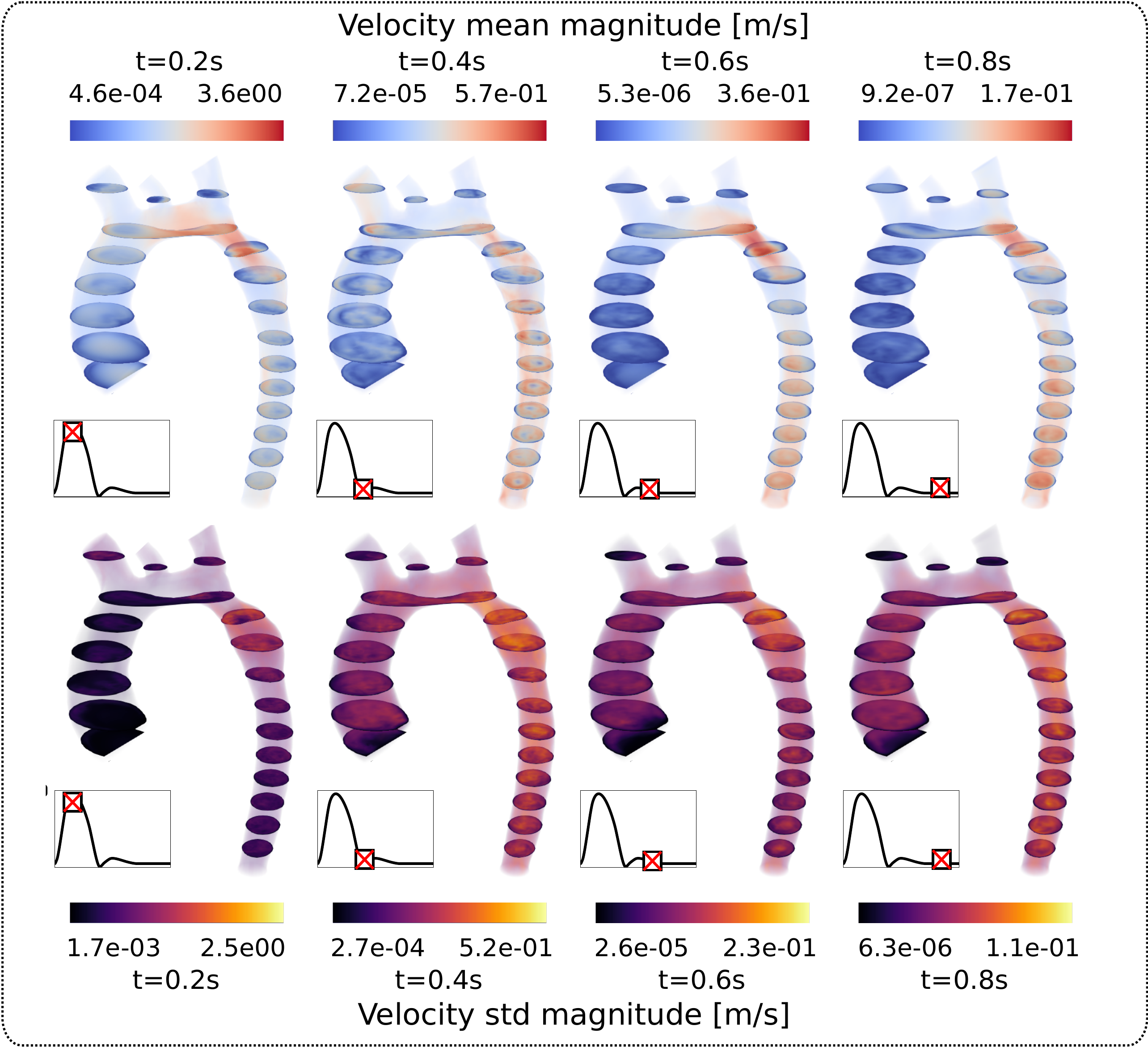}
  \caption{Velocity mean and standard deviation over $10$ sampled perturbations of one selected test geometry (the base geometry for batch 1) with $\alpha_r=1.0$, and for
  $t \in \left\{0.2\text{s}, 0.4\text{s}, 0.6\text{s}, 0.8\text{s}\right\}$.}
  \label{fig:velocity_snapshot}
\end{figure}

\begin{figure}[h!]
  \centering
  \includegraphics[width=0.7\textwidth]{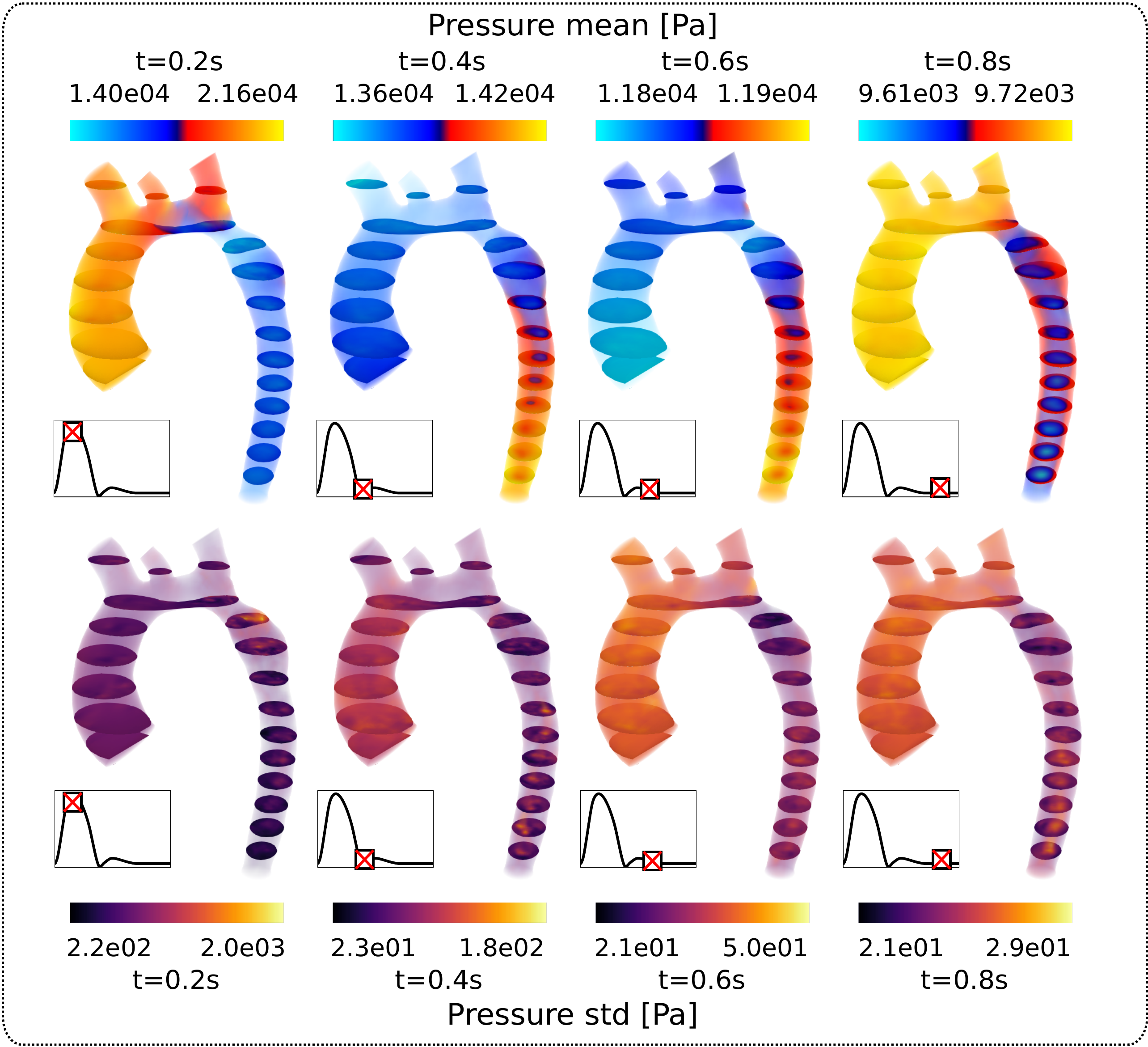}
  \caption{Pressure mean and standard deviation over $10$ sampled
  perturbations of one selected test geometry (the base geometry for batch 1), with $\alpha_r=1.0$, and for $t \in \left\{0.2\text{s}, 0.4\text{s}, 0.6\text{s}, 0.8\text{s}\right\}$}
  \label{fig:pressure_snapshot}
\end{figure}

To visualize the variability of the perturbed domains, Figure~\ref{fig:batch_variation} shows the
sample standard deviation of the vertex coordinates for batches 0, 10, 20, and 29.
Figure~\ref{fig:mesh_stats} gives an overview of the variability within each batch, our source of geometric uncertainty. Figures~\ref{fig:velocity_snapshot} and \ref{fig:pressure_snapshot} show the variability of velocity and pressure fields for one selected test geometry (the base geometry for batch 1) with $\alpha_r=1.0$ at different time steps. Most of the uncertainty in the velocity field is concentrated during diastole, in the region of the descending aorta, as observed also in~\cite{bovsnjak2025geometric}, due to the development of secondary flow patterns.

\subsection{Shape uncertainty quantification on selected biomarkers}
\label{sec:shape_uq_numerical_experiments}

We will compare the impact of geometric variation on a number of quantities of interest:

\begin{itemize}
  \item Overall maximum velocity magnitude throughout, and mean
  magnitude of the velocity components across cross--sections orthogonal to the aorta's centerline.

  \item Outflow through the descending aorta ($Q_\mathrm{desc}$).

  \item Overall mean pressure $\bar P$ and pressure difference $\Delta P$ between the descending aorta outlet and the inlet.

  \item Wall shear stress (WSS), defined as
  \[
    \tau_w(\bx) = \nu \frac \partial {\partial \bn} \Big(\bu - \left(\bu \cdot \bn\right)\bn \Big)(\bx),
  \]
  where $\bn = \bn(\bx)$ is the outer normal at a boundary point $\bx \in \Gamma$, so that homogeneous shear flow
  over a plane corresponds to wall shear stress in the direction opposite to the flow. The WSS will be analyzed mostly in terms of its mean magnitude
  over $\Gamma_\mathrm{wall}$ or over a reference patch $\Gamma_\mathrm{ref} \subset \Gamma_\mathrm{wall}$ on the underside of the descending aorta.
  \item Oscillatory shear index (OSI), defined as
  \[
    \mathrm{OSI}(\bx) = \frac 1 2 \left( 1 - \frac {\left|\int_2^3 \tau_w(\bx) \right| \mathrm dt} {\int_2^3 |\tau_w(\bx)| \mathrm dt} \right),
  \]
which measures how strongly $\tau_w$ oscillates over a given length
  of time during the simulated cardiac cycle, at a given boundary point $\bx$.
  Note that, unlike the other quantities introduced here, the OSI is not time-dependent. We will, as with $\tau_w$ itself,
  consider the mean OSI over $\Gamma_\mathrm{wall}$ or $\Gamma_\mathrm{ref}$.

  \item Secondary flow degree (SFD), defined, for a cross section $S$, as
  \[
    \mathrm{SFD}(S) = \frac
      {\int_S \left| \bu - (\bu \cdot \bn) \bn \right| \mathrm d\bx}
      {\int_S \left| \bu \cdot \bn \right| \mathrm d\bx},
  \]
which measures the ratio of tangential to normal flow over a cross--section $S$.
  Low SFD corresponds to strongly directed flow through $S$, whereas high SFD indicates a large amount of in--plane flow
  due to eddies or (in the aortic arch) lateral outlets.

  \item Normalized flow displacement (NFD), defined, for a cross-section $S$, as
  \begin{equation}\label{eq:nfd}
    \mathrm{NFD}(S) = \frac {|\bx_\bn(S) - \bx_g(S)|} {r_H(S)}.
  \end{equation}
  The NFD is a measure of \textit{flow eccentricity}, i.e., of the normalized
  distance between the moment of the flow through $S$,
  $\bx_\bn(S) = \frac {\int_S \left| (\bu \cdot \bn) \right| \bx \mathrm d\bx} {\int_S \left| \bu \cdot \bn \right| \mathrm d \bx}$,
  and its geometric centroid, $\bx_g(S) = \frac {\int_S \bx \mathrm d\bx} {\int_S \mathrm d \bx}$.
  In \eqref{eq:nfd}, the normalizing factor $r_H(S)$ is the hydraulic radius, which is typically defined
  as the ratio of area to (wetted) perimeter; we use the more stable estimate
  $r_H(S) = \frac 3 4 \frac {\int_S \left| \bx - \bx_G(S) \right| \mathrm d \bx}{\int_S \mathrm d \bx}$,
  which matches the conventional definition if $S$ is a disc.

\end{itemize}

\begin{figure}[h!]
\centering
  \includegraphics[width=0.23\textwidth]{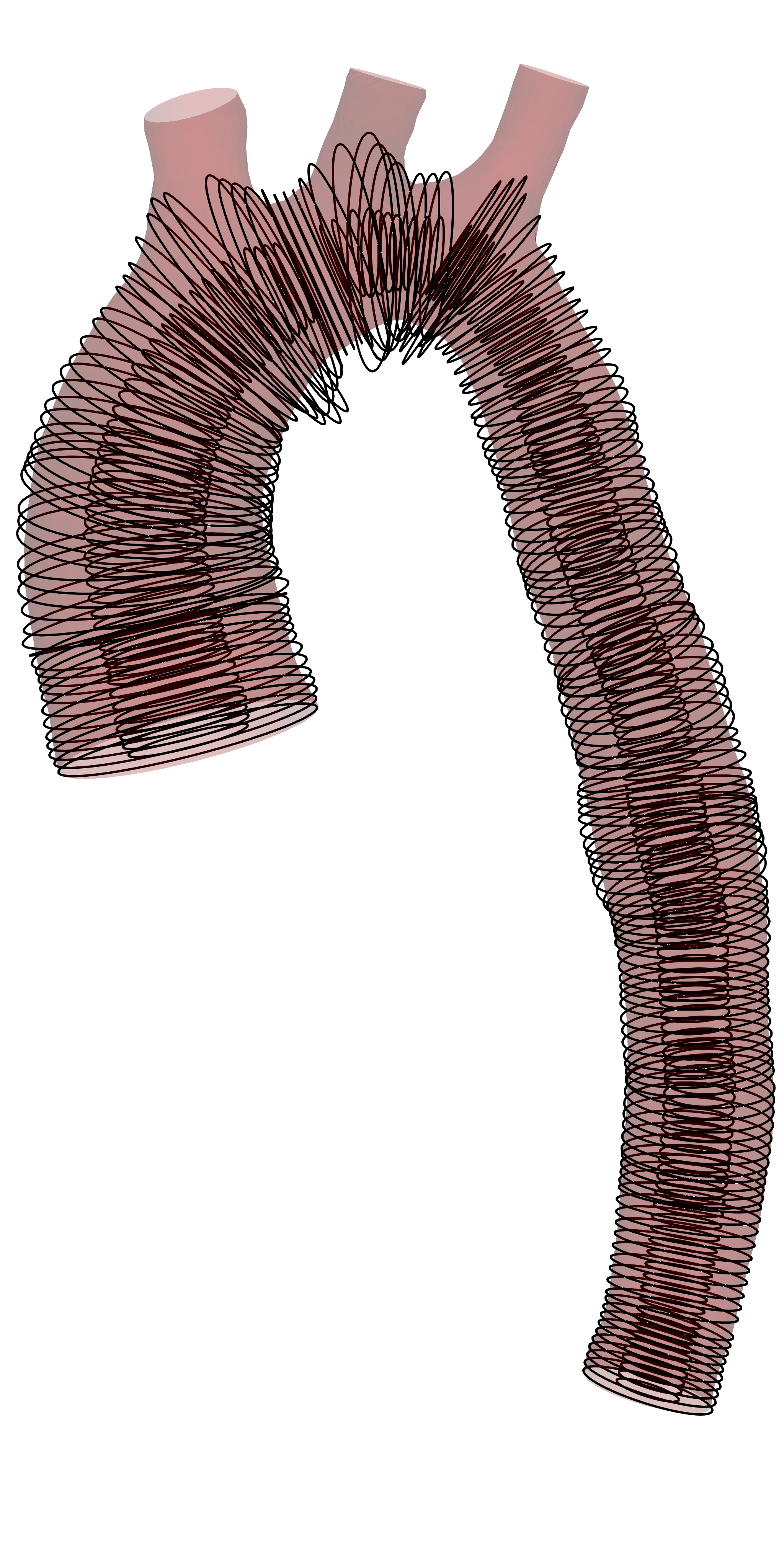}
  \caption
  {
    Left:
    selected geometry (from batch 20) showing  the cross--sections used for computing averages of  quantities of interests.
    The outer rings are in-plane circles matching an estimate of the outer radius of each cross--section.
    The inner rings show the estimated hydraulic radius $r_H$.
  }
  \label{fig:cross_sections}
\end{figure}

Figure~\ref{fig:cross_sections} shows the cross-sections $S$ used for the computation of the quantities of interest for one case in batch 20.

\begin{figure}[h!]
\begin{center}
  \includegraphics[width=0.23\textwidth]{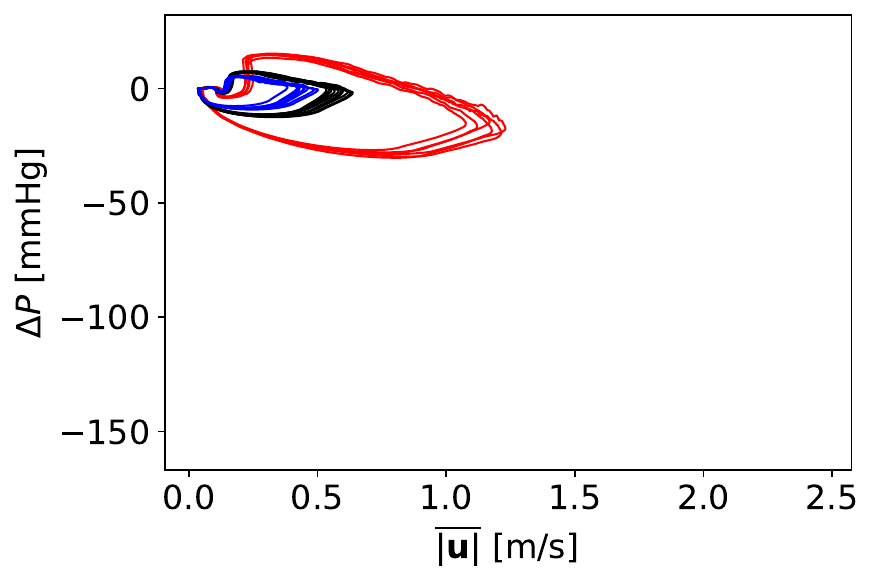}
  \includegraphics[width=0.23\textwidth]{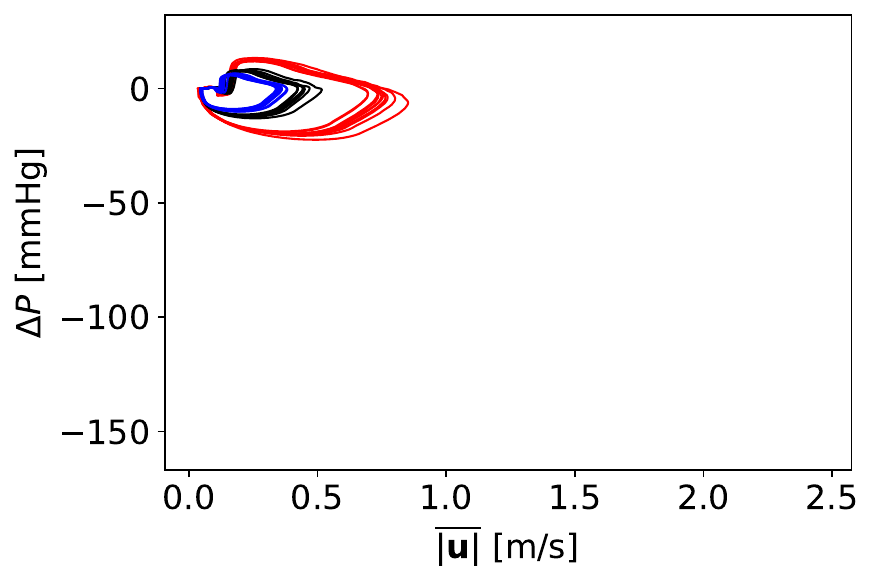}
  \includegraphics[width=0.23\textwidth]{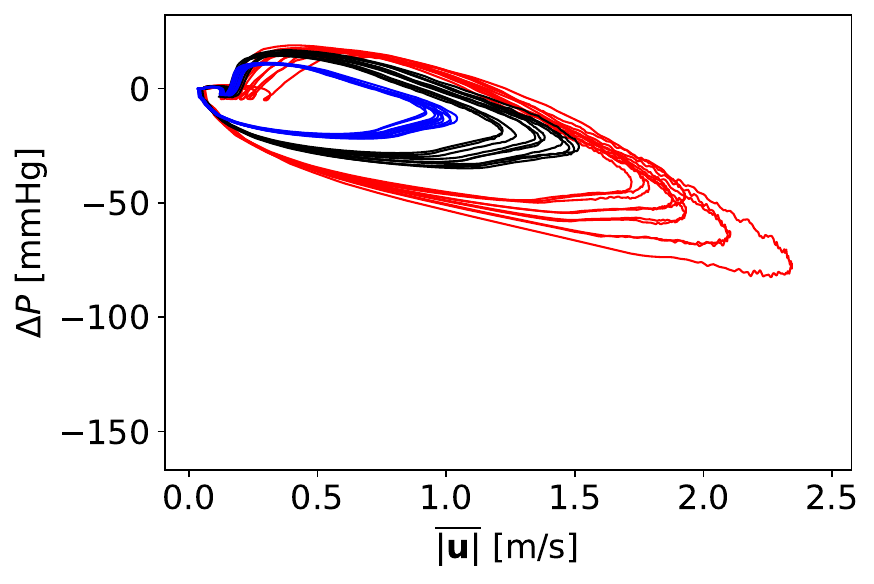}
  \includegraphics[width=0.23\textwidth]{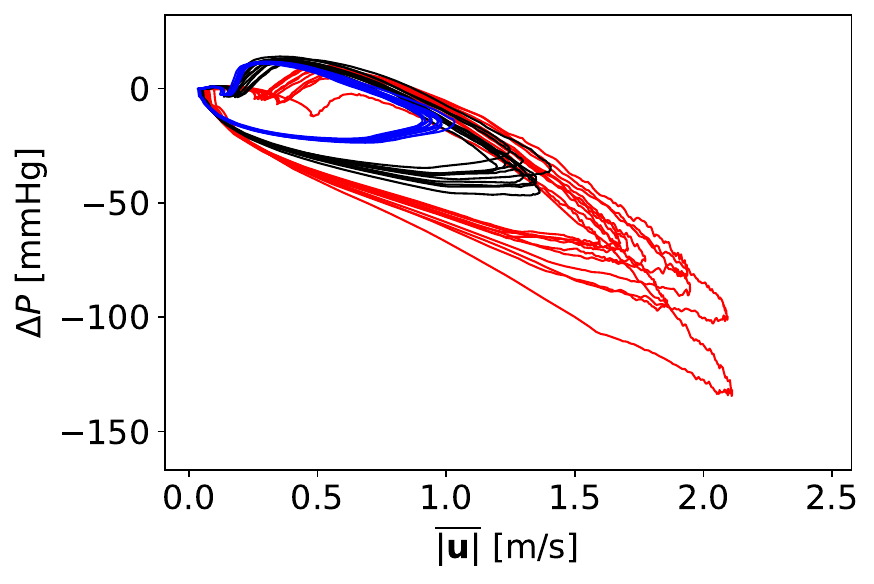}\\[0.2cm]
    \includegraphics[width=0.23\textwidth]{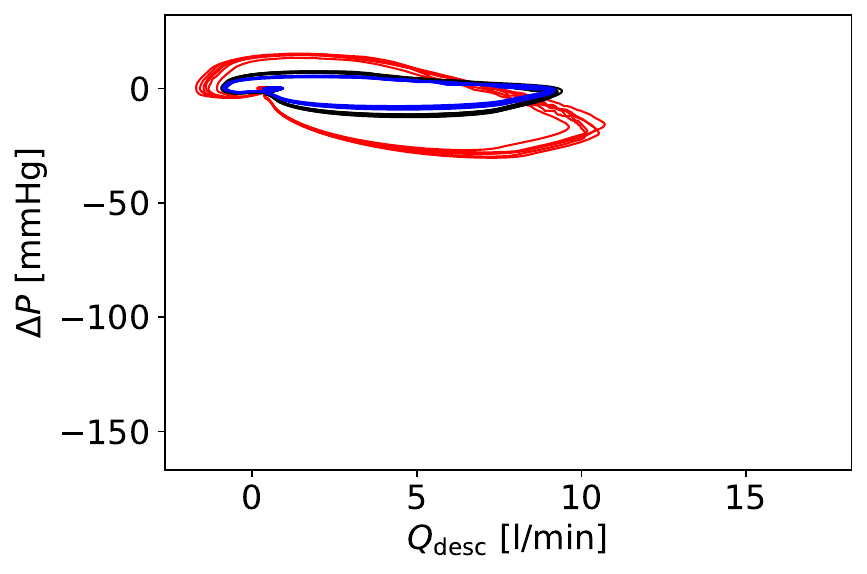}
  \includegraphics[width=0.23\textwidth]{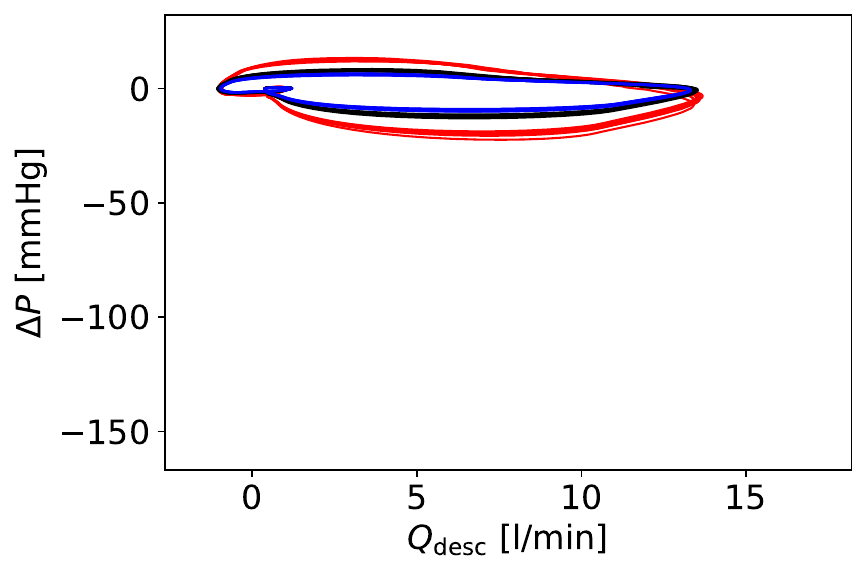}
  \includegraphics[width=0.23\textwidth]{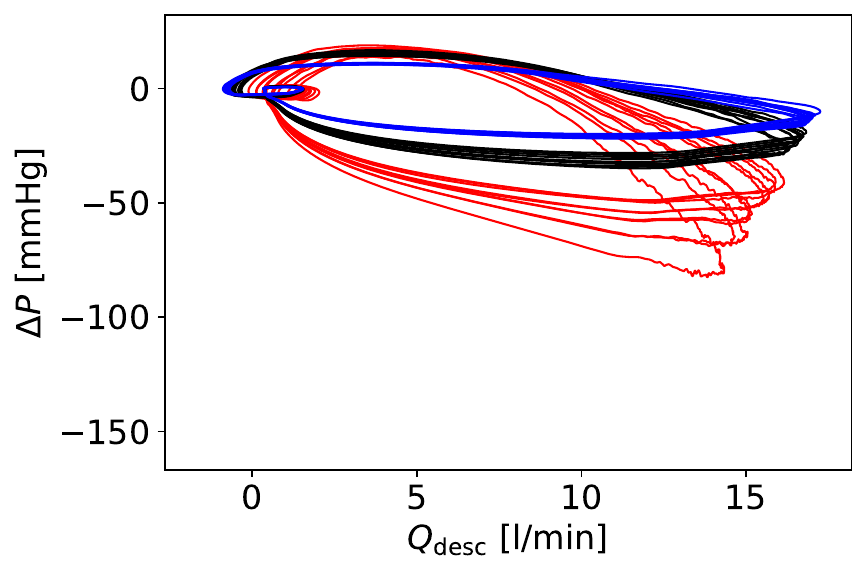}
  \includegraphics[width=0.23\textwidth]{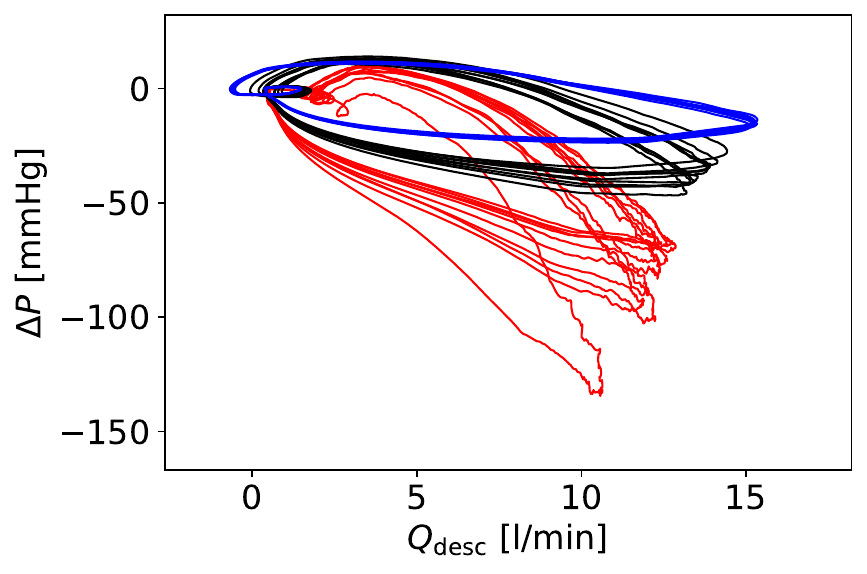}
  \end{center}
  \caption
  {
    Top row: Pressure difference between descending aorta and inlet versus mean cross-sectional normal velocity magnitude (averaged over the length of the domain)
    over one cardiac cycle (time goes counterclockwise).
    Bottom row: Pressure difference between descending aorta and inlet versus outflow at the descending aorta versus
    pressure difference between descending aorta and inlet
    over one cardiac cycle.
    From left to right: batches 0, 10, 20, and 29.
    \legendnote
  }
  \label{fig:loop_u_p}
\end{figure}

Figures~\ref{fig:loop_u_p} show the results for the pressure drop as a function of the averaged cross-sectional velocity and of the outflow at the descending aorta for batches 0, 10, 20, and 29. The cases with larger vessel radii (batches 0 and 10), where the pressure drop is lower (to achieve the same volumetric flow), show
less sensitive behavior with respect to the geometry variation. Similarly, within the same batch,
the highest impact of the geometry is seen for the set of simulations with reduced radius ($\alpha_r = 0.7$).

\begin{figure}[h!]
\begin{center}
  \includegraphics[width=0.23\textwidth]{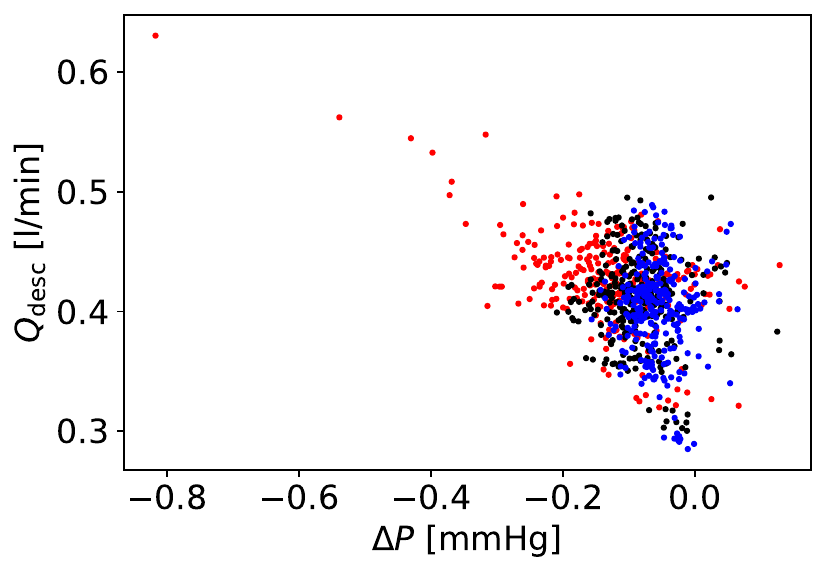}
  \includegraphics[width=0.23\textwidth]{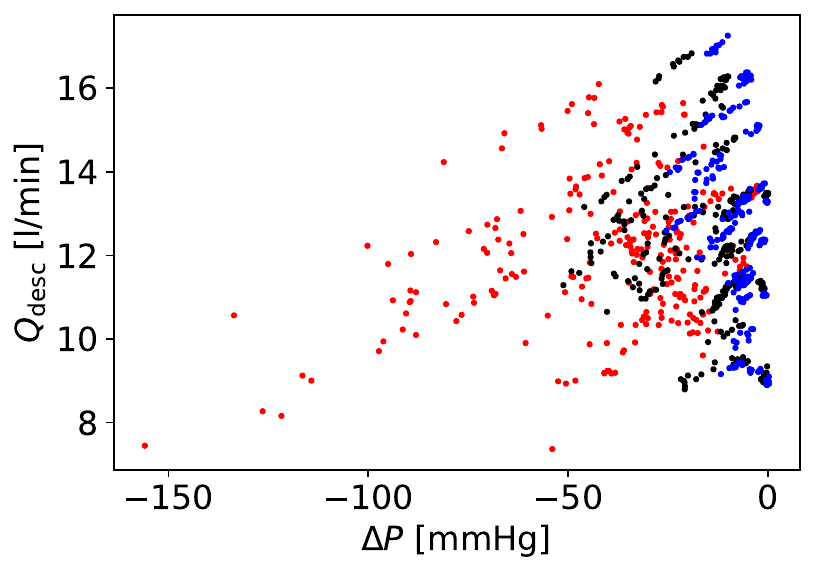}
  \includegraphics[width=0.23\textwidth]{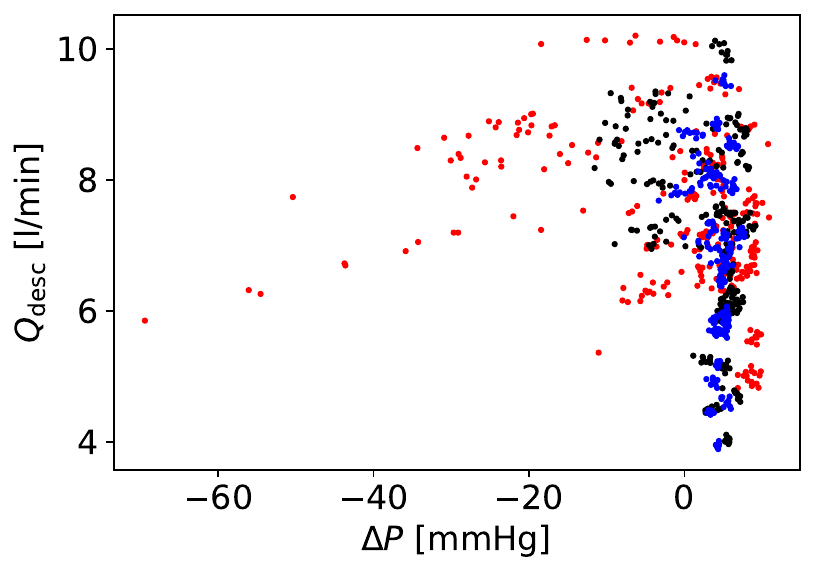}
  \includegraphics[width=0.23\textwidth]{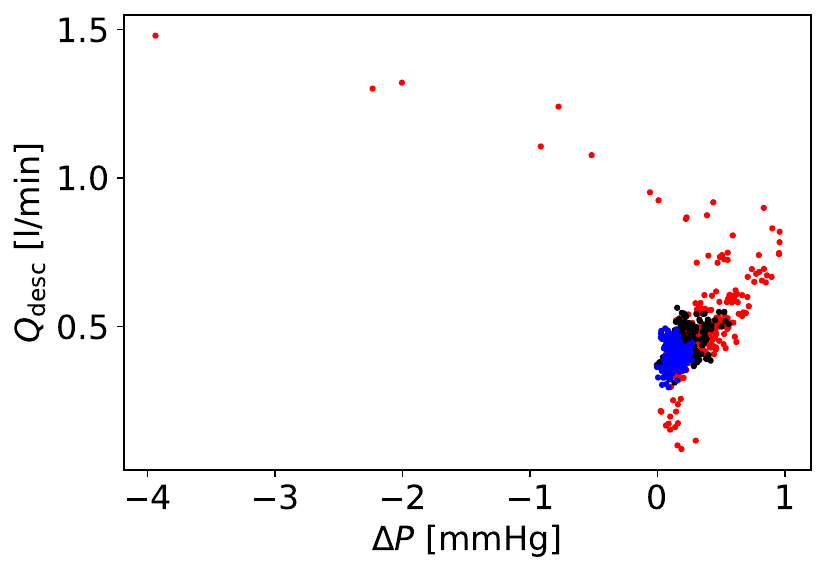} \\
    \includegraphics[width=0.23\textwidth]{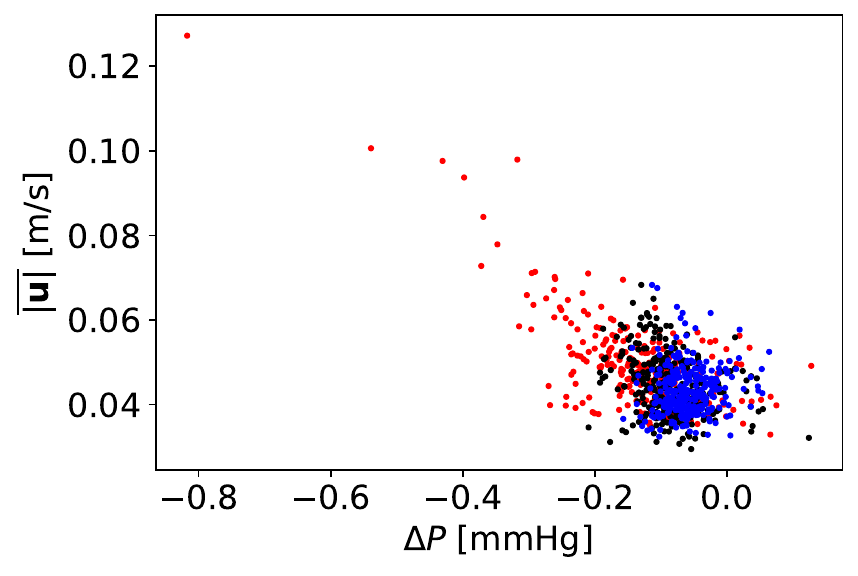}
  \includegraphics[width=0.23\textwidth]{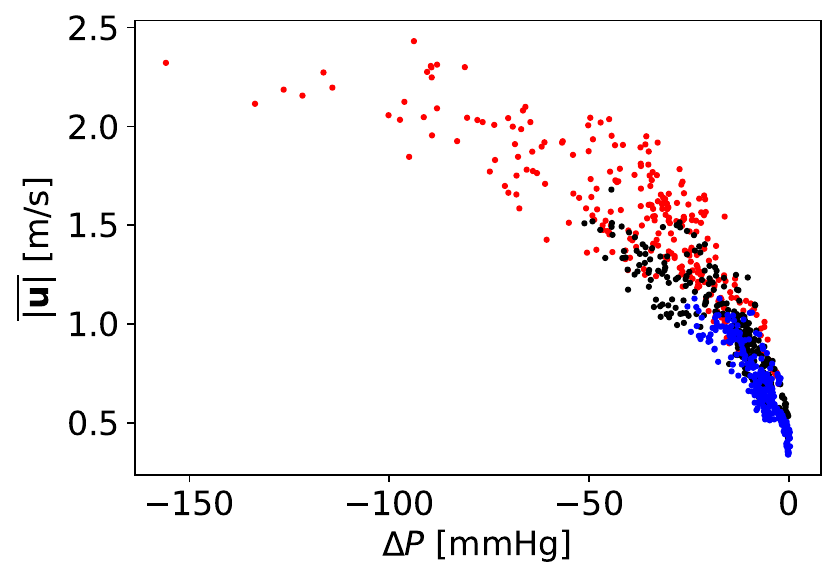}
  \includegraphics[width=0.23\textwidth]{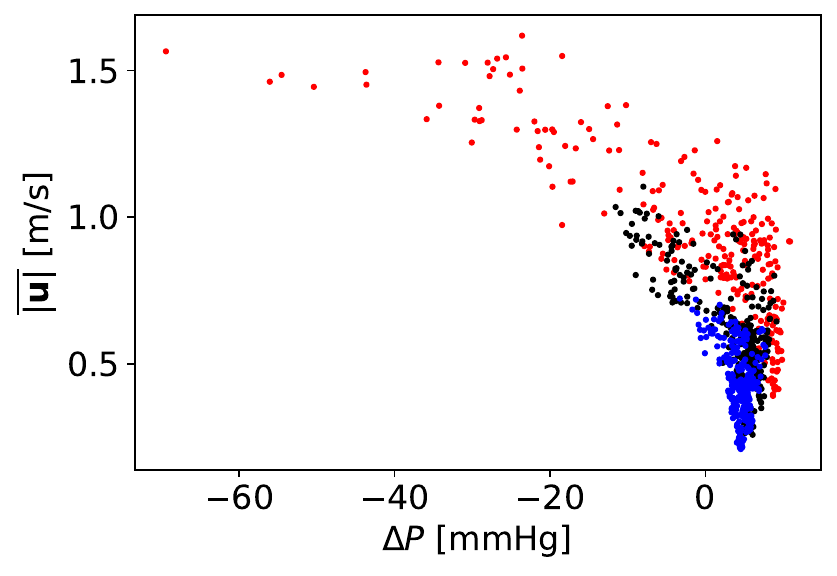}
  \includegraphics[width=0.23\textwidth]{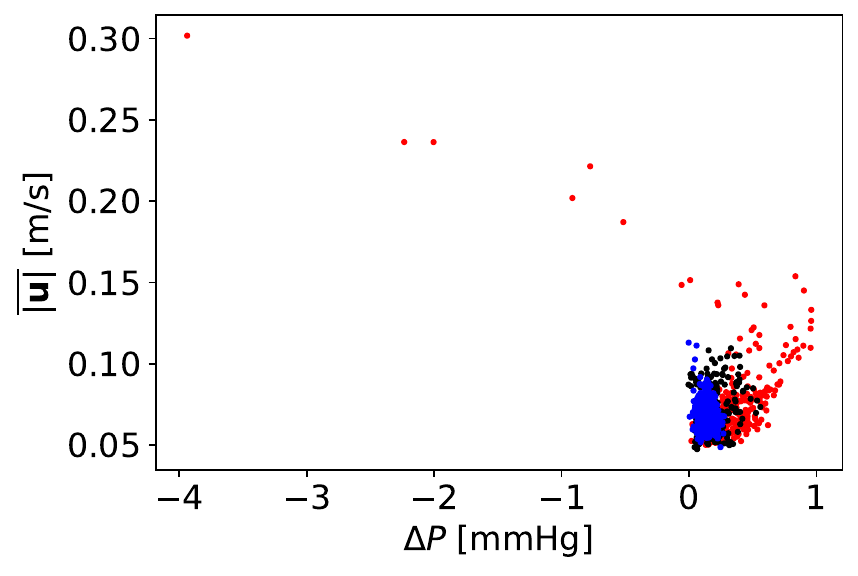}
\end{center}
  \caption
  {
    Top row: Pressure difference between descending aorta and inlet versus outflow at the descending aorta
    for all simulations. Bottom row: Pressure difference between descending aorta and inlet versus mean cross--sectional velocity magnitude
    (averaged over the length of the domain).
    From left to right: times $t \in \{ 2, 2.16, 2.3, 2.7 \}$,
    corresponding to the end of diastole, peak flow, end of systole, and middle diastole.
    \legendnote
  }
  \label{fig:scatter_Q_p}
\end{figure}

Figure~\ref{fig:scatter_Q_p} analyzes the variability of the results at different instants in time. In particular,
the fraction of outflow through the descending aorta at peak systole varies notably.
This behavior can be seen as a consequence of the differences in flow splits, which are imposed through the Windkessel boundary conditions
(Section~\ref{sec:numerical_model_calibration}). The cases with lower radii are also those showing
the highest pressure variability.
Figure~\ref{fig:scatter_Q_p} illustrates the velocity--pressure relationship for all simulations at different time steps.
At peak systole ($t = \SI{2.16}{s}$), which corresponds to large pressure differences, larger pressures and large variations
are always associated with narrower geometries.

\begin{figure}[htb!]
\centering
  \includegraphics[width=0.23\textwidth]{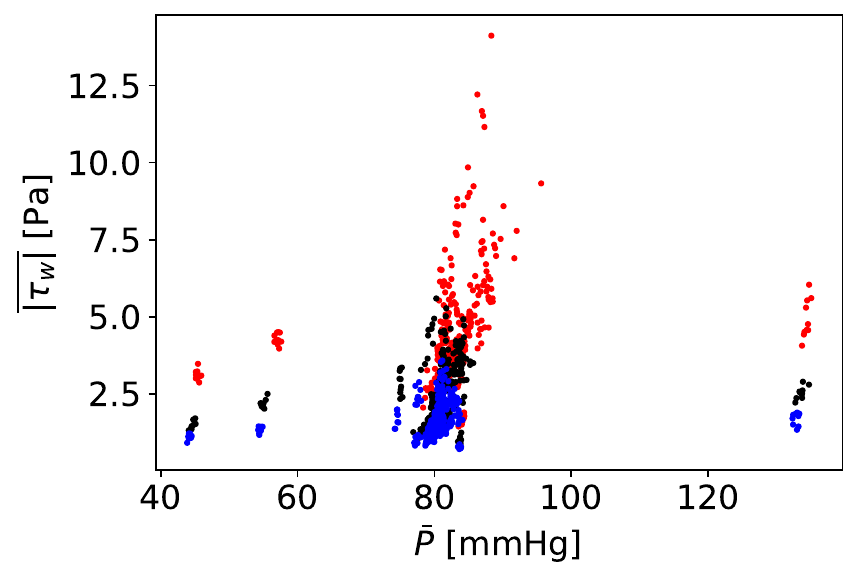}
  \includegraphics[width=0.23\textwidth]{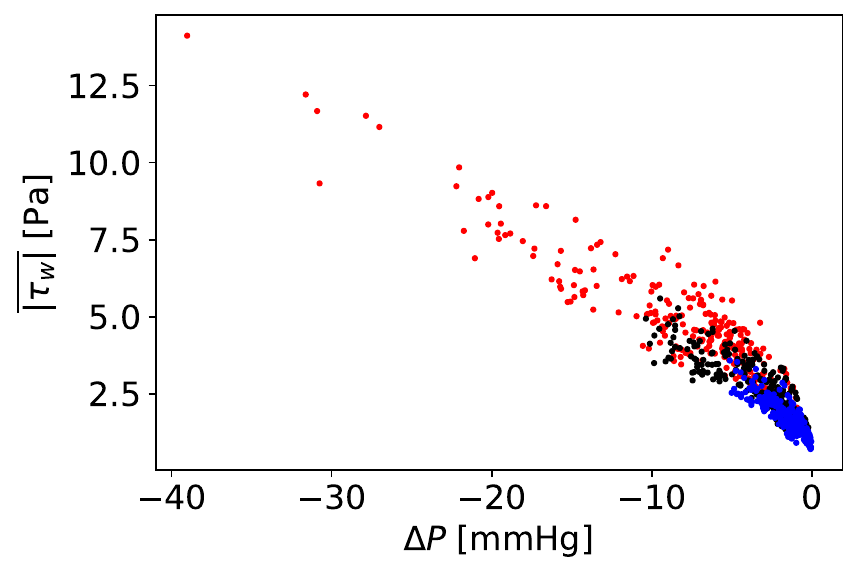}\\
    \includegraphics[width=0.23\textwidth]{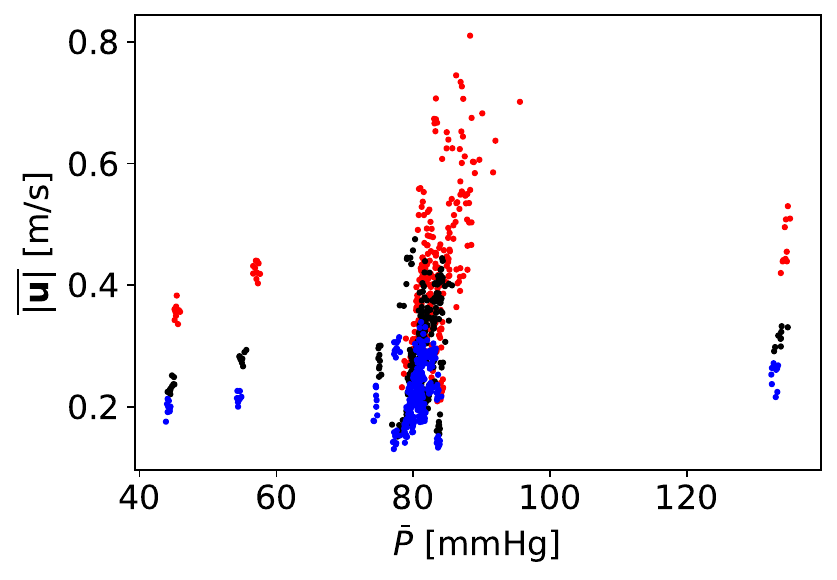}
  \includegraphics[width=0.23\textwidth]{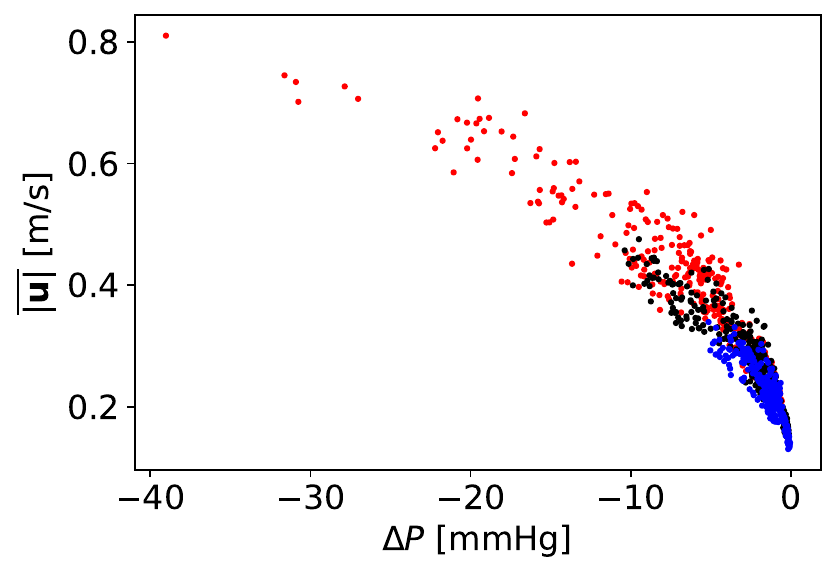}
  \caption
  {
  Mean wall shear stress (top) and mean velocity magnitude (bottom) versus
    mean pressure (left) and mean pressure drop (right), for all simulations,
    averaged over time. Note the clear batch separation in the
    absolute mean pressure compared to the pressure drop.
    \legendnote
  }
  \label{fig:scatter_P_bar}
\end{figure}

Figure~\ref{fig:scatter_P_bar} shows that the correlation
between pressure and velocity, as well as between pressure and wall shear stress, is clearer when examined in terms of
pressure differences rather than absolute pressure, since the latter is strongly separated between batches.

\begin{figure}[h!]
  \includegraphics[width=0.23\textwidth]{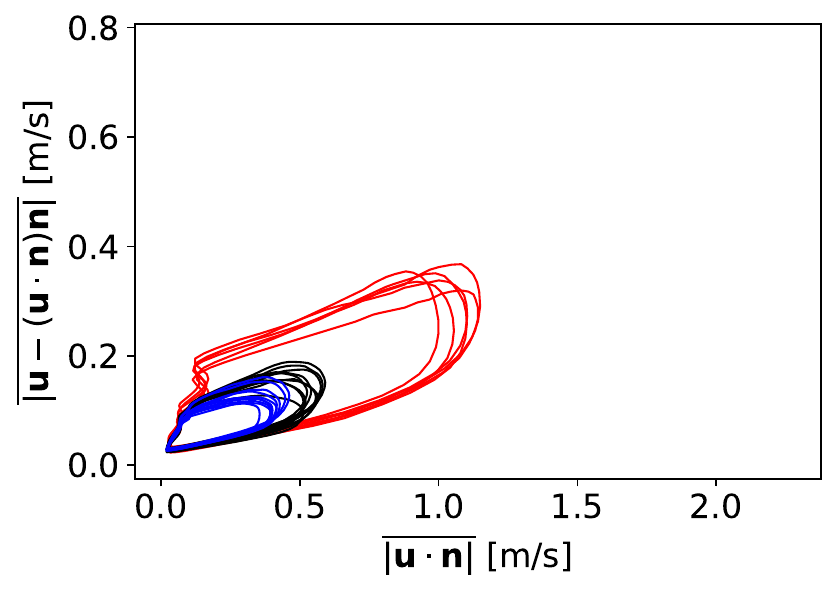}
  \includegraphics[width=0.23\textwidth]{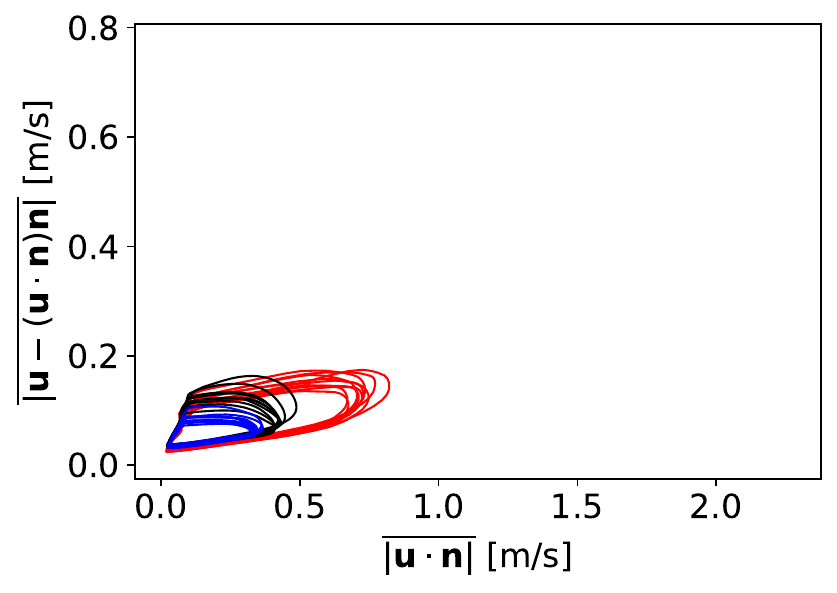}
  \includegraphics[width=0.23\textwidth]{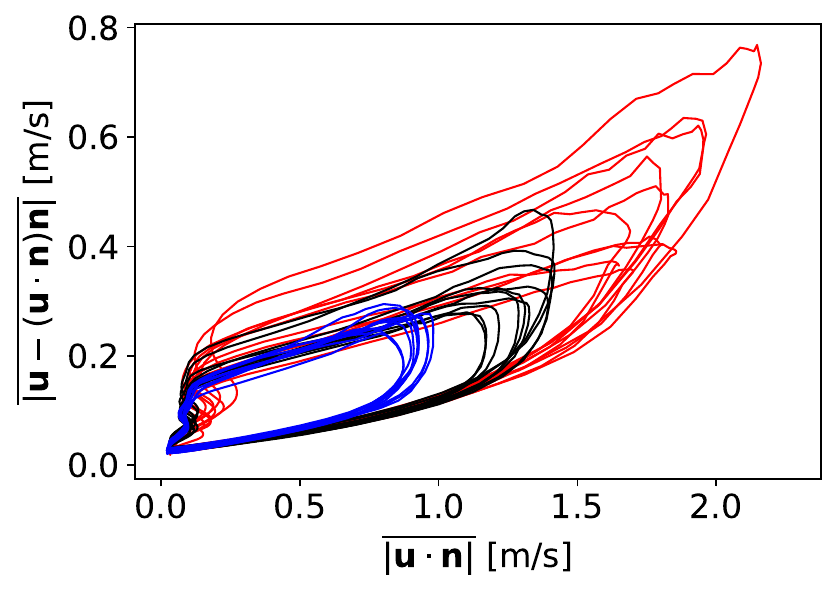}
  \includegraphics[width=0.23\textwidth]{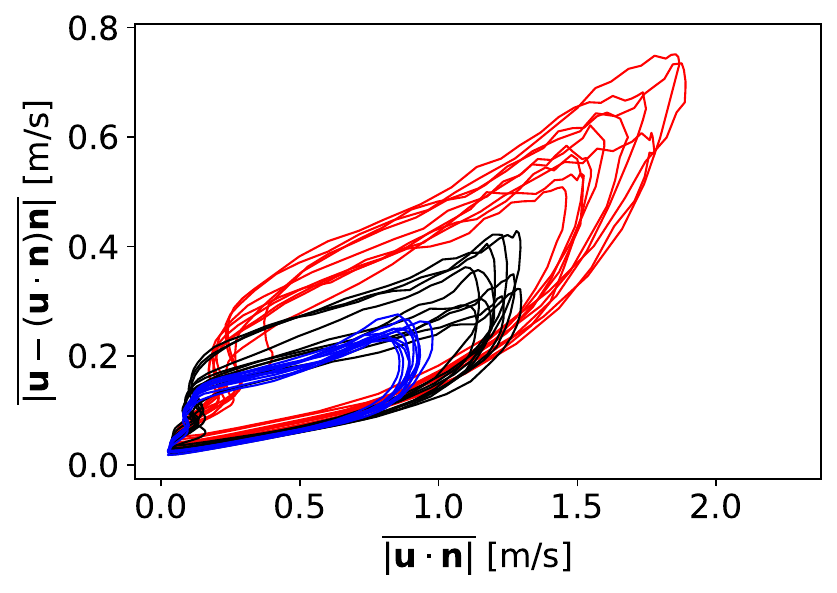}\\

  \includegraphics[width=0.23\textwidth]{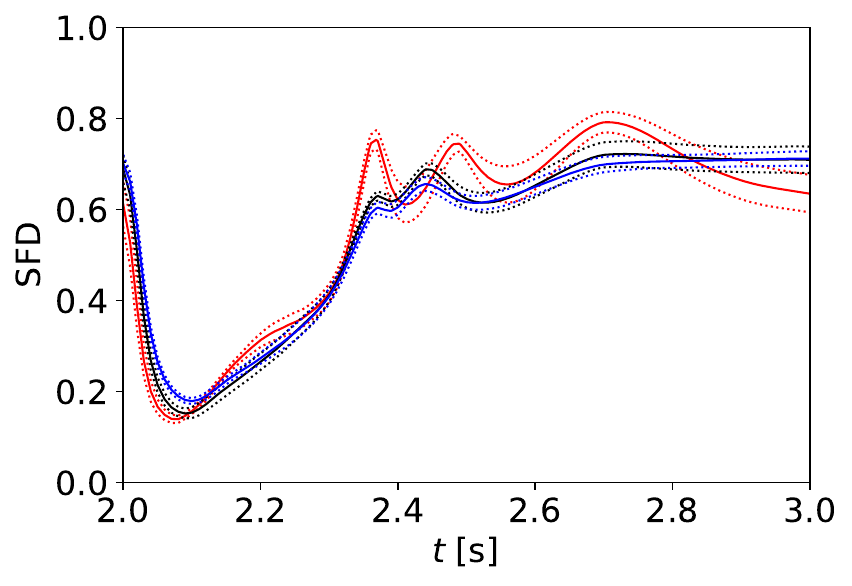}
  \includegraphics[width=0.23\textwidth]{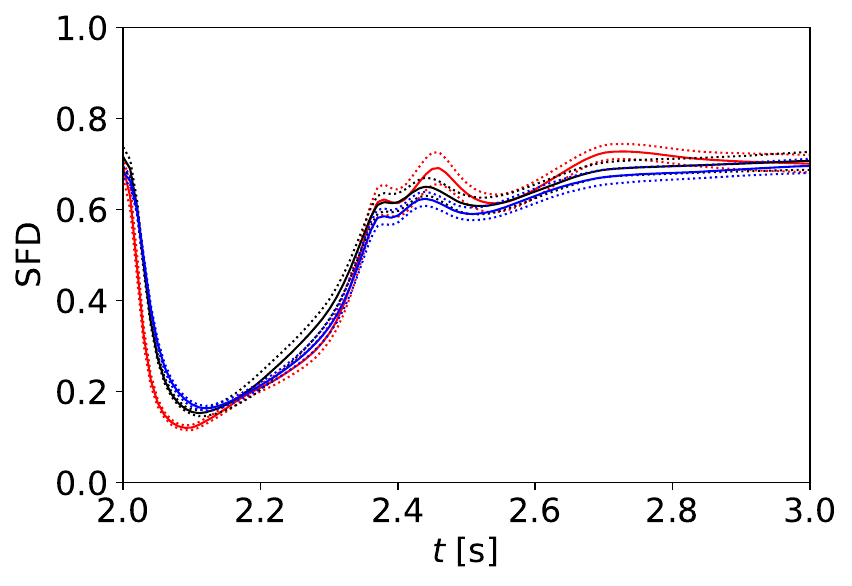}
  \includegraphics[width=0.23\textwidth]{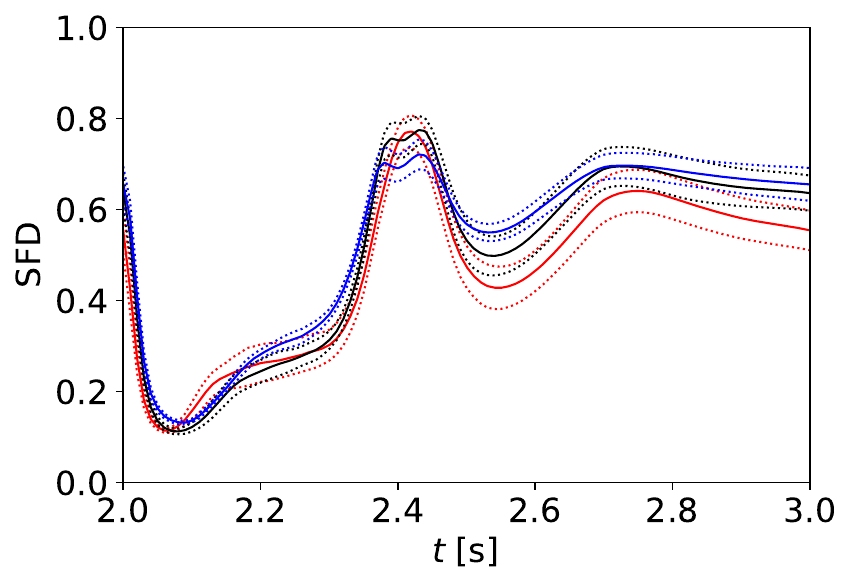}
  \includegraphics[width=0.23\textwidth]{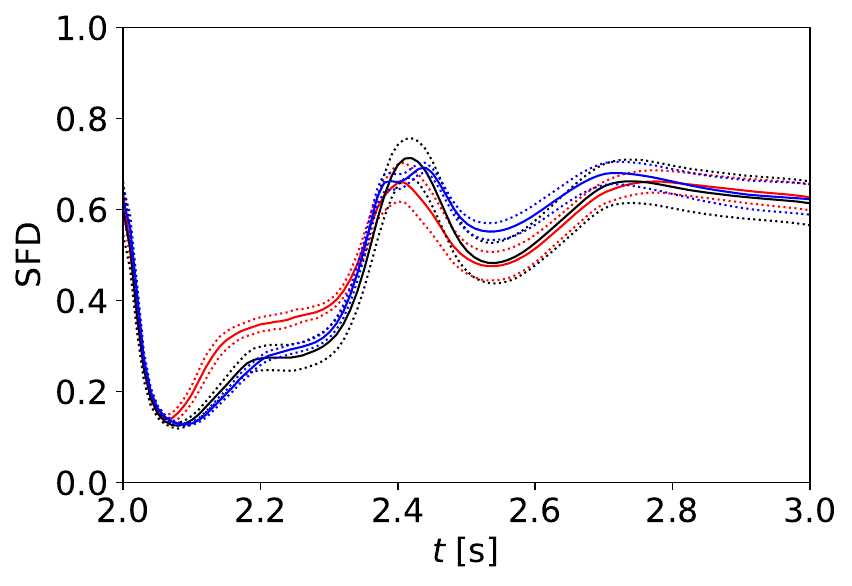} \\

  \includegraphics[width=0.23\textwidth]{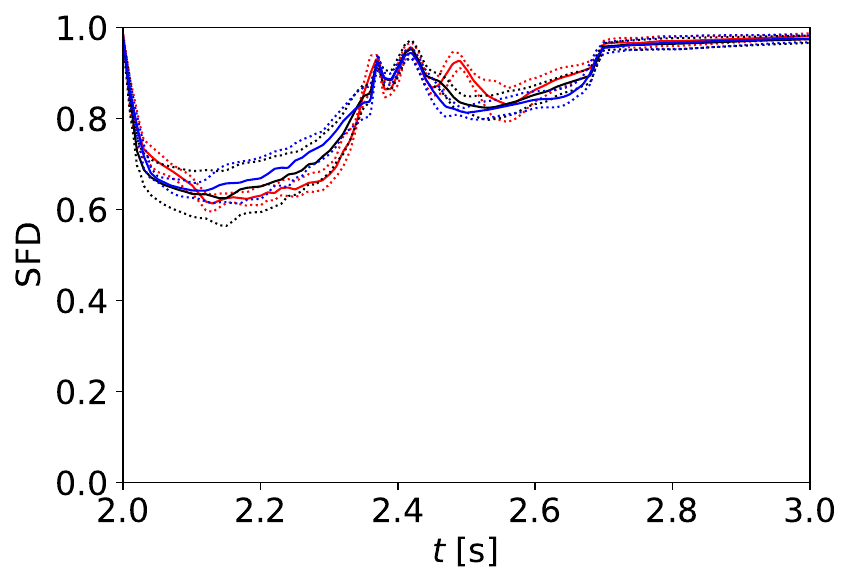}
  \includegraphics[width=0.23\textwidth]{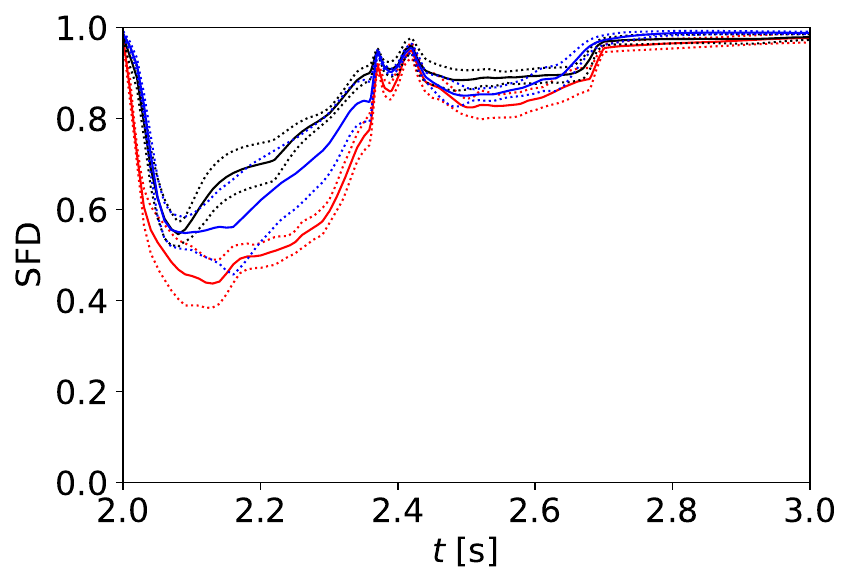}
  \includegraphics[width=0.23\textwidth]{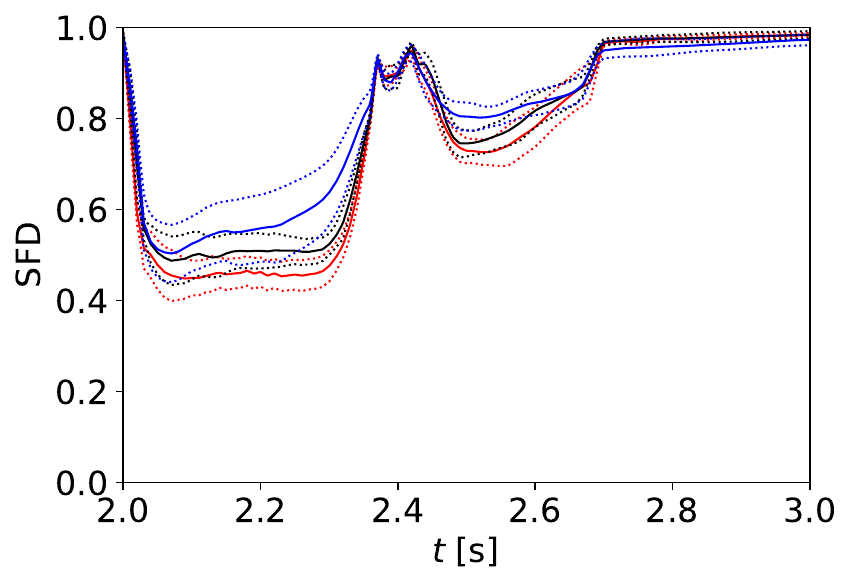}
  \includegraphics[width=0.23\textwidth]{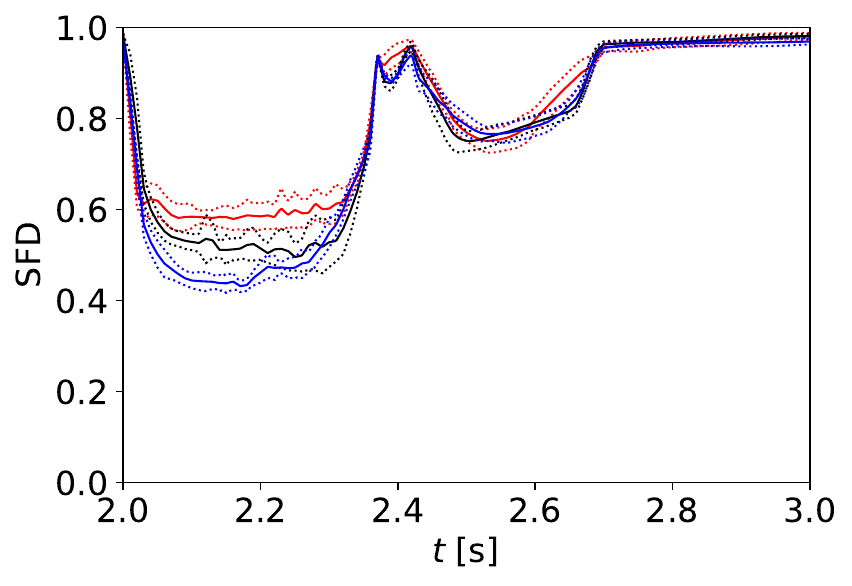}
  \caption
  {
    Top row:
    Cross-sectional tangential velocity versus normal velocity over one cardiac cycle.
    The time runs counterclockwise around these loops.
    During systole the flow accelerates and the normal flow dominates, while, during diastole, the presence of dissipating eddies might result in larger tangential velocity.
    Middle row: SFD (averaged over the length of the domain) over one cardiac cycle.
    Bottom row: SFD (maximum over the length of the domain) over one cardiac cycle.
    From left to right: batches 0, 10, 20, and 29.
    \legendnote
  }
  \label{fig:loop_norm_tang}
\end{figure}

The sensitivity of overall secondary flow (normal vs. tangential velocity) is analyzed in Figure~\ref{fig:loop_norm_tang}.
As for the pressure difference, cases with reduced radius ($\alpha_r = 0.7$) are more sensitive to geometrical uncertainty.
However, the secondary flow degree (SFD) (Figure~\ref{fig:loop_norm_tang}, middle and bottom) is considerably less
sensitive to geometric variation. The clearest observation is that, as expected, normal (forward/backward) flow
dominates when flow rates are high, leading to low SFD, whereas the slower flow during diastole exhibits SFD closer to 1,
corresponding to decaying turbulence throughout the aorta.

\begin{figure}[htp!]
  \centering
  \includegraphics[width=0.7\textwidth]{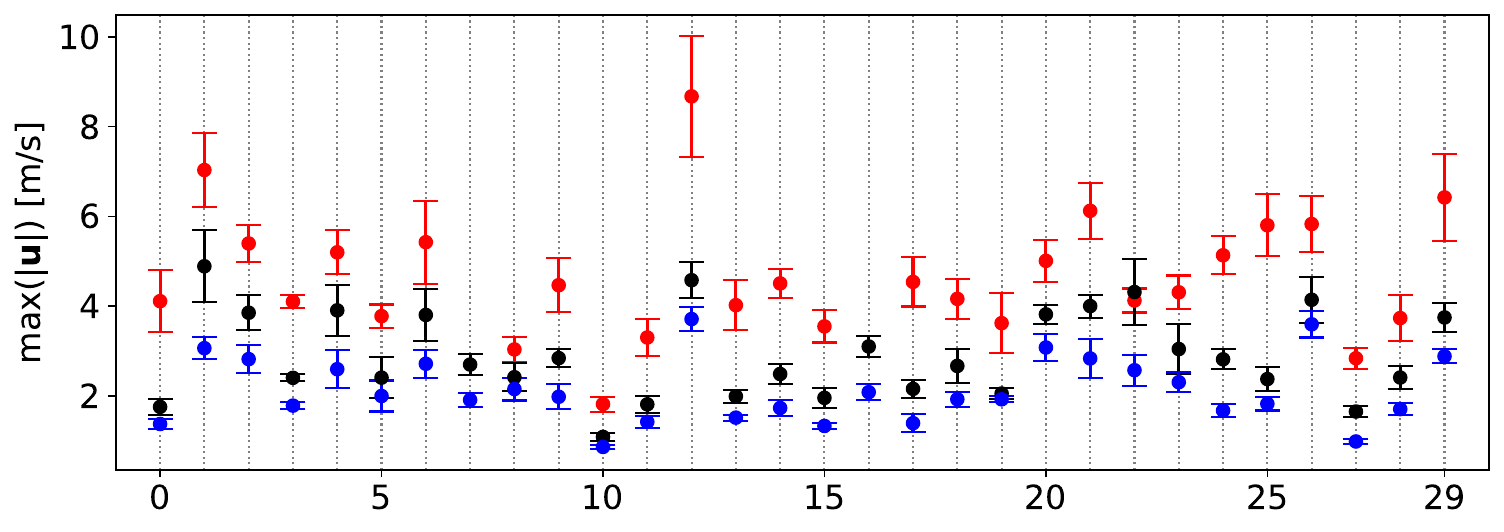}
  \includegraphics[width=0.7\textwidth]{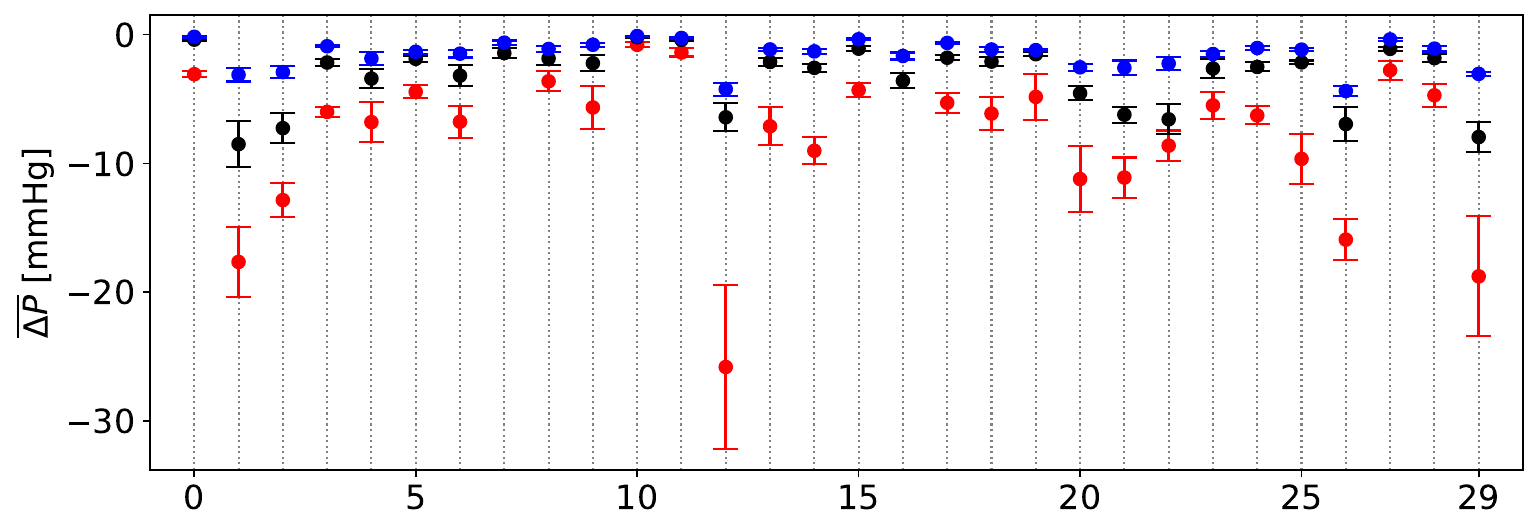}
  \includegraphics[width=0.7\textwidth]{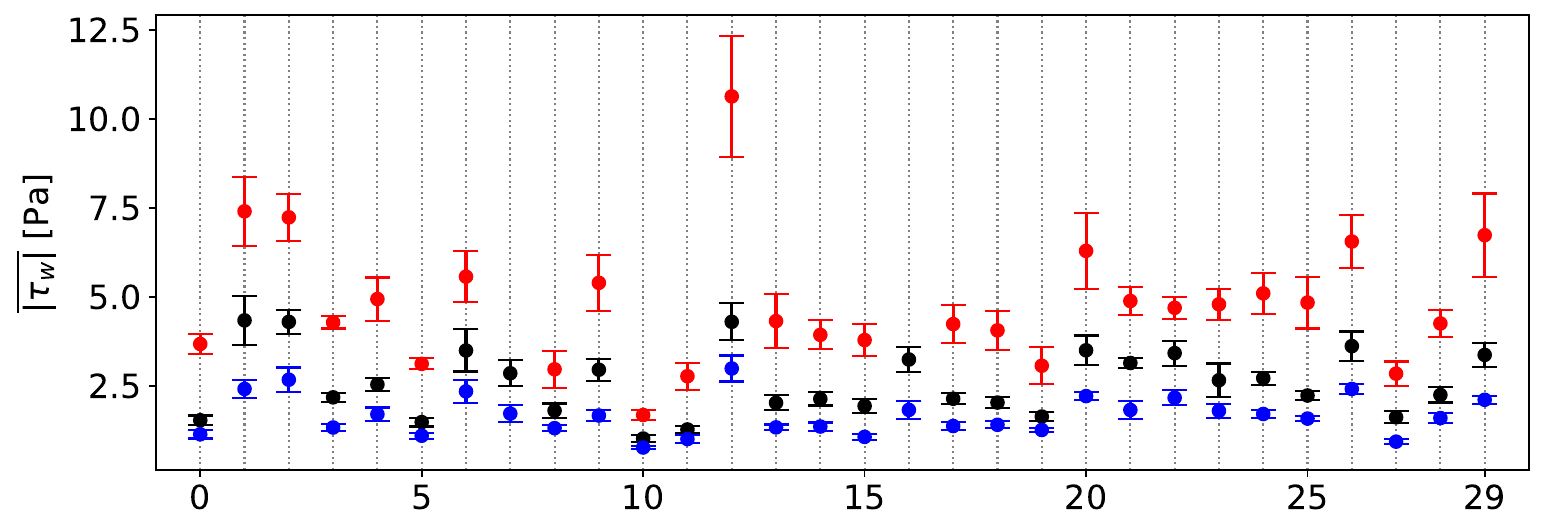}
  \includegraphics[width=0.7\textwidth]{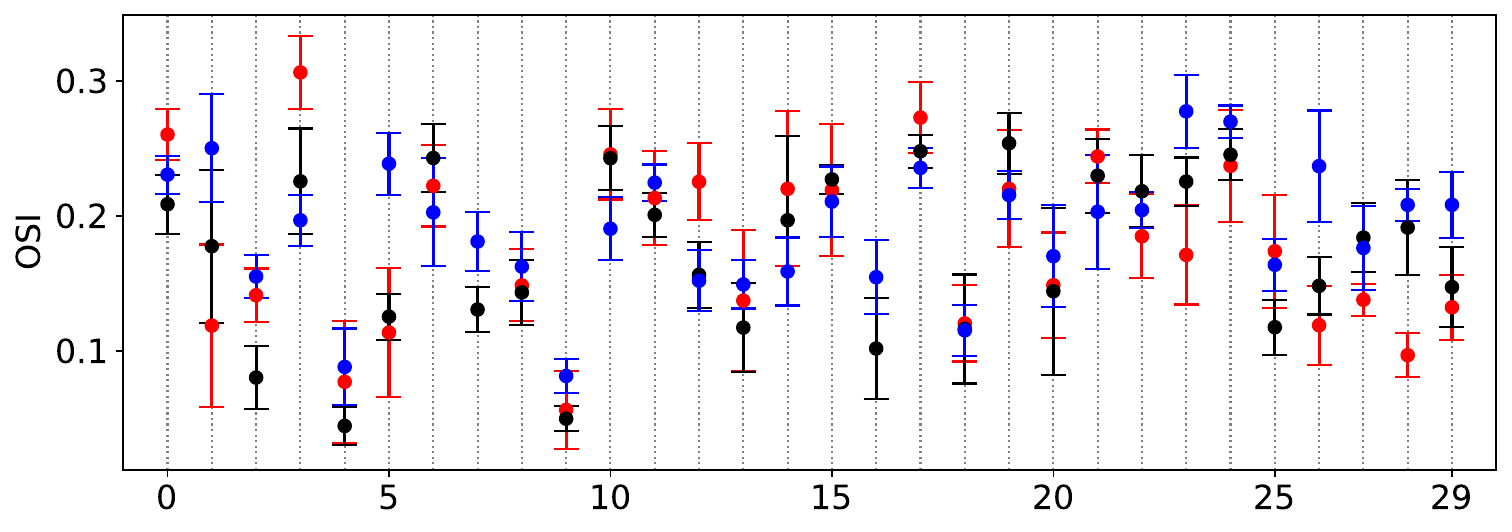}
  \caption{
  Sample means and standard deviations for different quantities over the whole dataset of geometries, divided according to the corresponding batch
   (from 0 to 29, in the $x$-axis).
  First row: maximum velocity magnitude at peak systole ($t = 2.16$s);
  second row: largest pressure difference between inlet and descending aorta over the cardiac cycle;
  third row: mean wall shear stress magnitude over the whole blood vessel wall and over the cardiac cycle;
  Fourth row: mean oscillatory shear index over a reference patch on the underside of the descending aorta.
  \legendnote }\label{fig:tables}
\end{figure}

Figure \ref{fig:tables} shows sample means and standard deviations for velocity, pressure, wall shear stress, and OSI.
Additional tables with detailed statistics are provided also in appendix \ref{sec:sim_details}.

%
%
In most cases, one can observe a correlation between geometries with smaller vessel radius and larger and more sensitive
velocities and pressures.
For one batch (number 12), we obtain rather different results. This geometry is smaller than the others (see Figure~\ref{fig:all_test_geometries}), and this behavior could be
a consequence of having scaled the boundary conditions (inlet flow and outlet lumped parameters) without taking into account the overall volume.
A different behavior can be seen for the OSI. In this case, sensitivities are comparable across batches and radius variation.
Considering the geometry variability of the dataset (see, e.g., the geometries in Figures~\ref{fig:shape_examples_batches}), the results suggest that, although
the wall shear stress sensitivity varies across the batches, its directional variability and the related sensitivity are comparable.
Detailed statistics on the biomarkers are provided in Appendix~\ref{sec:sim_details}.

\subsection{Sample size study}
\label{sec:shape_uq_convergence}
The numerical study in Section~\ref{sec:shape_uq_numerical_experiments} prioritized the variability of the geometry within the test dataset, with the limitation of considering
only a limited sample (a maximum of $n_\mathrm{s} = 10$ simulations for each individual batch). In this section, we study the
influence of considering, for a single batch (batch 3, with $\alpha_r = 1$), a larger sample with an additional $n_\mathrm{s}^\prime = 100$ geometries.

We compare the resulting statistics with those obtained with the smaller sample size in terms of distributions of quantities of interest, as well as in terms
of the Wasserstein--1 distance defined, for two one--dimensional real--valued random variables $X_1$ and $X_2$, as
\begin{equation}\label{eq:w1}
\mathcal W_1 (X_1, X_2) := \int_{\mathbb R} \left| F_1(x) - F_2(x) \right| \mathrm d x,
\end{equation}
depending on the respective cumulative distribution functions $F_1$ and $F_2$.

Note that, by a change of variables, one can write
  \[
    \mathcal W_1 (cX_1, cX_2) = c \mathcal W_1 (X_1, X_2)
  \]
  for any $c > 0$. If $X_i$ are concentrated at $x_i$ (i.e., $F_i$ are step functions), one obtains
  $\mathcal W_1 (X_1, X_2) = |x_1 - x_2|$. Hence, if the random variables $X_i$ represent two distributions of a
  physical field in a given unit, $\mathcal W_1 (X_1, X_2)$ is interpretable as a quantity in the same unit.

\begin{figure}[h!]
  \includegraphics[width=.4\textwidth]{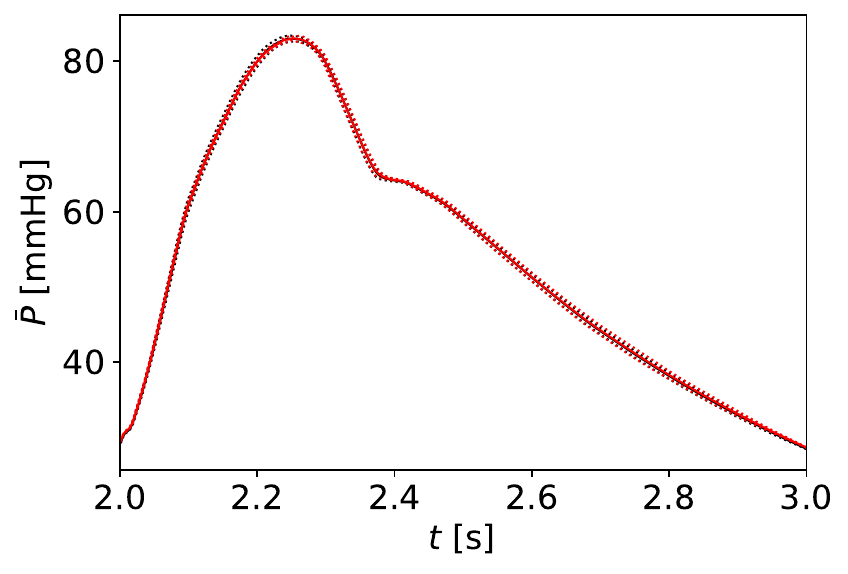}
  \includegraphics[width=.4\textwidth]{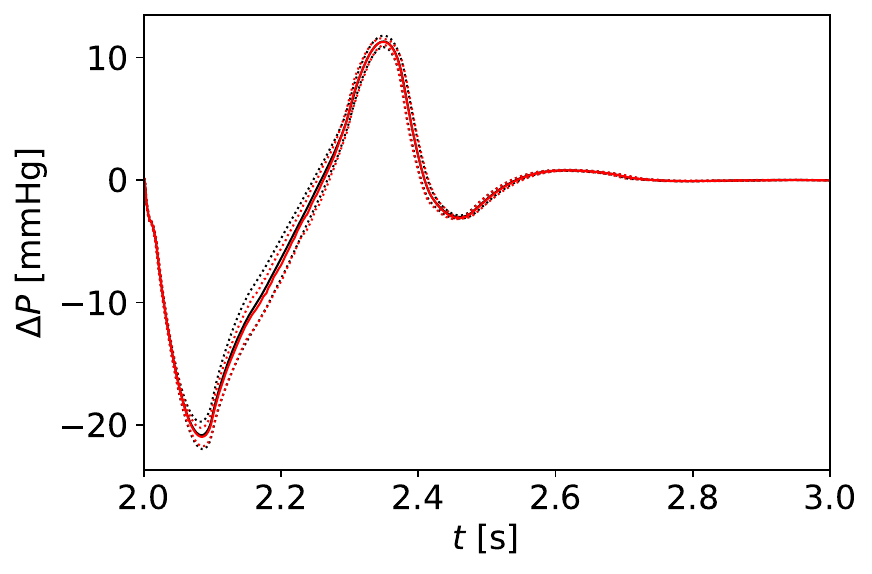}\\
    \includegraphics[width=.4\textwidth]{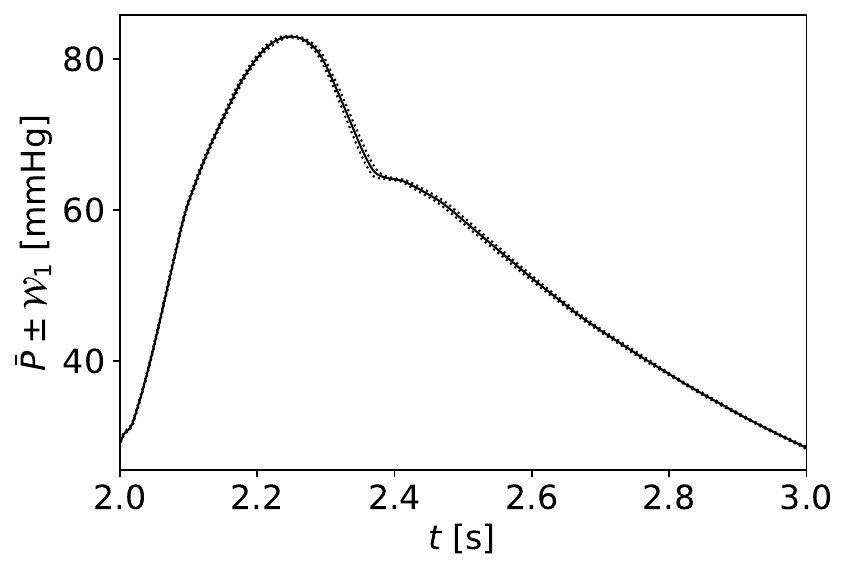}
  \includegraphics[width=.4\textwidth]{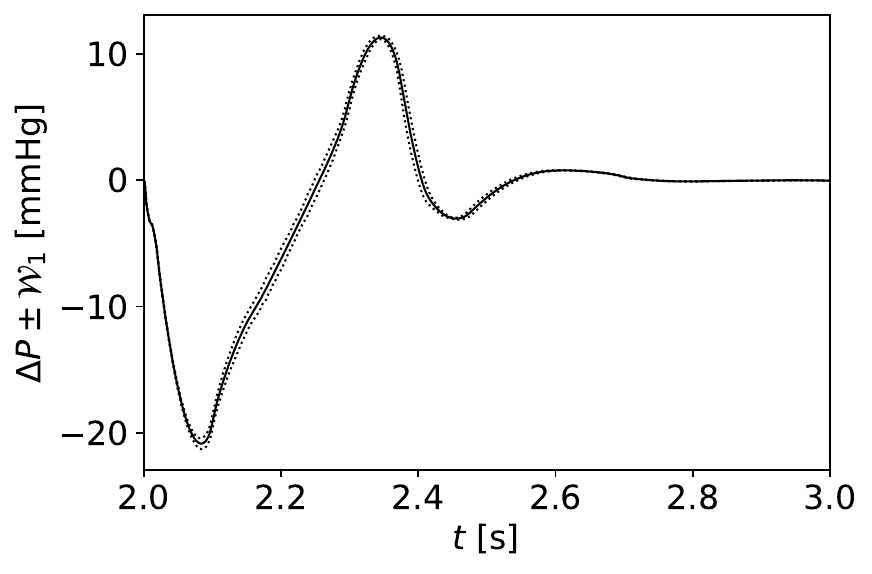}
  \caption
  {
    Top row: Distribution (mean and standard deviation) of mean pressure in the aorta (left)
    and pressure difference between descending aorta and inlet (right) over time. \compnote\
    Bottom row: Joint mean of the separate distributions and Wasserstein--1 distance between them.
  }
  \label{fig:comp_time_P}
\end{figure}

\begin{figure}[h!]
  \includegraphics[width=.4\textwidth]{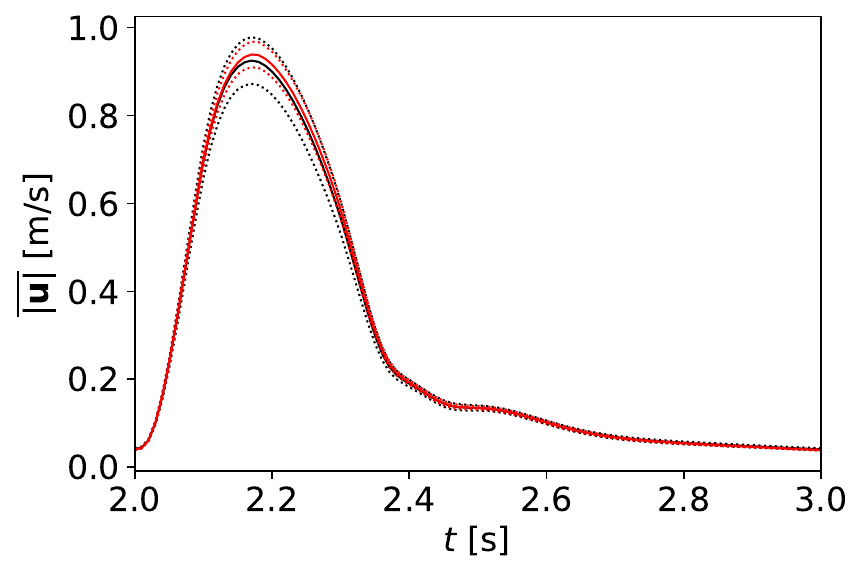}
  \includegraphics[width=.4\textwidth]{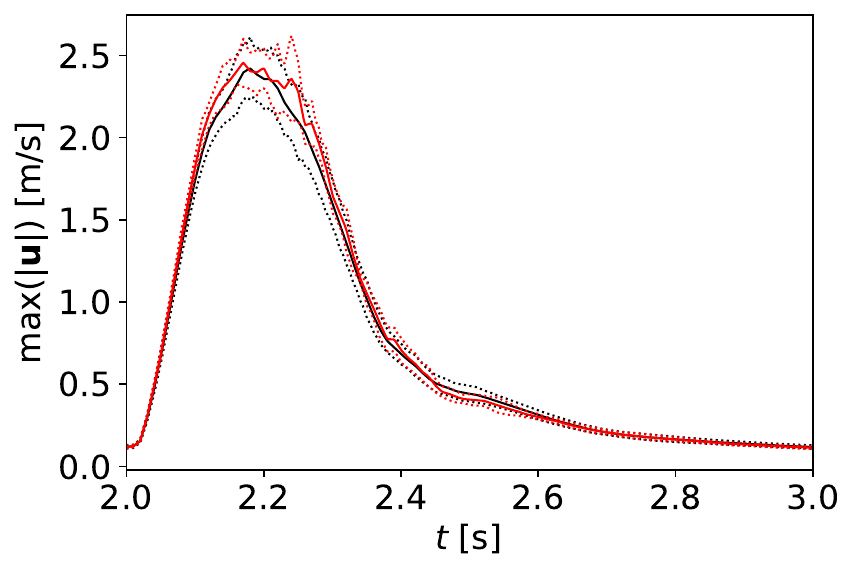}\\
    \includegraphics[width=.4\textwidth]{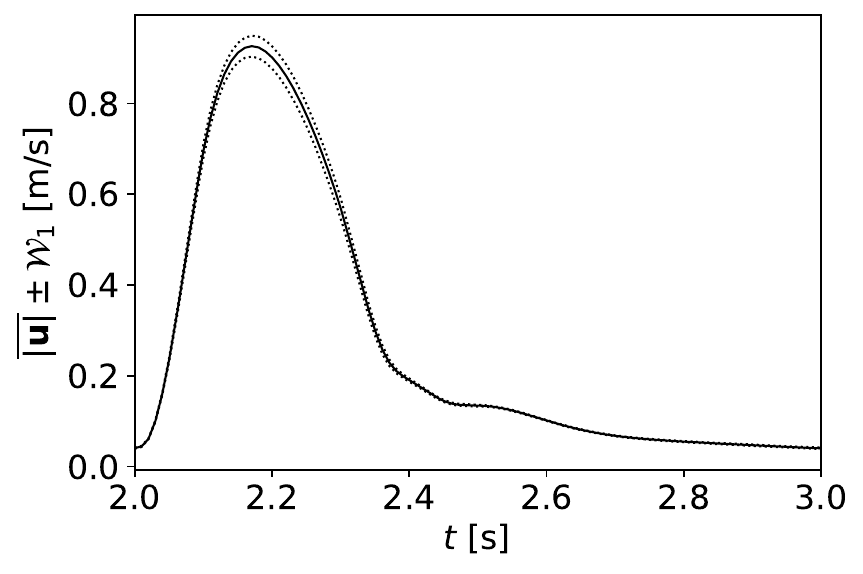}
  \includegraphics[width=.4\textwidth]{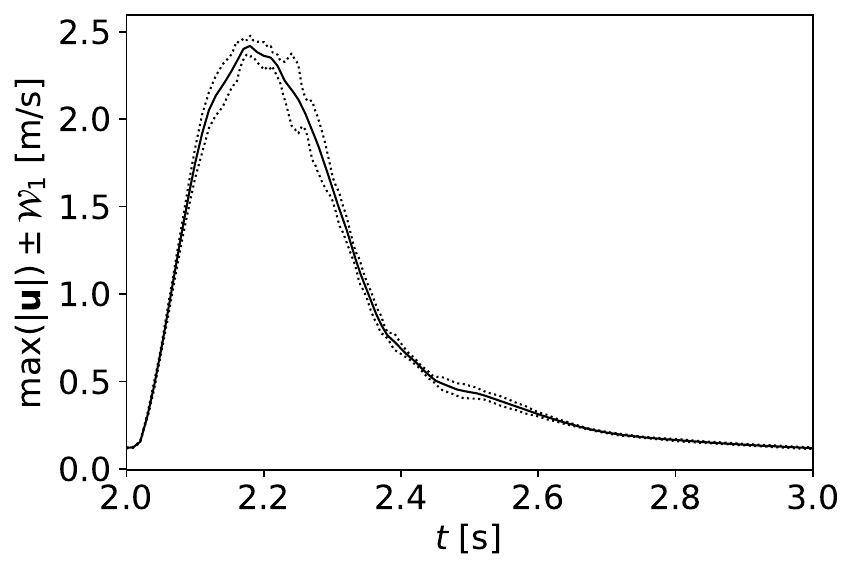}
  \caption
  {
    Top row: Distribution (mean and standard deviation) of mean cross--sectional velocity magnitude (left)
    and overall maximum velocity magnitude (right)
    over time. \compnote\
    Bottom row: Joint mean of the separate distributions and Wasserstein--1 distance between them.
  }
  \label{fig:comp_time_u}
\end{figure}

Figure~\ref{fig:comp_time_P} shows the distributions of mean pressure and pressure differences for the two samples, showing that the statistics obtained in both
cases are comparable. Quantifying this by the Wasserstein--1 distance (bottom row) does not reveal any surprises --- the
distributions are clearly very close compared to the overall scale of the pressure.
Similar conclusions can be drawn for the mean and maximum velocity magnitude (along the centerline and over the whole
domain), Figure~\ref{fig:comp_time_u}, although here some elevated sensitivity at peak flow can be observed.

\begin{figure}[h!]
  \includegraphics[width=.3\textwidth]{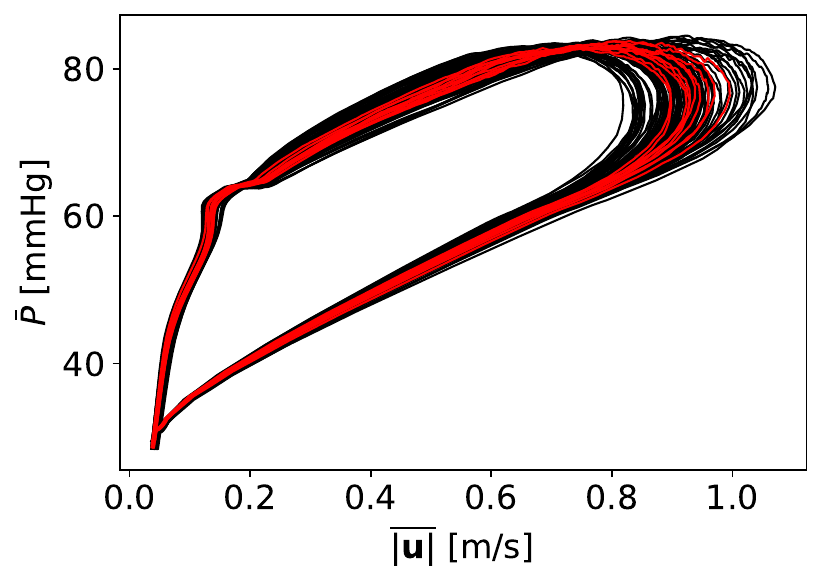}
  \includegraphics[width=.3\textwidth]{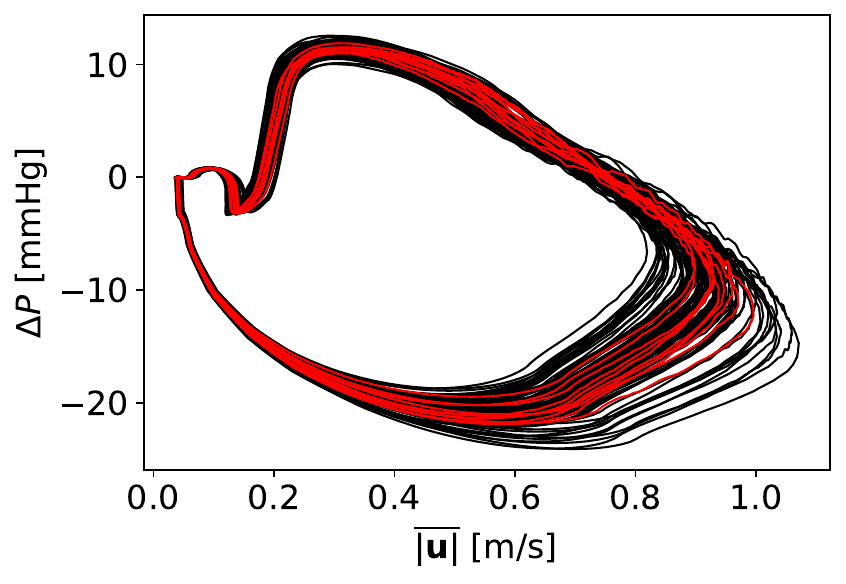}\\
  \includegraphics[width=.3\textwidth]{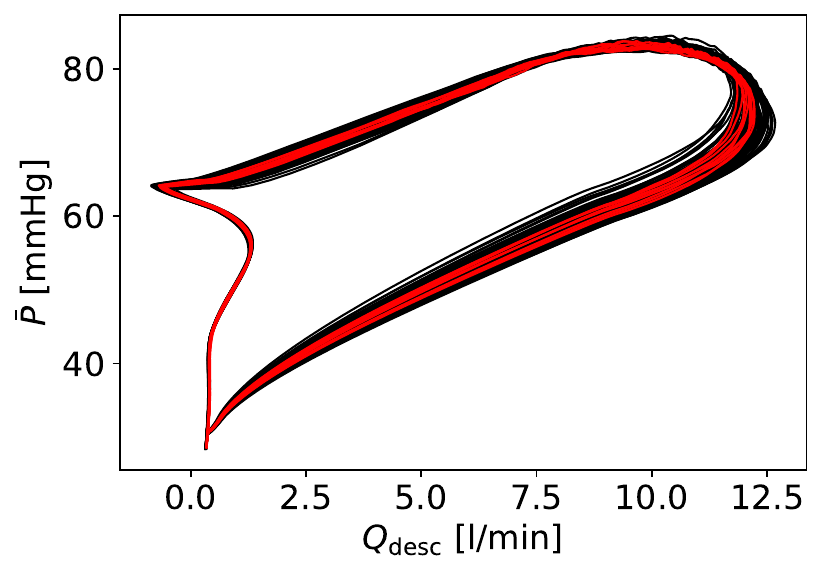}
  \includegraphics[width=.3\textwidth]{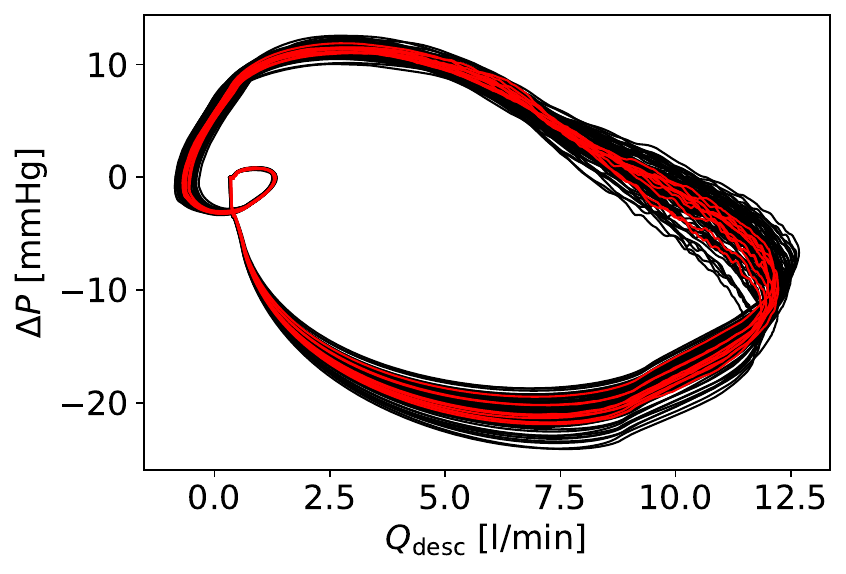}
  \caption
  {
    Left: Mean pressure in the aorta versus
    mean cross--sectional velocity magnitude, averaged over the length of the domain (top) and outflow through the descending aorta (bottom),
    over one cardiac cycle.
    Right:  Pressure difference between descending aorta and inlet versus
    mean cross--sectional velocity magnitude, averaged over the length of the domain (top) and outflow through the descending aorta (bottom),
    over one cardiac cycle.    \compnote
  }
  \label{fig:comp_loop_u_Q_p}
\end{figure}

Figures~\ref{fig:comp_loop_u_Q_p} compare pressure and velocity quantities of interest over a full cycle. Also in this case, the samples of different sizes behave very similarly.
A similar agreement between the two cases was also observed for the remaining quantities of interest.

\begin{figure}[h!]
  \includegraphics[width=.3\textwidth]{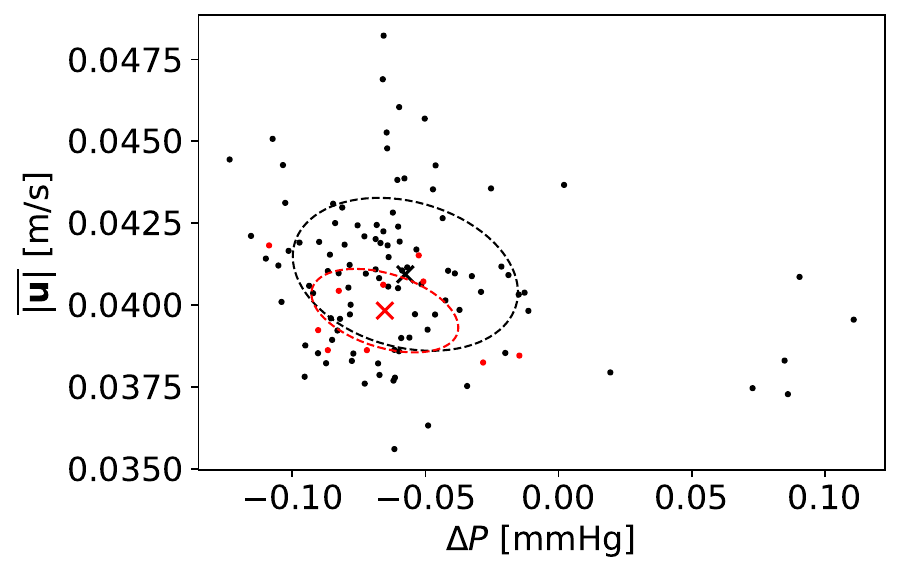}
  \includegraphics[width=.3\textwidth]{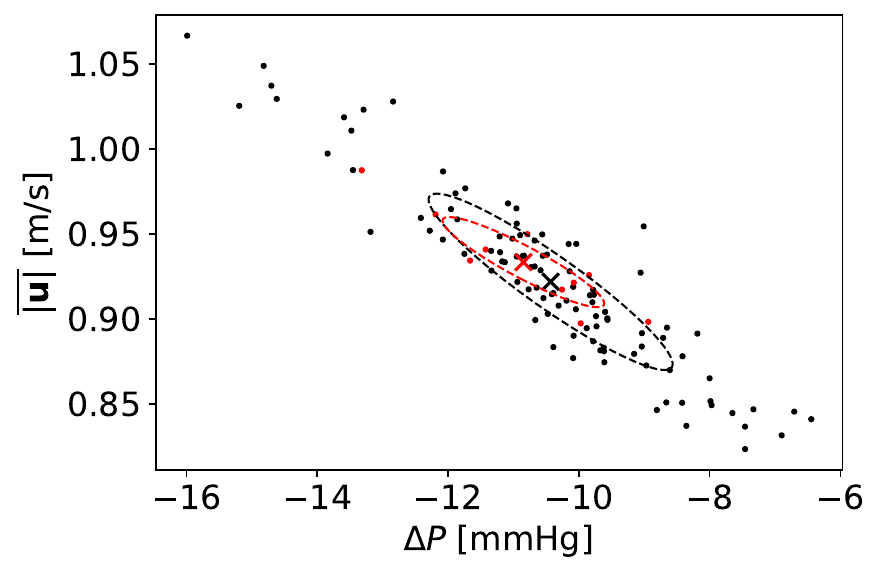}\\
  \includegraphics[width=.3\textwidth]{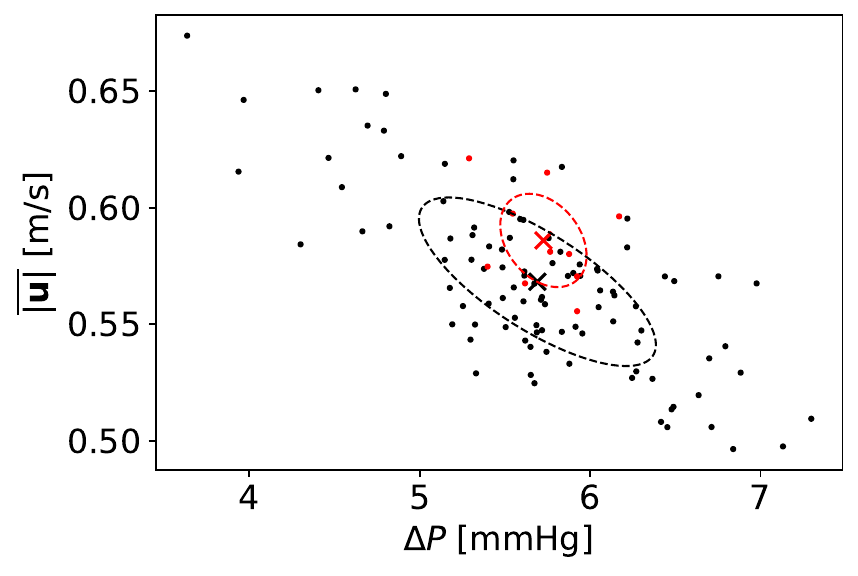}
  \includegraphics[width=.3\textwidth]{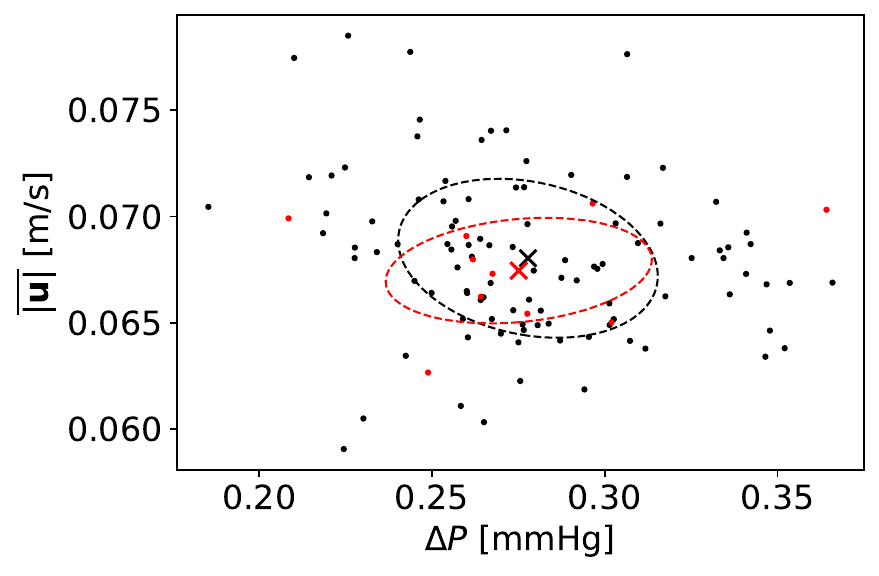}
  \caption
  {
    Pressure difference between descending aorta and inlet versus mean cross--sectional velocity magnitude
    (averaged over the length of the domain) for all simulations at times $t \in \{ 2, 2.16, 2.3, 2.7 \}$,
    corresponding to the end of diastole (top-left), peak flow (top-right), end of systole (bottom-left), and middle diastole (bottom-right).
    \compnote
    Crosses and ellipses indicate sample means and standard deviations.
  }
  \label{fig:comp_scatter_u_p}
\end{figure}

Analyzing the results at selected time steps (Figure~\ref{fig:comp_scatter_u_p}), one can see that, as expected, the larger sample allows us to capture a broader
range of physiological conditions.

\clearpage
\section{Conclusions}
\label{sec:conclusions}

We propose a new approach for shape generative modeling based on the combination of stochastic
interpolants and LDDMM registration. The method leverages pairwise diffeomorphic maps computed
via multilevel ResNet-LDDMM shape registration to obtain the time-dependent conditioned drift on
intermediate domains required for training the stochastic interpolants, allowing, in particular, the bridging of
distributions defined on different three-dimensional domains.

We validated the framework by considering a cohort of synthetic aortic geometries, generating variations of
the available shapes to perform domain uncertainty quantification studies for blood flow simulations,
monitoring different quantities of interest (such as velocity, pressure differences, wall shear stresses, and OSI).
To this end, the generative model was used to generate different batches of perturbed domains
starting from 30 test geometries, considering both increases and decreases in vessel radius. Our results show
that the sensitivity with respect to domain changes is considerably higher for geometries with reduced
lumen size. The pattern differs only in the case of OSI, suggesting that the variability of the
stress vector is more affected by variations in the surface shape rather than by the diameter of the
vessel.
For this application, we considered a conditioning variable related to a latent space representation of
the aortic shapes, i.e., a given number of centerline points and the corresponding radii of the inscribed
spheres at each point. This is a natural choice for this particular setting, and it is also consistent
with previous studies that consider the generation of digital shapes using statistical shape models (SSMs).
However, the framework can be adapted to handle different latent descriptions.
For the presented numerical simulations, the Navier–Stokes equations are solved via a high-order,
discontinuous Galerkin matrix-free solver which relies on hexahedral volume meshes. While the choice of the
numerical method is independent from the generative model, it is extremely important, from the point
of view of the application, to be able to rely on a performant solver, particularly for uncertainty
quantification studies requiring multiple simulations. Hence, while prescribing the surface deformation can be
achieved via the generative model, it is necessary to preserve the internal mesh structure that guarantees
solver performance. To this end, we propose an iterative methodology for the deformation of the
volume mesh based on the solution of a fictitious elasticity problem, combined with an optimization step
for controlling the mesh aspect ratio. These preliminary results have been obtained by considering only a rather limited sample of geometries
and conditioning variables. To address this issue, we performed numerical simulations using a larger
sample (100 geometries instead of 10) for one particular test geometry, showing good agreement between
the statistics in both cases.

A limitation of the proposed approach is that it does not allow for the study of local domain uncertainty, as shown in Figure~\ref{fig:interpolation} (right). This limitation could be addressed by enriching the training dataset with local deformations. Imposing metric constraints would improve the correspondence between the latent variables associated with centerline points and radii, and the actual centerline points and radii of the generated shapes. We remark that, although in the case of the aorta a natural encoding via the centerline is available, the methodology can nonetheless be extended to truly nonparametric shapes, such as general anatomical organs, provided they belong to the same class of bi-Lipschitz homeomorphisms. In this regard, the current LDDMM stochastic interpolant can be extended to generate shapes in different classes of bi-Lipschitz homeomorphisms by considering multiple template shapes~\cite{KONG2024103293}. These extensions will be explored in future work.

Current and future research, based on the results of this work, includes the application of this approach to different clinical contexts (different organs and pathological conditions, simulation of devices and treatments) for tasks related not only to uncertainty quantification, but also to virtual patient generation for in silico trials, and shape optimization.
Current and future research, based on the results of this work, includes the application of this approach to different clinical contexts (different organs and pathological conditions, simulation of devices and treatments) for tasks related not only to uncertainty quantification, but also to virtual patient generation for in silico trials, and shape optimization.

\section*{Acknowledgements}
Funded by the Deutsche Forschungsgemeinschaft (DFG, German Research
Foundation) under Germany´s Excellence Strategy – The Berlin Mathematics
Research Center MATH+ (EXC-2046/1, EXC-2046/2, project ID: 390685689).

\printbibliography
\newpage
\appendix
\section{Training of the conditional LDDMM stochastic interpolant}
\label{sec:training_details}

To approximate the conditioned drift $b_t^{\theta}[\bc]$ we employ a graph neural network (GNN), whose architecture is a MeshGraphNet  implemented in the open-source framework {\tt physicsnemo}~\cite{physicsnemo2025}.

The processor size is $10$, and the hidden dimension of the node and edge encoder and decoder is $128$. 
The input layer is defined by the template hexahedral mesh, whose number of vertices will be denoted by $N_0$. 
Node features are a concatenation of a feature embedding of  $I_t\in\mathbb{R}^{N_0\times 3}$, considering,
for $n_f = 64$ Fourier features,
\begin{itemize}
\item a feature embedding $\varphi_{I_t}:\mathbb{R}^{N_0\times 3}\rightarrow \mathbb{R}^{N_0\times 3\cdot n_{\text{f}}\cdot 2}$, defined as
\begin{equation}
\varphi_{I_t}(I_t) = \left[\cos(\mathbf{v}\otimes I_t), \sin(\mathbf{v}\otimes I_t)\right],
\end{equation}
where $\otimes$ is the Kronecker product of $I_t\in\mathbb{R}^{N_0\times 3}$ with $\mathbf{v}=\{1,2,\hdots, 2^{n_{\text{f}}-1}\}\in\mathbb{R}^{n_{\text{f}}}$;
\item a time sinusoidal encoding $\varphi_t:\mathbb{R}\rightarrow\mathbb{R}^{3\cdot n_{\text{f}}\cdot 2}$ defined as $\varphi_t=\varphi^2_t\circ \varphi^1_t$, where
\begin{equation}
\varphi^1_t:\mathbb{R}\to\mathbb{R}^{d}, \; \varphi^1_t(t) = \left[\sin(\omega_0 t), \cos(\omega_0 t), \dots, \sin(\omega_{n_{\text{f}}/2-1} t), \cos(\omega_{n_{\text{f}}/2-1} t)\right],
\end{equation}
where the frequencies are given by $\omega_i = \exp\left(-\ln(10000) \cdot \frac{2i}{d}\right)$, for $i=0, \hdots, d/2-1$, with
$d=94\cdot 4$ being the embedding dimension, and 
$\varphi^2_t:\mathbb{R}^{d}\rightarrow\mathbb{R}^{3\cdot n_{\text{f}}\cdot 2}$ 
is a FNN architecture with $1$ hidden layer of dimension $128$ and \textit{ReLU} activation, while the final activation is a \textit{tanh} activation;
\item an additional feed-forward neural  network $\varphi_{\bc}:\mathbb{R}^{d}\rightarrow\mathbb{R}^{3\cdot n_{\text{f}}\cdot 2}$  with 
the same structure as $\varphi^2_t$.
\end{itemize}

%
For defining the underlying graph structure, instead of the mesh edges, we consider each node to be connected to its $k=12$ nearest neighbors.
This choice improves the information propagation in the GNN.

The training is performed for $2750$ epochs on $1209$ training data with a batch size $n_{\text{b}}=4$, using the Adam optimizer with learning rate $10^{-4}$. A scheduler that reduces the learning rate by a factor of $0.5$ every time there is a plateau of $100$ epochs in the training error is employed.

A sigma noise schedule~\cite{karras2022elucidating} is employed for the diffusion coefficient $\sigma_t$: $\rho=1$ and the initial and final values are set to $\sigma_{\text{max}}=0.002$ and $\sigma_{\text{min}}=0.001$, respectively
\begin{equation*}
\sigma_t = \left(\sigma_{\text{max}}^{1/\rho} + t(\sigma_{\text{min}}^{1/\rho} - \sigma_{\text{max}}^{1/\rho})\right)^{\rho}.
\end{equation*}

\begin{figure}[htp!]
  \includegraphics[width=0.6\textwidth]{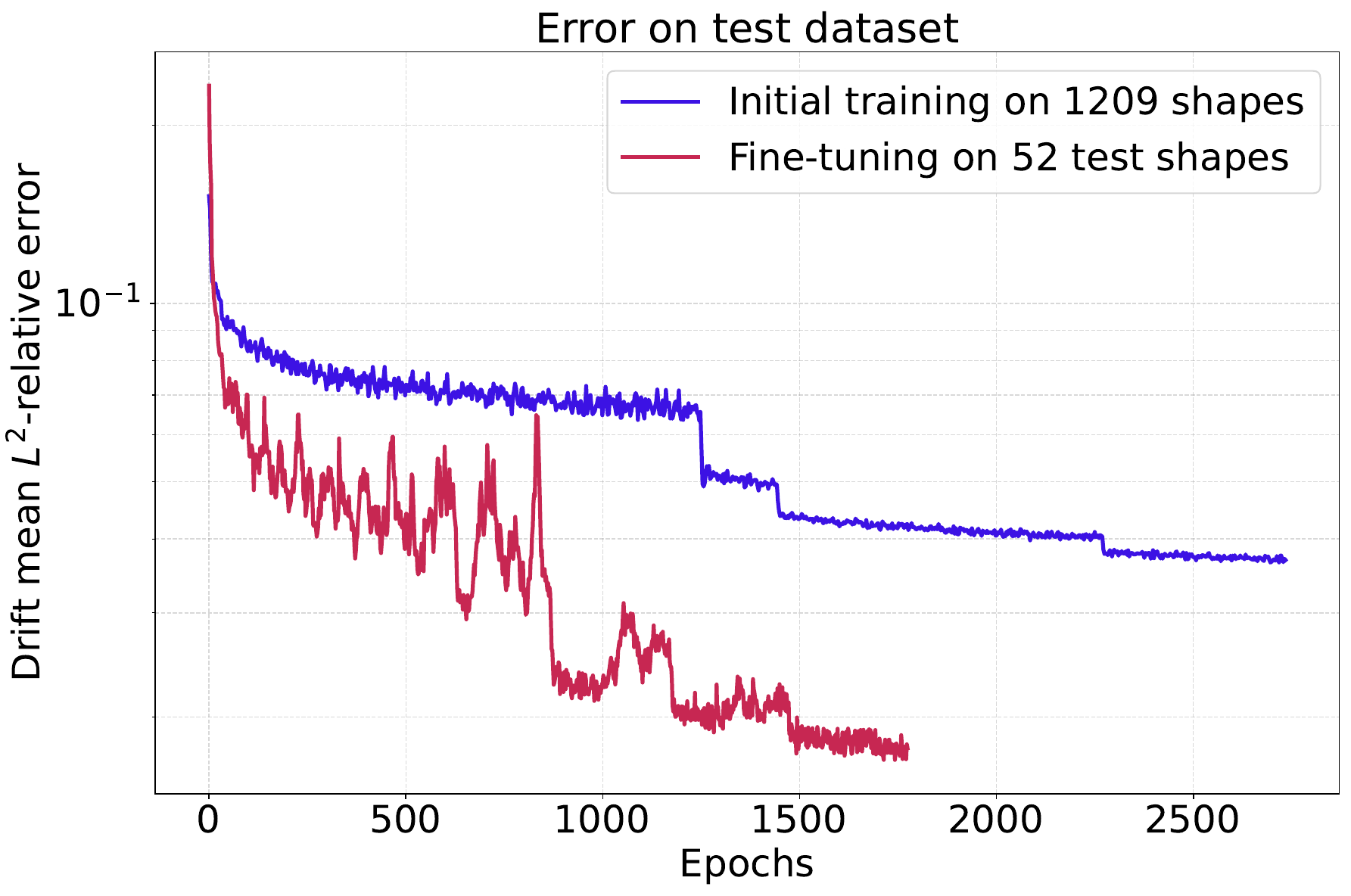}
  \caption{$L^2$-relative errors on the drift during training (blue curve) and fine-tuning (red curve) phases, evaluated only on the test set of $30$ shapes.}
  \label{fig:training_lddmm}
\end{figure}

\begin{figure}[htp!]
  \includegraphics[width=0.75\textwidth]{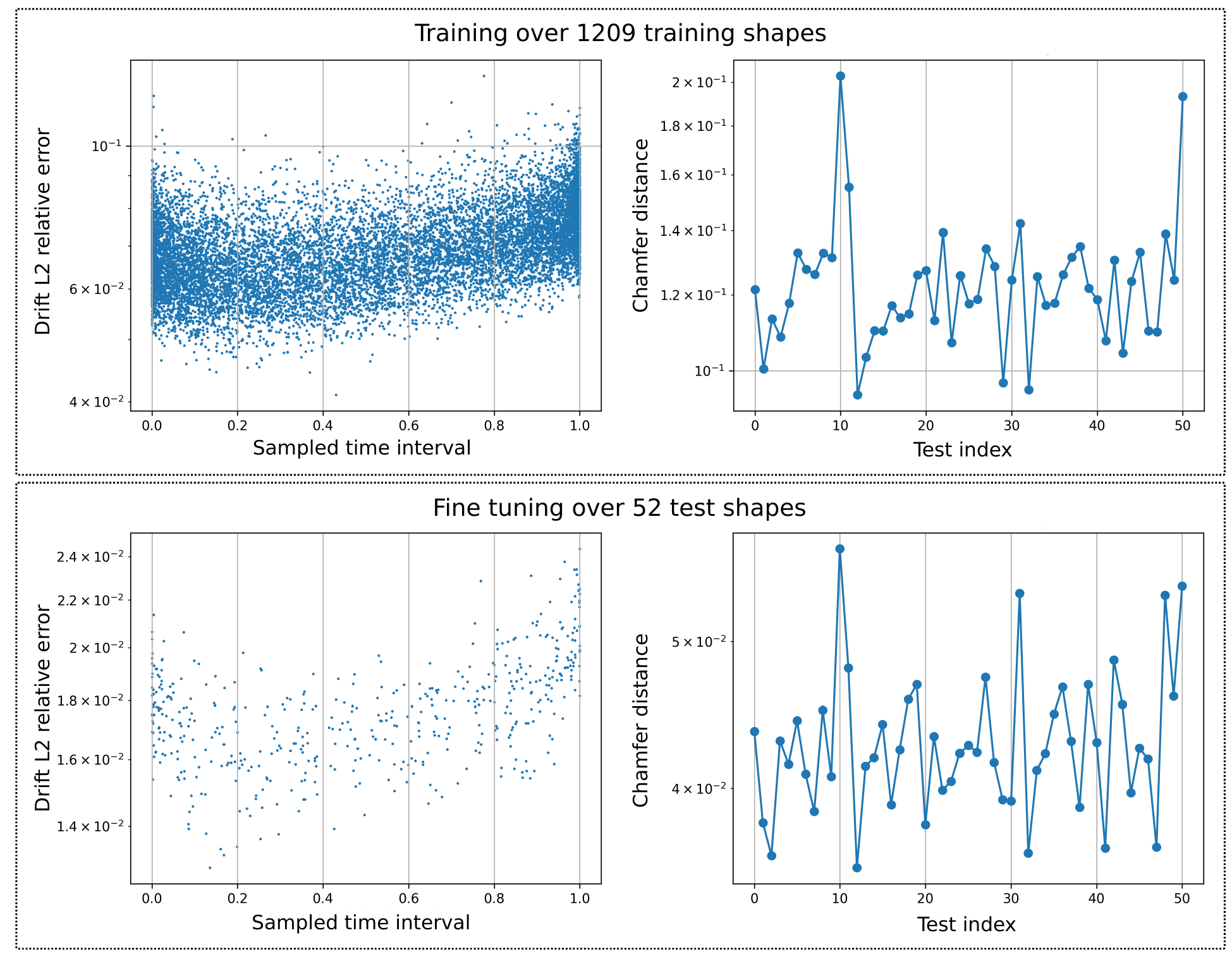}
  \caption{Left: Drift $L^2$-relative error on the test set during training and fine-tuning for different sampled time instances. Right: Chamfer distance between the target shapes and the generated shapes with the learned drift on the test set during training and fine-tuning.}
  \label{fig:chamfer_lddmm}
\end{figure}

After the training a fine-tuning phase is performed on the $30$ test shapes for $1750$ epochs with the same training parameters. 
The $L^2$-relative errors on the drift evaluated on the test set of $30$ shapes during the fine-tuning phases are shown in Figure~\ref{fig:training_lddmm}. 
Figure~\ref{fig:chamfer_lddmm} shows the $L^2$-relative error during training and fine-tuning phases for different sampled time instances, as well 
as the Chamfer distance between the target shapes and the generated shapes with the learned drift.

\newcommand{\arapprox}[1]{\widetilde{\mathrm{Asp}}(#1;\Psi)}
\section{Smoothing algorithms for mesh aspect ratio optimization}
\label{sec:mesh_smoothing}
Let $\Psi$ denote a volume displacement field obtained by solving the iterative linear elastic problem \eqref{eq:intermediate elastic problem-psi} (section \ref{subsec:mesh_transport}).
For each (hexahedral) mesh cell $K$ after mesh deformation, we consider the trilinear reference map
\[
  F_{K;\Psi}(\bx): [0,1]^3 \to K.
\]
defined by mapping the vertices of a reference cube to the vertices of $K$.
Under the assumption that problem \eqref{eq:intermediate elastic problem-psi} was solved successfully,
the Jacobian $J = \nabla F_{K;\Psi}(\bx) \in \mathbb R^{3 \times 3}$
is well-defined and non-singular for each point $\bx \in [0,1]^3$.
In particular, since the deformation is assumed to preserve cell orientation, the Jacobian has three positive singular values
$0 < \sigma_1(\bx) \leq \sigma_2(\bx) \leq  \sigma_3(\bx)$ at each point $\bx \in [0,1]^3$.

We define the aspect ratio of the cell $K$ under the deformation $\Psi$ as the maximum of the pointwise ratio
between the largest and the smallest singular value of $J$:
\begin{equation}\label{eq:aspect-ratio}
  \mathrm{Asp}(K;\Psi) = \max_{\bx \in [0,1]^3} \frac {\sigma_3(\bx)} {\sigma_1(\bx)}.
\end{equation}

\begin{remark}
The aspect ratio \eqref{eq:aspect-ratio} can be seen as the deviation of the shape of $K$ from a cube. Indeed,
$\mathrm{Asp}(K;\Psi)\geq 1$, and the minimum value of $1$ corresponds to the case of all equal singular values, at each point $\bx \in [0,1]^3$.
\end{remark}

In practice, we consider an approximated aspect ratio
\begin{equation}\label{eq:aspect-ratio-app}
\widetilde{\mathrm{Asp}}(K;\Psi) = \max_{\bx \in S_\mathrm{asp}} \frac {\sigma_3(\bx)} {\sigma_1(\bx)},
\end{equation}
using uniformly spaced points $S_\mathrm{asp} = \left\{ 0, \frac 1 4, \frac 1 2, \frac 3 4, 1 \right\}^3$.


Let us now define, for a given displacement field, the set
\[
  \mathcal I(\Psi) = \left\{ K : \arapprox{K} > \frac{3}{4} \max_{K^\prime} \arapprox{K^\prime} \right\}
\]
of \textit{bad cells}, i.e., of mesh cells whose approximated aspect ratio is larger than 75\% of the maximum value over the whole domain.
We consider the optimization problem
\begin{equation}\label{eq:psi-smoothing}
\Psi^* = \argmin_{\Psi |_{\partial \Omega} = \Psi_{\text{s}} } \mathcal{L}(\Psi) ,\quad
\mathcal{L}(\Psi) :=
\frac 1 {\# \mathcal I(\Psi)} \sum_{K \in \mathcal I(\Psi)}
\arapprox{K}^2 \,.
\end{equation}
where $\Psi_{\text{s}}$ is the prescribed deformation on the surface mesh (see equation~\eqref{eq:intermediate elastic problem-psi}).
The minimization problem \eqref{eq:psi-smoothing} has been solved with \texttt{PyTorch}  using a modified
adaptive gradient descent method with an initial value of $\Psi_{N_\text{steps}}$, stopping when a threshold of
$\mathcal{L} < 100$ was reached, which in each case corresponded to a maximum approximate aspect ratio between $10$ and $20$.

Additional changes were made to prevent the reemergence of inverted cells in the course of this procedure or convergence to poor local minima. Note also that as we are using a multigrid numerical solver, this procedure was actually performed in sequence from the coarsest to the finest level, fixing inherited coordinates at each step.

\section{Additional details on biomarker statistics}
\label{sec:sim_details}

This supplementary section details mesh variation statistics (table~\ref{tab:mesh_stats}), as well as sample means and standard deviations for
velocity (table~\ref{tab:u_216_max}), pressure (table~\ref{tab:P_rel_min}), wall shear stress (table~\ref{tab:wss_mag_mean}), 
and OSI (table~\ref{tab:osi_patch}) for all considered batches.

\begin{table}[!t]
  \caption
  {
    Mesh variation statistics. Here $\overline d_c$ is the mean pairwise chamfer
    distance over each batch, whereas $\overline \sigma$ is the root mean square
    sample standard deviation of the vertex coordinates. Both are in units of millimeters.
  }
  \label{tab:mesh_stats}
  \begin{tabular}{c|c|c||c|c|c||c|c|c}
	$\alpha_r = 0.7$ & $\overline d_c$ & $\overline \sigma$ & $\alpha_r = 1$ & $\overline d_c$ & $\overline \sigma$ & $\alpha_r = 1.3$ & $\overline d_c$ & $\overline \sigma$ \\ \hline
	0 & 1.419 & 1.79 & 0 & 1.923 & 2.902 & 0 & 2.043 & 2.875 \\
	1 & 1.561 & 2.389 & 1 & 1.871 & 2.901 & 1 & 1.712 & 2.207 \\
	2 & 1.223 & 1.694 & 2 & 1.359 & 1.938 & 2 & 1.735 & 2.791 \\
	3 & 1.302 & 1.848 & 3 & 1.483 & 1.949 & 3 & 1.602 & 1.966 \\
	4 & 1.42 & 2.031 & 4 & 1.432 & 1.943 & 4 & 1.631 & 2.674 \\
	5 & 1.444 & 2.019 & 5 & 1.904 & 3.128 & 5 & 1.571 & 2.12 \\
	6 & 1.497 & 2.058 & 6 & 1.861 & 2.602 & 6 & 2.035 & 2.974 \\
	7 & --- & --- & 7 & 1.54 & 2.494 & 7 & 1.56 & 2.131 \\
	8 & 2.041 & 3.192 & 8 & 1.634 & 2.208 & 8 & 1.672 & 2.311 \\
	9 & 1.562 & 2.38 & 9 & 1.89 & 3.07 & 9 & 1.457 & 1.815 \\
	10 & 1.773 & 2.14 & 10 & 1.894 & 2.367 & 10 & 1.626 & 1.682 \\
	11 & 1.764 & 2.615 & 11 & 1.695 & 2.15 & 11 & 1.989 & 2.51 \\
	12 & 1.592 & 2.933 & 12 & 1.568 & 2.468 & 12 & 1.8 & 2.893 \\
	13 & 2.111 & 3.639 & 13 & 1.675 & 2.583 & 13 & 1.351 & 1.641 \\
	14 & 1.547 & 2.465 & 14 & 1.597 & 2.431 & 14 & 1.601 & 2.255 \\
	15 & 1.899 & 3.077 & 15 & 1.931 & 2.785 & 15 & 1.803 & 2.299 \\
	16 & --- & --- & 16 & 1.586 & 2.767 & 16 & 1.795 & 2.854 \\
	17 & 1.611 & 2.341 & 17 & 1.576 & 2.101 & 17 & 1.93 & 2.789 \\
	18 & 2.099 & 3.273 & 18 & 1.785 & 2.889 & 18 & 1.253 & 1.333 \\
	19 & 2.122 & 3.53 & 19 & 1.665 & 2.288 & 19 & 1.539 & 2.013 \\
	20 & 1.617 & 2.494 & 20 & 1.622 & 2.382 & 20 & 1.46 & 1.985 \\
	21 & 1.453 & 2.074 & 21 & 1.418 & 1.677 & 21 & 2.059 & 3.355 \\
	22 & 1.172 & 1.518 & 22 & 1.674 & 2.566 & 22 & 1.735 & 2.572 \\
	23 & 1.412 & 1.898 & 23 & 1.968 & 2.825 & 23 & 1.956 & 3.214 \\
	24 & 1.471 & 2.167 & 24 & 1.797 & 2.832 & 24 & 1.802 & 2.507 \\
	25 & 1.591 & 2.413 & 25 & 1.543 & 2.103 & 25 & 1.572 & 2.445 \\
	26 & 1.523 & 2.16 & 26 & 1.799 & 2.814 & 26 & 1.47 & 2.105 \\
	27 & 1.595 & 2.069 & 27 & 2.095 & 3.103 & 27 & 1.984 & 2.948 \\
	28 & 1.532 & 2.296 & 28 & 1.467 & 1.796 & 28 & 1.531 & 2.098 \\
	29 & 1.586 & 2.534 & 29 & 1.427 & 1.893 & 29 & 1.467 & 2.454 \\
\end{tabular}

\end{table}

\begin{table}[!t]
  \begin{center}
    \caption[Maximum velocity]
    {
      Maximum velocity magnitude (in m/s) per batch at $t = 2.16$ (peak systole),
      sample means and standard deviations.
    }

    \label{tab:u_216_max}

    \begin{tabular}{c|c|c||c|c|c||c|c|c}
$\alpha_r = 0.7$ & $\mu$ & $\sigma$ & $\alpha_r = 1.0$ & $\mu$ & $\sigma$ & $\alpha_r = 1.3$ & $\mu$ & $\sigma$ \\ \hline
0 & $4.108$ & $0.696$ & 0 & $1.750$ & $0.177$ & 0 & $1.368$ & $0.109$ \\
1 & $7.032$ & $0.828$ & 1 & $4.886$ & $0.800$ & 1 & $3.059$ & $0.250$ \\
2 & $5.395$ & $0.418$ & 2 & $3.850$ & $0.393$ & 2 & $2.815$ & $0.313$ \\
3 & $4.098$ & $0.142$ & 3 & $2.398$ & $0.076$ & 3 & $1.778$ & $0.072$ \\
4 & $5.198$ & $0.490$ & 4 & $3.903$ & $0.568$ & 4 & $2.590$ & $0.423$ \\
5 & $3.774$ & $0.260$ & 5 & $2.402$ & $0.465$ & 5 & $1.992$ & $0.346$ \\
6 & $5.424$ & $0.925$ & 6 & $3.801$ & $0.579$ & 6 & $2.712$ & $0.315$ \\
7 & --- & --- & 7 & $2.694$ & $0.228$ & 7 & $1.906$ & $0.151$ \\
8 & $3.031$ & $0.277$ & 8 & $2.412$ & $0.316$ & 8 & $2.147$ & $0.255$ \\
9 & $4.464$ & $0.602$ & 9 & $2.838$ & $0.209$ & 9 & $1.975$ & $0.278$ \\
10 & $1.811$ & $0.166$ & 10 & $1.081$ & $0.097$ & 10 & $0.857$ & $0.048$ \\
11 & $3.298$ & $0.419$ & 11 & $1.806$ & $0.189$ & 11 & $1.416$ & $0.130$ \\
12 & $8.675$ & $1.354$ & 12 & $4.576$ & $0.407$ & 12 & $3.708$ & $0.273$ \\
13 & $4.019$ & $0.562$ & 13 & $1.986$ & $0.148$ & 13 & $1.510$ & $0.066$ \\
14 & $4.504$ & $0.317$ & 14 & $2.483$ & $0.215$ & 14 & $1.725$ & $0.170$ \\
15 & $3.545$ & $0.358$ & 15 & $1.951$ & $0.221$ & 15 & $1.324$ & $0.069$ \\
16 & $5.503$ & --- & 16 & $3.097$ & $0.243$ & 16 & $2.078$ & $0.183$ \\
17 & $4.538$ & $0.548$ & 17 & $2.152$ & $0.205$ & 17 & $1.387$ & $0.202$ \\
18 & $4.159$ & $0.446$ & 18 & $2.663$ & $0.381$ & 18 & $1.920$ & $0.170$ \\
19 & $3.618$ & $0.675$ & 19 & $2.047$ & $0.122$ & 19 & $1.919$ & $0.065$ \\
20 & $5.007$ & $0.470$ & 20 & $3.814$ & $0.207$ & 20 & $3.074$ & $0.299$ \\
21 & $6.123$ & $0.625$ & 21 & $3.998$ & $0.257$ & 21 & $2.831$ & $0.430$ \\
22 & $4.122$ & $0.268$ & 22 & $4.312$ & $0.742$ & 22 & $2.568$ & $0.348$ \\
23 & $4.306$ & $0.374$ & 23 & $3.040$ & $0.562$ & 23 & $2.301$ & $0.220$ \\
24 & $5.132$ & $0.423$ & 24 & $2.810$ & $0.223$ & 24 & $1.669$ & $0.154$ \\
25 & $5.801$ & $0.695$ & 25 & $2.370$ & $0.260$ & 25 & $1.819$ & $0.150$ \\
26 & $5.828$ & $0.626$ & 26 & $4.139$ & $0.511$ & 26 & $3.592$ & $0.295$ \\
27 & $2.835$ & $0.236$ & 27 & $1.646$ & $0.129$ & 27 & $0.978$ & $0.049$ \\
28 & $3.730$ & $0.518$ & 28 & $2.405$ & $0.265$ & 28 & $1.700$ & $0.141$ \\
29 & $6.422$ & $0.965$ & 29 & $3.744$ & $0.326$ & 29 & $2.880$ & $0.159$ \\
\end{tabular}

  \end{center}
\end{table}

\begin{table}[!t]
  \begin{center}
    \caption[Maximum pressure drop]
    {
      Largest pressure difference between inlet and descending aorta over one cardiac cycle (in mmHg),
      sample means and standard deviations.
    }

    \label{tab:P_rel_min}

    \begin{tabular}{c|c|c||c|c|c||c|c|c}
$\alpha_r = 0.7$ & $\mu$ & $\sigma$ & $\alpha_r = 1.0$ & $\mu$ & $\sigma$ & $\alpha_r = 1.3$ & $\mu$ & $\sigma$ \\ \hline
0 & $28.644$ & $1.120$ & 0 & $11.824$ & $0.427$ & 0 & $8.618$ & $0.377$ \\
1 & $88.027$ & $12.313$ & 1 & $43.453$ & $6.466$ & 1 & $21.899$ & $0.787$ \\
2 & $63.888$ & $5.698$ & 2 & $36.947$ & $4.547$ & 2 & $20.059$ & $1.510$ \\
3 & $42.669$ & $1.190$ & 3 & $20.990$ & $0.762$ & 3 & $13.680$ & $0.311$ \\
4 & $40.048$ & $6.825$ & 4 & $21.498$ & $2.175$ & 4 & $14.061$ & $0.969$ \\
5 & $29.633$ & $1.412$ & 5 & $16.304$ & $0.374$ & 5 & $12.987$ & $0.387$ \\
6 & $47.970$ & $5.846$ & 6 & $29.287$ & $3.956$ & 6 & $19.765$ & $1.208$ \\
7 & --- & --- & 7 & $12.421$ & $0.615$ & 7 & $8.622$ & $0.392$ \\
8 & $31.649$ & $4.326$ & 8 & $19.859$ & $1.453$ & 8 & $13.917$ & $0.595$ \\
9 & $41.759$ & $6.948$ & 9 & $23.142$ & $1.697$ & 9 & $14.836$ & $0.796$ \\
10 & $20.012$ & $0.942$ & 10 & $11.997$ & $0.444$ & 10 & $9.477$ & $0.361$ \\
11 & $21.471$ & $1.368$ & 11 & $11.015$ & $0.429$ & 11 & $9.040$ & $0.379$ \\
12 & $111.399$ & $22.018$ & 12 & $31.068$ & $4.557$ & 12 & $22.435$ & $1.983$ \\
13 & $42.676$ & $6.047$ & 13 & $20.369$ & $1.047$ & 13 & $14.811$ & $0.413$ \\
14 & $45.675$ & $4.582$ & 14 & $22.484$ & $1.245$ & 14 & $14.173$ & $0.314$ \\
15 & $33.101$ & $2.233$ & 15 & $17.511$ & $0.476$ & 15 & $11.313$ & $0.386$ \\
16 & $57.824$ & --- & 16 & $23.487$ & $1.808$ & 16 & $14.042$ & $0.675$ \\
17 & $31.135$ & $2.935$ & 17 & $18.584$ & $0.527$ & 17 & $11.783$ & $0.375$ \\
18 & $37.853$ & $4.359$ & 18 & $19.838$ & $0.644$ & 18 & $14.422$ & $0.452$ \\
19 & $30.507$ & $4.736$ & 19 & $15.413$ & $0.421$ & 19 & $12.476$ & $0.321$ \\
20 & $59.358$ & $10.298$ & 20 & $31.828$ & $2.152$ & 20 & $20.841$ & $0.719$ \\
21 & $53.520$ & $5.594$ & 21 & $30.180$ & $1.310$ & 21 & $17.860$ & $1.253$ \\
22 & $46.991$ & $3.758$ & 22 & $33.863$ & $3.379$ & 22 & $18.590$ & $0.829$ \\
23 & $41.321$ & $4.160$ & 23 & $24.068$ & $2.784$ & 23 & $14.496$ & $0.500$ \\
24 & $41.528$ & $2.528$ & 24 & $22.245$ & $0.666$ & 24 & $14.692$ & $0.287$ \\
25 & $49.545$ & $7.475$ & 25 & $23.881$ & $0.559$ & 25 & $16.854$ & $0.383$ \\
26 & $70.156$ & $5.729$ & 26 & $35.460$ & $3.147$ & 26 & $25.416$ & $1.248$ \\
27 & $28.887$ & $2.554$ & 27 & $18.003$ & $0.955$ & 27 & $12.013$ & $0.337$ \\
28 & $34.776$ & $2.610$ & 28 & $19.231$ & $1.041$ & 28 & $14.515$ & $0.503$ \\
29 & $86.733$ & $19.557$ & 29 & $39.778$ & $3.418$ & 29 & $22.793$ & $0.496$ \\
\end{tabular}

  \end{center}
\end{table}

\begin{table}[!t]
  \begin{center}
    \caption[Mean WSS magnitude]
    {
      Mean wall shear stress magnitude over the whole blood vessel wall and one cardiac cycle (in Pa),
      sample means and standard deviations.
    }

    \label{tab:wss_mag_mean}

    \begin{tabular}{c|c|c||c|c|c||c|c|c}
$\alpha_r = 0.7$ & $\mu$ & $\sigma$ & $\alpha_r = 1.0$ & $\mu$ & $\sigma$ & $\alpha_r = 1.3$ & $\mu$ & $\sigma$ \\ \hline
0 & $3.671$ & $0.273$ & 0 & $1.532$ & $0.133$ & 0 & $1.137$ & $0.118$ \\
1 & $7.392$ & $0.968$ & 1 & $4.333$ & $0.682$ & 1 & $2.409$ & $0.248$ \\
2 & $7.223$ & $0.659$ & 2 & $4.291$ & $0.337$ & 2 & $2.669$ & $0.342$ \\
3 & $4.276$ & $0.172$ & 3 & $2.173$ & $0.132$ & 3 & $1.327$ & $0.086$ \\
4 & $4.931$ & $0.604$ & 4 & $2.533$ & $0.192$ & 4 & $1.699$ & $0.190$ \\
5 & $3.113$ & $0.155$ & 5 & $1.478$ & $0.123$ & 5 & $1.097$ & $0.081$ \\
6 & $5.560$ & $0.719$ & 6 & $3.488$ & $0.590$ & 6 & $2.341$ & $0.321$ \\
7 & --- & --- & 7 & $2.849$ & $0.360$ & 7 & $1.719$ & $0.241$ \\
8 & $2.960$ & $0.513$ & 8 & $1.800$ & $0.202$ & 8 & $1.309$ & $0.092$ \\
9 & $5.386$ & $0.791$ & 9 & $2.949$ & $0.312$ & 9 & $1.663$ & $0.145$ \\
10 & $1.678$ & $0.137$ & 10 & $1.018$ & $0.093$ & 10 & $0.766$ & $0.043$ \\
11 & $2.766$ & $0.374$ & 11 & $1.272$ & $0.107$ & 11 & $1.003$ & $0.115$ \\
12 & $10.617$ & $1.704$ & 12 & $4.292$ & $0.516$ & 12 & $2.983$ & $0.365$ \\
13 & $4.313$ & $0.751$ & 13 & $2.020$ & $0.208$ & 13 & $1.331$ & $0.081$ \\
14 & $3.927$ & $0.410$ & 14 & $2.129$ & $0.183$ & 14 & $1.355$ & $0.114$ \\
15 & $3.781$ & $0.452$ & 15 & $1.933$ & $0.197$ & 15 & $1.064$ & $0.082$ \\
16 & $6.217$ & --- & 16 & $3.235$ & $0.359$ & 16 & $1.821$ & $0.255$ \\
17 & $4.229$ & $0.523$ & 17 & $2.135$ & $0.152$ & 17 & $1.369$ & $0.117$ \\
18 & $4.053$ & $0.539$ & 18 & $2.027$ & $0.161$ & 18 & $1.403$ & $0.095$ \\
19 & $3.061$ & $0.514$ & 19 & $1.635$ & $0.126$ & 19 & $1.258$ & $0.047$ \\
20 & $6.283$ & $1.070$ & 20 & $3.493$ & $0.415$ & 20 & $2.208$ & $0.111$ \\
21 & $4.877$ & $0.388$ & 21 & $3.137$ & $0.141$ & 21 & $1.820$ & $0.256$ \\
22 & $4.686$ & $0.303$ & 22 & $3.414$ & $0.352$ & 22 & $2.162$ & $0.214$ \\
23 & $4.784$ & $0.438$ & 23 & $2.653$ & $0.471$ & 23 & $1.795$ & $0.200$ \\
24 & $5.091$ & $0.576$ & 24 & $2.714$ & $0.183$ & 24 & $1.707$ & $0.111$ \\
25 & $4.832$ & $0.728$ & 25 & $2.225$ & $0.122$ & 25 & $1.579$ & $0.081$ \\
26 & $6.546$ & $0.732$ & 26 & $3.611$ & $0.409$ & 26 & $2.408$ & $0.146$ \\
27 & $2.839$ & $0.340$ & 27 & $1.616$ & $0.167$ & 27 & $0.931$ & $0.064$ \\
28 & $4.247$ & $0.374$ & 28 & $2.242$ & $0.213$ & 28 & $1.593$ & $0.130$ \\
29 & $6.724$ & $1.170$ & 29 & $3.364$ & $0.328$ & 29 & $2.105$ & $0.115$ \\
\end{tabular}

  \end{center}
\end{table}

\begin{table}[!t]
  \begin{center}
    \caption[Reference patch OSI]
    {
      Mean oscillatory shear index over a reference patch on the underside of the descending aorta,
      sample means and standard deviations.
    }

    \label{tab:osi_patch}

    \begin{tabular}{c|c|c||c|c|c||c|c|c}
$\alpha_r = 0.7$ & $\mu$ & $\sigma$ & $\alpha_r = 1.0$ & $\mu$ & $\sigma$ & $\alpha_r = 1.3$ & $\mu$ & $\sigma$ \\ \hline
0 & $0.260$ & $0.019$ & 0 & $0.209$ & $0.022$ & 0 & $0.231$ & $0.014$ \\
1 & $0.119$ & $0.060$ & 1 & $0.178$ & $0.057$ & 1 & $0.250$ & $0.040$ \\
2 & $0.142$ & $0.020$ & 2 & $0.081$ & $0.024$ & 2 & $0.155$ & $0.016$ \\
3 & $0.307$ & $0.027$ & 3 & $0.226$ & $0.039$ & 3 & $0.197$ & $0.019$ \\
4 & $0.077$ & $0.045$ & 4 & $0.044$ & $0.014$ & 4 & $0.088$ & $0.028$ \\
5 & $0.114$ & $0.048$ & 5 & $0.125$ & $0.017$ & 5 & $0.239$ & $0.023$ \\
6 & $0.222$ & $0.030$ & 6 & $0.243$ & $0.025$ & 6 & $0.203$ & $0.040$ \\
7 & --- & --- & 7 & $0.131$ & $0.017$ & 7 & $0.181$ & $0.022$ \\
8 & $0.149$ & $0.027$ & 8 & $0.144$ & $0.024$ & 8 & $0.163$ & $0.026$ \\
9 & $0.056$ & $0.029$ & 9 & $0.050$ & $0.009$ & 9 & $0.082$ & $0.013$ \\
10 & $0.246$ & $0.033$ & 10 & $0.243$ & $0.024$ & 10 & $0.191$ & $0.023$ \\
11 & $0.213$ & $0.035$ & 11 & $0.201$ & $0.016$ & 11 & $0.225$ & $0.013$ \\
12 & $0.225$ & $0.028$ & 12 & $0.156$ & $0.024$ & 12 & $0.152$ & $0.023$ \\
13 & $0.137$ & $0.052$ & 13 & $0.117$ & $0.033$ & 13 & $0.149$ & $0.018$ \\
14 & $0.220$ & $0.057$ & 14 & $0.197$ & $0.063$ & 14 & $0.159$ & $0.025$ \\
15 & $0.219$ & $0.049$ & 15 & $0.227$ & $0.011$ & 15 & $0.211$ & $0.026$ \\
16 & $0.045$ & --- & 16 & $0.102$ & $0.038$ & 16 & $0.155$ & $0.028$ \\
17 & $0.273$ & $0.026$ & 17 & $0.248$ & $0.012$ & 17 & $0.236$ & $0.015$ \\
18 & $0.120$ & $0.028$ & 18 & $0.116$ & $0.040$ & 18 & $0.115$ & $0.019$ \\
19 & $0.220$ & $0.043$ & 19 & $0.254$ & $0.022$ & 19 & $0.216$ & $0.018$ \\
20 & $0.149$ & $0.039$ & 20 & $0.144$ & $0.062$ & 20 & $0.170$ & $0.038$ \\
21 & $0.244$ & $0.020$ & 21 & $0.230$ & $0.027$ & 21 & $0.203$ & $0.042$ \\
22 & $0.185$ & $0.031$ & 22 & $0.218$ & $0.027$ & 22 & $0.204$ & $0.013$ \\
23 & $0.171$ & $0.037$ & 23 & $0.226$ & $0.018$ & 23 & $0.278$ & $0.027$ \\
24 & $0.237$ & $0.041$ & 24 & $0.245$ & $0.019$ & 24 & $0.270$ & $0.012$ \\
25 & $0.174$ & $0.042$ & 25 & $0.118$ & $0.020$ & 25 & $0.164$ & $0.019$ \\
26 & $0.119$ & $0.029$ & 26 & $0.148$ & $0.021$ & 26 & $0.237$ & $0.041$ \\
27 & $0.138$ & $0.012$ & 27 & $0.184$ & $0.026$ & 27 & $0.177$ & $0.031$ \\
28 & $0.097$ & $0.016$ & 28 & $0.192$ & $0.035$ & 28 & $0.208$ & $0.012$ \\
29 & $0.132$ & $0.024$ & 29 & $0.147$ & $0.030$ & 29 & $0.208$ & $0.024$ \\
\end{tabular}

  \end{center}
\end{table}




\end{document}